\documentclass{article}
\PassOptionsToPackage{numbers, compress}{natbib}
\usepackage[final]{neurips_2026}
\makeatletter
\providecommand{\@trackname}{\@neuripsordinal\ Conference on Neural Information Processing Systems (NeurIPS \@neuripsyear).}
\makeatother

\usepackage[utf8]{inputenc} % allow utf-8 input
\usepackage[T1]{fontenc}    % use 8-bit T1 fonts
\usepackage{hyperref}       % hyperlinks
\usepackage{url}            % simple URL typesetting
\usepackage{booktabs}       % professional-quality tables
\usepackage{amsfonts}       % blackboard math symbols
\usepackage{amsmath}        % math environments and \text{}
\usepackage{amsthm}         % theorem environments
\usepackage{nicefrac}       % compact symbols for 1/2, etc.
\usepackage{microtype}      % microtypography
\usepackage{todonotes}

\usepackage{booktabs}
\usepackage{multirow}
\usepackage{graphicx}
\usepackage{subcaption}
\usepackage{makecell}
\usepackage{xspace}
\usepackage{pgfplots}
\usepgfplotslibrary{groupplots}  % Load the library
\pgfplotsset{compat=1.18}  
\usepackage{listings}

\usepackage{algorithm}
\usepackage{algpseudocode}

\usepackage{enumitem}
\usepackage{wrapfig}
\usepackage{caption}

\definecolor{ragcolor}{HTML}{FFB74D}
\definecolor{gptcolor}{HTML}{81D4FA}
\definecolor{rlmcolor}{HTML}{CE93D8}
\definecolor{sliderscolor}{HTML}{8C1515}

\usepackage{amssymb}

\usepackage{minitoc} 
\usepackage{tikz}
\usetikzlibrary{calc}
\usepackage{pgfplots}
\usepackage{subcaption}
\pgfplotsset{compat=1.18}
\definecolor{opTeal}{RGB}{26,188,156}
\definecolor{opBlue}{RGB}{52,152,219}
\definecolor{opRed}{RGB}{231,76,60}
\definecolor{opOrange}{RGB}{243,156,18}
\definecolor{opYellow}{RGB}{241,196,15}
\definecolor{opGreen}{RGB}{46,204,113}
\definecolor{opPurple}{RGB}{155,89,182}

\definecolor{stageone}{HTML}{666666}
\definecolor{stagetwo}{HTML}{7B68AE}
\definecolor{stagethree}{HTML}{3A7BD5}
\definecolor{stagefour}{HTML}{D4880F}
\definecolor{stagefive}{HTML}{1A7F64}

\usepackage[table]{xcolor}

\definecolor{ScoreGreen}{HTML}{2E7D32}

\newcommand{\system}{SLIDERS\xspace}

\newcommand{\EE}{\textrm{EE}}

\newcommand{\EEllm}{\textrm{EE}_\textrm{LLM}}

\title{SLIDERS: Evidence Synthesis with Autonomous Database Reconciliation for Systematic Reviews\thanks{Code: \url{https://github.com/stanford-oval/sliders} \quad Website: \url{https://sliders.genie.stanford.edu/}}}
\author{
  Harshit Joshi \quad Priyank Shethia \quad Jadelynn Dao \quad Monica S. Lam \\
  Computer Science Department, Stanford University\\
  \texttt{\{hj, lam\}@cs.stanford.edu} \\
}

\begin{document}
\doparttoc
\faketableofcontents

\maketitle

\begin{abstract}
Systematic reviews requires comprehensive evidence synthesis from large document corpora to answer targeted research questions and are foundational in finance, social sciences, and other technical fields. 
Manual construction of evidence tables is labor-intensive, and recent LLM-based assistants relying on embedding or keyword search often fail to meet the coverage standards of systematic reviews.
We introduce SLIDERS, a novel LLM-based methodology for automatically assembling evidence tables tailored to research questions for systematic reviews. 
SLIDERS extracts full-text excerpts that serve as direct evidence or as provenance for structured data. 
Core to SLIDERS is an autonomous evidence reconciliation coding agent that iteratively writes SQL to analyze and reconcile extracted evidence, bringing together information fragmented across documents, resolving inconsistencies and removing duplicates across records.
We evaluate \system{} on BrowseComp+, achieving 90\% accuracy, and show that it outperforms all baselines on additional benchmarks within the LLM’s context window. To assess scalability, we introduce two new evidence-synthesis benchmarks, WikiCeleb100 (3.9M tokens) and FinQ100 (36M tokens), which include follow-up questions beyond the original research query. Although \system{} synthesizes evidence knowing only the original question, it answers 78.9\% and 55.2\% of all questions correctly on WikiCeleb100 and FinQ100, respectively.

\end{abstract}

\begin{figure}[h]
    \centering
    \includegraphics[width=\linewidth]{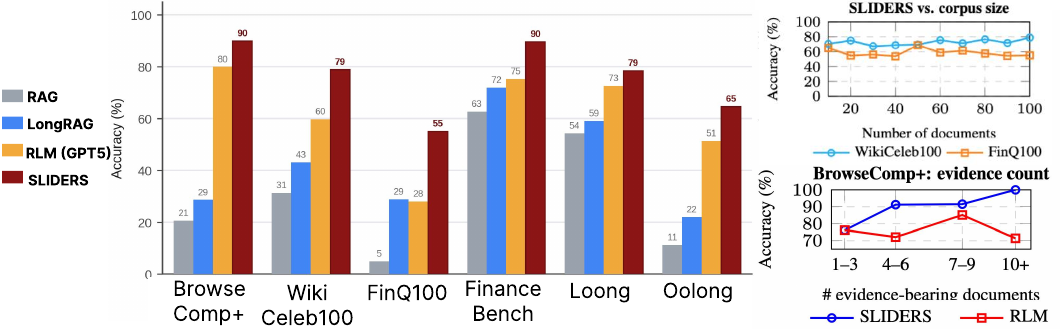}
    \caption{SLIDERS, constructs a database from large set of documents and autonomously reconciles inconsistencies through iterative investigation. (left) It outperforms all baselines including RAG, LongRAG and RLM (GPT-5) on all six benchmarks  and (right) maintains high accuracy as number of documents and evidence dispersion per question increases.}
    \label{fig:placeholder}
\end{figure}

\section{Introduction}

Systematic reviews are foundational to evidence-based research in domains such as medicine, social sciences, policy, finance, and law~\cite{petticrew2006systematic,tranfield2003towards,quinn2025systematic,baude2017making,keele2007guidelines}. By rigorously synthesizing evidence to answer targeted research questions, they inform scientists and decision-makers in research, policy, and practice. However, conducting a high-quality systematic review is a complex, resource-intensive undertaking. The process involves identifying eligible studies, extracting and standardizing relevant data, and integrating findings into a coherent synthesis, while adhering to transparent reporting standards~\cite{lasserson2024starting,chandler2019cochrane,page2021prisma}. Each stage requires substantial expert effort. A recent analysis indicates a mean completion time of over a year (67.3 weeks) per review~\cite{borah2017analysis}. Even after narrowing down eligible studies, reviewers must manually extract data from numerous sources, with domain specific averages ranging from 20 to 72 documents~\cite{wu2026can}, making the process highly labor-intensive and time-consuming~\cite{li2019randomized,gartlehner2024data}.
This investment of time and effort represents a major barrier to timely and scalable knowledge synthesis.

We address the challenge of automating information extraction from scientific documents after they have been identified as relevant. LLMs hold considerable promise for accelerating evidence synthesis~\cite{riis2025chatgpt,gartlehner2024data}, but their use  poses important challenges. First, when processing lengthy documents or extensive contexts, LLMs may miss or inadequately extract relevant information, leading to incomplete or inaccurate results~\cite{smyth2026quantifying,goldman2024really,bertsch2025oolong}. Second, the combined length of all documents considered in a systematic review often exceeds the LLM’s context window, making it impractical to analyze all texts simultaneously in a single prompt~\cite{liang2025surveyx,wang-etal-2026-arxiv2table}. Overcoming these limitations is crucial to fully realizing the potential of LLMs for large-scale, automated evidence synthesis.

This paper examines a key question for automating evidence synthesis: How can we accurately extract all relevant evidence from a document collection that may exceed an LLM’s context window and answer questions in systematic reviews? One common approach is RAG, which selects passages relevant to a given question~\citep{lewis2020retrieval,jiang2024longrag,edge2024graphrag}. However, this selective retrieval can omit evidence necessary for a comprehensive synthesis. While retrieving more passages can increase coverage, it also further constrains the available context length. Alternatively, chunk-based extraction processes documents in segments and collects structured records~\citep{zhang2024coa,docetl,zhao2024longagent}. For instance, DocETL extracts records according to user-defined schemas~\citep{docetl}. Yet as the number of extracted records grows, their combined length may also exceed the context window, rendering simple concatenation infeasible for large collections.

{\bf First contribution: SLIDERS evidence synthesis agent.} This paper introduces SLIDERS, a fully automated, scalable, and database-centric agent for evidence synthesis from large document collections. Similar to previous chunk-based extraction methods, SLIDERS extracts information chunk by chunk and stores it in a database. However, SLIDERS introduces several key innovations. First, it automatically determines the database schema, eliminating the need for manual design. Recognizing that users commonly refine their queries or pose follow-up questions as they interact with results, SLIDERS proactively designs the schema to accommodate a broad range of related and subsequent inquiries, greatly enhancing flexibility. This allows the cost of extraction be amortized across the different queries. In addition, to mitigate issues associated with out-of-context analysis, SLIDERS carefully constructs the context for each chunk.

Second, and more importantly, empirical findings indicate that chunk-based extraction will inevitably produce a very ``dirty'' database, one in which the extracted information is often redundant, contradictory, or incomplete, all of which must be reconciled for comprehensive synthesis. To address this challenge, \system{} incorporates an autonomous coding agent that systematically reconciles and curates the database.

The resulting clean, reconciled database serves as a powerful foundation for SLIDERS’ third key innovation: flexible and dynamic exploration of evidence. By integrating LLM language understanding with the power of database queries, SLIDERS can answer not only the original review questions but also a wide range of follow-up inquiries without the need for repeated extraction.
In sum, SLIDERS fully leverages the scalability and versatility of databases to efficiently store, curate, and answer questions across arbitrarily large collections of long documents.

Database cleaning and reconciliation is a longstanding area of research, encompassing various approaches for error detection and repair\citep{chu2016data}. This includes traditional methods rooted in statistical regularities and integrity constraints\citep{beskales2010sampling, rekatsinas2017holoclean}, as well as more recent LLM-based techniques designed to identify typos and outliers at the column level\citep{zhang2025data}, or approaches leveraging labeled data to surface underlying data quality issues\citep{ni2024iterclean, ahmad2023retclean}.
However, these methods often focus narrowly on resolving column-level inconsistencies, rather than addressing the gross redundancies, inconsistencies, and incompleteness caused by chunk-wise extraction. 

Effective reconciliation relies on leveraging the full text, just as LLMs do when documents fit within the context window. To support this at scale, SLIDERS is designed as an index, precisely recording the source text spans and reasoning for each extracted value. This design ensures that sources remains accessible for reconciliation, even when the document collection exceeds the LLM’s context window.

\textbf{Second contribution: An effective autonomous database reconciliation coding agent.} 

Despite the sophistication of today’s LLMs, instructing them to directly reconcile databases is not viable. For example, LLMs would attempt to ingest the entire database in their context, which does not scale, and their justifications for resolving conflicting data can diverge from users’ objectives.

To address these challenges, we have developed a robust reconciliation strategy encoded directly in the agent’s prompt. At scale, the agent issues tailored SQL commands for each database table, first identifying reconciliation issues and then resolving them based on observed data. The process begins with grouping records by primary keys, consolidating related facts from different sections of a single document or across multiple documents. The agent's instructions specify strategies for addressing a range of error types, including duplications, inconsistencies, and partial or complementary information, using detailed analysis of supporting text spans and extraction rationales. Additionally, the prompt outlines resolution priorities, such as preferring values derived from direct, reliable quotations over those that are computed. Crucially, the reconciliation code adapts dynamically to the structure and anomalies of the data, enabling context-aware and scalable curation.

To ensure the fidelity of edits, the agent is further instructed to verify its changes by comparing the proposed database updates against the intended update. If modifications deviate, they are automatically reverted and the agent receives feedback for refinement. This iterative loop enables robust, automated, and context-sensitive database reconciliation without human supervision.

\begin{figure}[t]
    \centering
    \includegraphics[width=0.99\linewidth]{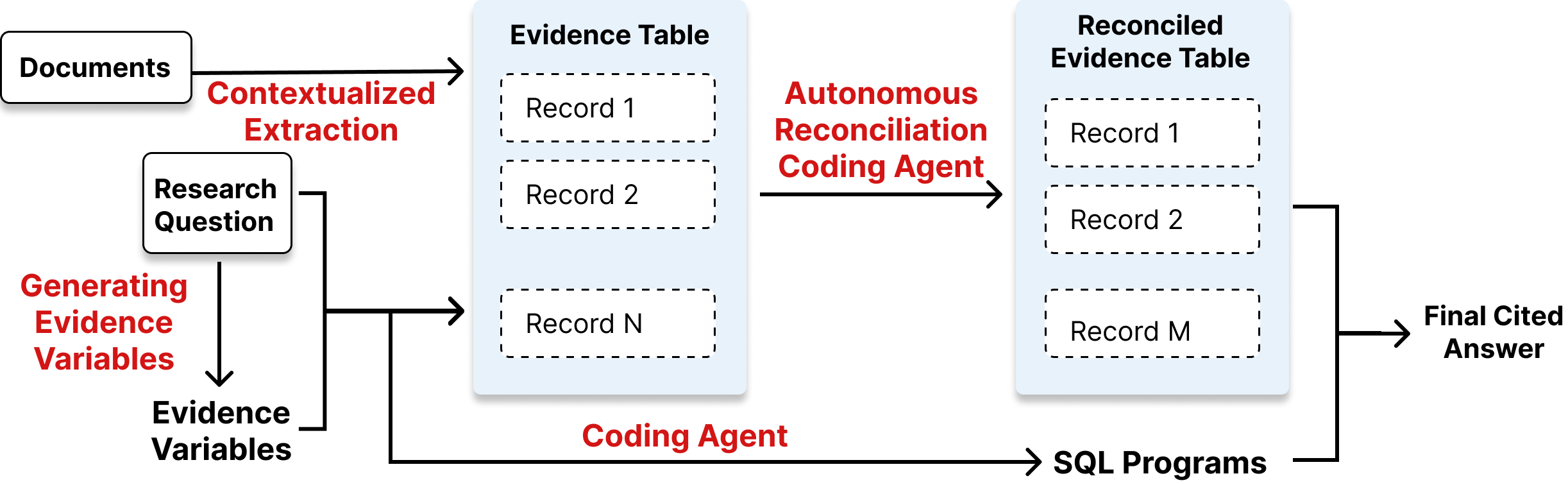}
    \caption{\system{} extracts evidence into a database with links to supporting source spans. An autonomous agent investigates the database, identifies reconciliation tasks, and updates records using their source context. The reconciled table supports synthesis and follow-up analysis.}
    \label{fig:main-intro}
    \vspace{-4mm}
\end{figure}

{\bf Third contribution: a large-scale evidence management dataset.} In real-world applications, systematic reviews are often conducted on datasets that far exceed the context windows of current LLMs. Yet, to our knowledge, there are no existing datasets of this scale, and none that support multi-turn interactions or follow-up questioning. This paper introduces two new benchmarks: WikiCeleb100 and FinQ100, each comprising 100 documents drawn from Wikipedia and 10-Q financial filings, respectively. We further evaluate \system{} on BrowseComp+~\cite{chen2025browsecomp}, a deep-research benchmark, as well as on other document QA datasets that fit within an LLM’s context window.
Our results show that \system{} outperforms the strongest baselines by 19 and 27 percentage points on WikiCeleb100 and FinQ100, respectively. Notably, \system{} also achieves superior performance on tasks where the input documents fit entirely within the LLM’s context window.

\section{Related Works}

\textbf{Long context modeling.}
Training-based approaches expand model capacity via position-agnostic training~\citep{he2024never}, multi-document instruction tuning~\citep{liu2025mdcure}, or representation-level compression into fixed working memory~\citep{eyuboglu2025cartridges,yu2025memagent}.
Agentic systems instead divide work across roles or recursive sub-tasks~\citep{zhang2024coa,zhao2024longagent,li2024graphreader,narayan2025minions,zhang2025recursivelanguagemodels}.
\system{} is complementary to these efforts.
Elicit~\citep{elicit2026systematic} supports search, screening, data extraction and synthesis with sentence-level citations across up to 200 papers. \system{} focuses on scaling to several documents and automatically reconciling extracted evidence.

\textbf{Chunking Methods.}
RAG grounds LLM outputs by retrieving chunks from an external corpus before synthesis~\citep{lewis2020retrieval,guu2020retrieval,karpukhin2020dense,izacard-grave-2021-leveraging,khattab2020colbert,izacard2021unsupervised}.
Long-context, adaptive, iterative, graph-based, and hierarchy-based variants improve retrieval scope or organization~\citep{jiang2024longrag,edge2024graphrag,liu2024ra,jiang2023active,trivedi-etal-2023-interleaving,asai2023self,yan2024corrective,sarthi2024raptor,sarmah2024hybridrag,guo2024lightrag,gutierrez2024hipporag}.
However, using embedding-based similarity match can miss several several key details as evidence maybe temporal or spread across several documents.
Once can retrieve more chunks, but simply expanding context also does not ensure reliable use of all relevant evidence~\citep{Liu2023LostMiddle}.
In contrast, \system{} derives evidence variables from the research question and extracts evidence from all the chunks that are relevant to the problem.
Other chunking methods, scale by splitting inputs into chunks and processing each chunk independently like \system{}, but aggregating the outputs by providing everything to the LLM. 
This pattern appears in divide-and-conquer summarization~\citep{gidiotis2020divide}, LLM MapReduce pipelines~\citep{zhou2025llmmapreduce}, optimized document-processing workflows~\citep{docetl}, domain-specific extraction systems~\citep{han2024xbrl,wang2025finsage}, and hierarchical or hybrid retrieval methods~\citep{sarmah2024hybridrag,choe2025hirec}. While these approaches scale for extraction, when the number of evidences grows, they are also constrained by LLM's context window.  \system{} instead keeps chunk-level extractions in database and performs database reconciliation using the coding agent.

\section{The SLIDERS Method}
\label{sec:methodology}

\system{} constructs a database of document-grounded evidence and autonomously reconciles inconsistencies through iterative investigation. The framework proceeds as follows: First, it automatically generates evidence variables and extracts structured records from document chunks. Next, a reconciliation agent queries the database, examines supporting source passages, and writes custom code to systematically group related records and resolve redundancy, contradictions, or incompleteness. Finally, a response synthesis agent leverages the curated database to answer both the initial research question and an array of related follow-up inquiries.

\textbf{Task Formulation. }
Given a research question $q$ and a set of eligible documents $\mathcal{D}$, the objective is to construct a reusable evidence database $\mathcal{Z}_{q}$ that enables robust answers to the research question using the curated evidence.
The evidence database may contain multiple evidence tables. Each table is structured with columns corresponding to question-relevant evidence variables and rows representing document-backed evidence records; each record is linked to its source document, supporting text span, and provenance metadata. The evidence variables may capture typed values (e.g., dates, categories, measurements) as well as free-text fields (e.g., rationales, or descriptions).
Crucially, the database is designed to be reusable across related research questions. At answer time, a coding agent dynamically reorganizes the evidence to produce the precise view required to address the current question.

\subsection{Automated Generation of Evidence Variables}
\label{sec:generating-dimensions}
A central innovation of \system{} is its ability to automatically generate a tailored set of evidence variables directly from the research question and a representative sample of the document metadata.

\textbf{Definition 1.}
The \emph{Evidence Variable Generation Task} takes a question $q$ and document metadata $\mathcal{M}$, and returns a set of evidence variables for the evidence table:
$$\mathcal{V}_{q, \mathcal{M}} = \{v_1, \ldots, v_n\}, \quad \text{where each }
v = \langle \textit{vn}, d, \tau, \lambda, \iota \rangle.$$
Here, $\textit{vn}$ is the evidence variable name; $d$ is a semantic description; $\tau$ is the data type; $\lambda$ specifies the normalization rule; and $\iota$ is the extraction instruction for the variable.

All evidence variables are strongly typed, enabling support for precise selection, aggregation, filtering, and comparison during synthesis.
Normalization rules, such as currency conversion or date formatting, are included to ensure consistency and comparability between independently extracted excerpts.
Furthermore, each variable is recorded alongside its supporting source text, rationale, and detailed provenance, ensuring that all synthesized answers can be directly traced back to the original document language.
This design enables transparent, document-grounded synthesis and interactive re-exploration of the evidence database for subsequent queries.

We instruct the LLM to extract not only variables relevant to the immediate research question, but also broader latent attributes that can support future or follow-up queries. This results in a flexible evidence schema that serves both current and anticipated information needs.

By building a comprehensive evidence table up front, SLIDERS enables subsequent questions to be answered rapidly through database queries, eliminating the need to reprocess the original documents. For example, in financial research, a question about long-term borrowing may prompt SLIDERS to extract variables such as company, sector, debt instrument, stated reason, and time period—attributes that seamlessly support a variety of related follow-up questions.

\subsection{Extraction into the Evidence Table}
\label{sec:contextualized_extraction_gating}

\textbf{Definition 2.}
The \emph{Evidence Extraction Task}, \EE, takes a question $q$, document set $\mathcal{D}$, document metadata $\mathcal{M}$, and generated evidence variables $\mathcal{V}_{q,\mathcal{M}} = \{v_1,\ldots,v_n\}$,
and produces an evidence table\ $\mathcal{Z}_q$.
An evidence table is a set of evidences:
$
z = \langle c, m_c, \{e_{v_j}\}_{j=1}^n \rangle,
$
where $c$ is the source chunk, $m_c$ is chunk metadata, and $e_{v_j}$ is the entry associated with evidence variable $v_j$.
Each entry is a tuple
$\langle v, p, \gamma \rangle,$
where $v$ is the extracted value normalized according to the type, unit, and normalization rule of the dimension, $p$ is the relevant source span supporting the value, and $\gamma$ is a rationale linking the span to the extracted value.
For free-text dimensions, $v$ may be an excerpt, claim, caveat, event or support/contradiction note, while $p$ keeps the original excerpt for inspection and citation.
\[
\begin{aligned}
\EE(q, \mathcal{D}, \mathcal{M}, \mathcal{V}_{q, \mathcal{M}})
&= \bigcup_{d \in \mathcal{D},\, c \in \mathcal{C}_d}
\EE(q, c, m_c, \mathcal{V}_{q, \mathcal{M}}),\\
\EE(q, c, m_c, \mathcal{V}_{q, \mathcal{M}})
&=
\begin{cases}
\EEllm(q, c, m_c, \mathcal{V}_{q, \mathcal{M}}) & \text{if } R(q, c, m_c, \mathcal{V}_{q, \mathcal{M}}),\\
\emptyset & \text{otherwise}.
\end{cases}
\end{aligned}
\]
Here $R(\cdot)$ is the relevance gate, and $\EEllm$ is the extraction module.

We determine empirically that when a chunk contains no relevant excerpt for an evidence variable, an in-context extractor may hallucinate a plausible but unsupported value because the output format forces the model to conform to the schema~\citep{liu2023evaluating,bbc2025hallucination}.
To maintain high precision, we introduce a \emph{relevance gate}. 
Before extraction, the model first decides whether the chunk contains relevant evidence that can instantiate at least one generated evidence variable. 
Extraction is invoked only when the gate passes; otherwise the chunk contributes no excerpts to the evidence table.

The extraction task over the document set can be parallelized, as it operates independently on each chunk. For details on our technique for constructing metadata that enriches context and improves extraction quality, please refer to Appendix~\ref{app:chunking}.

\subsection{Autonomous Evidence Reconciliation}
\label{sec:answer_time_curation}
Once the initial evidence table has been assembled through chunk-based extraction, it inevitably contains a high-recall but ``dirty'' collection of document-grounded records.
Entries are often redundant, fragmented, or even contradictory, and information relevant to a single question may be scattered across multiple segments~\citep{sollaci2004introduction,mann1988rhetorical}. Moreover, as the scale of the document set grows, the raw evidence table can far exceed the context capacity of large language models.

\textbf{Definition 3.}
The \emph{Reconciliation Task} takes a question $q$, evidence variables $\mathcal{V}_{q,\mathcal{M}}$, and a raw evidence table $\mathcal{Z}_{q}$ produced by chunk-level extraction, and returns a reconciled evidence table:
\[
\mathcal{Z}^{\textrm{R}}_{q}
=
\mathsf{Reconcile}(q,\mathcal{V}_{q, \mathcal{M}},\mathcal{Z}_{q}).
\]
The reconciled evidence table $\mathcal{Z}^{\textrm{R}}_{q}$ preserves the provenance spans, excerpts, and extraction rationales of the original records while organizing them into the final representation.

\paragraph{Discovering reconciliation tasks.}
We aim to develop an agent that operates with full autonomy, proactively identifying reconciliation needs by querying the database, rather than relying on pre-defined templates.
To accurately identify issues, the agent must examine both the values and their corresponding source documents, especially when disambiguating between similar extractions.

However, when a LLM agent is granted access to the database, two common pitfalls emerge.
First, the agent may only inspect the initial subset of rows (e.g., the first 10) and attempt to generalize issues from this limited sample. 
This approach fails because each extracted value may originate from a different source, and naïve generalization can lead to systematic errors (e.g., always selecting the maximum revenue, which may not be appropriate in every case).
Second, the agent may attempt to load the entire database context at once using queries like \texttt{SELECT * FROM table} which is neither scalable nor effective, as it quickly exhausts the available context window.

\begin{figure}[!t]
    \centering
    \includegraphics[width=\linewidth]{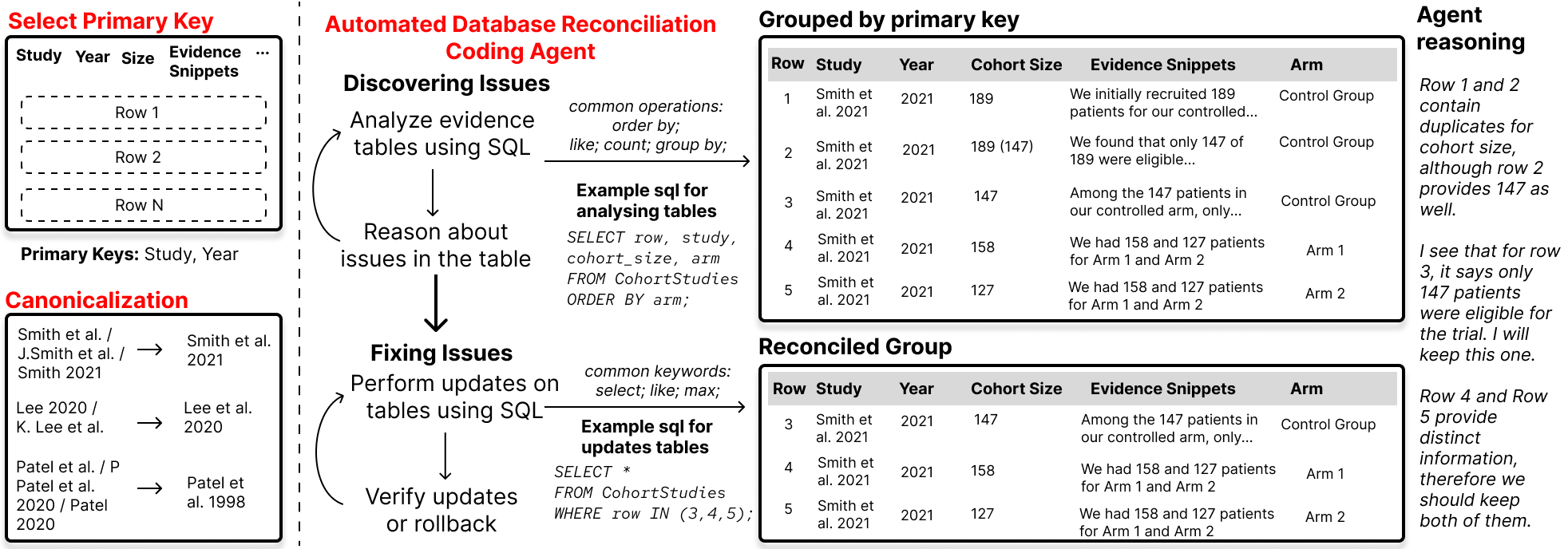}
    \caption{
The agent first selects the primary key and canonicalizes its values to group rows referring to the same study. It then analyzes the grouped records using SQL and evidence snippets to identify duplicates and resolve inconsistencies. Updates are verified and either retained or rolled back.
}
    \label{fig:reconciliation}
    \vspace{-4mm}
\end{figure}

Since the agent must consider each value alongside its provenance and extraction rationale, there is a need for strategies that allow the LLM to process the data in manageable segments. 
This requires grouping of data by bringing related information together, enabling effective discovery of reconciliation tasks without requiring comparisons across all rows simultaneously (Figure~\ref{fig:reconciliation}).
\iffalse
Why is pairwise enough?  We don't need this. 
Let there be $n$ evidence records in the raw evidence table $\mathcal{Z}_{q}$. 
A direct reconciliation procedure would compare all pairs $(z_i,z_j)$ where $1 \leq i < j \leq n$, requiring $O(n^2)$ comparisons. 
This is infeasible when the evidence table is large. 
\fi

\paragraph{Constructing evidence groups.}

To inspect related records together, the agent identifies a set of grouping variables and determines canonical (standardized) representations for their values. For each reconciliation task, these grouping variables are chosen based on the original question, and the evidence table. For example, in financial analysis, grouping might be done by company and fiscal year to bring together all records relevant to an annual report. In a clinical review, records might be grouped according to intervention, population, outcome, and time point, so that all instances referring to the same clinical comparison can be analyzed jointly.

For each record, the agent generates a grouping key by converting each relevant variable to its canonical form, ensuring that different surface expressions (such as name variations or synonyms) are treated consistently. Records that share the same canonical grouping key are collected into one group, allowing the agent to efficiently reconcile related data points within each group.

\iffalse
To inspect related records together, the agent derives grouping variables and canonical representations of their values. Let
$K_q = \{k_1,\ldots,k_s\} \subseteq \mathcal{V}_{q, \mathcal{M}}$
be the variables used for a reconciliation task. For example, company and fiscal year may identify records relevant to an annual financial question, whereas intervention, population, outcome, and time point may define a clinical comparison. If $z[k_j]$ is the value of variable $k_j$ in record $z$, the grouping key is
$
    \kappa_q(z) =
    \bigl(c_{k_1,q}(z[k_1]),\ldots,c_{k_s,q}(z[k_s])\bigr),
$
where $c_{k_j,q}$ maps equivalent surface forms to a canonical value. A group associated with key $a$ is
$
    g_a = \{z\in I : \kappa_q(z)=a\}.
$

Canonicalization brings related evidence together despite aliases or formatting differences. The agent uses source context to distinguish similar representations from values whose differences are meaningful.
\fi

\paragraph{Diagnosing records from source context.}
Within a group, the agent reviews the extracted values along with their associated source spans and rationales. 
We ask the model to analyze and diagnose one of the four reconciliation issues. 
First, there may be duplicate records that represent the same underlying finding; in such cases, these can be consolidated while retaining all relevant supporting sources.
Second, the group may include complementary records that address different parts of the question, and these can be combined or aggregated for a more comprehensive answer.
Third, there may be conflicting records that present genuinely different values for the same fields, requiring the agent to determine which value should be retained.
Finally, apparent conflicts may arise due to varying assumptions or levels of granularity; in these cases, all records should be preserved.

\paragraph{Writing Custom Reconciliation Code.}
The agent generates SQL code to address the issues identified during the diagnosis stage. We observed that, when prompted to fix problems by providing only a description of the issue to the LLM, its reasoning often diverged from the actual problem. For instance, the agent might select the earliest date to distinguish between two records without considering provenance or rationale, or it might choose computed values over those explicitly stated.

To mitigate these issues, we provide the coding agent with guidelines for each issue type (see Appendix F.1.3). This guidance enables accurate reconciliation across a range of scenarios, such as resolving conflicting field values, eliminating duplicates, or merging complementary records.

As illustrated in Figure~\ref{fig:reconciliation}, the agent correctly identifies the first two rows as duplicates, despite differing values. It further recognizes a conflict between the third row and the first two (e.g., cohort size: 147 vs. 189), and selects the correct value based on supporting evidence. Additionally, it retains rows four and five as they are associated with different study arms.

The SQL produced by the coding agent does not always precisely reflect its intent. 
The generated updates may also have unintended side effects. To address this, we introduce a verification agent that evaluates whether the database updates align with the reconciliation agent’s stated intent.

\subsection{Synthesizing Answers and Reusing Evidence}
Once the reconciled evidence table has been constructed, \system generates the response from the reconciled evidence table.
Importantly, the reconciled table may still be very large, so the agent uses evidence variables to retrieve a evidence context for final synthesis.
The final answer is then generated by a series of inspections of the selected evidence context, with citations to the supporting spans.
More importantly, this evidence table can be used for answering further questions.

\section{Evaluation on Benchmarks Beyond LLM Context Limits}
\label{sec:beyong_limits}

We evaluate \system's ability to synthesize evidences from documents.
In all tasks, the system is given a research question and a document corpus with the goal of answering the question.

We first evaluate on \textbf{BrowseComp+}~\cite{chen2025browsecomp}, a deep-research benchmark which tests whether systems can identify and synthesize evidence distributed across multiple documents.
To test the follow-up analysis of \system, we introduce \textbf{WikiCeleb100 (WC)}, a new corpus of 100 Wikipedia articles with a research question and 21 related follow-up questions, designed to test reusable evidence-table construction and subgroup comparison. 
WC comprises Wikipedia articles for the 100 most-viewed celebrity pages from November 2025 to January 2026 (\emph{3.9M tokens}); the research question is ``Who debuted at the youngest age across the following industries: Music, Film, Content Creation, and Other?''. It requires extracting each celebrity's date of birth, debut date, and industry from across all 100 articles. 
As a larger follow-up analysis benchmark, we present \textbf{FinQ100 (FQ)}. FQ comprises the most recent 10-Q filings of 100 randomly selected SEC-listed companies (\emph{36M tokens}) with 25 questions.
The research question is ``Which company has the lowest long-term borrowing?''. This question is harder because many filings do not state borrowing explicitly, requiring inference from context. Full descriptions of the benchmarks, construction protocols, and metrics are in Appendix~\ref{app:benchmark_descriptions}.

\textbf{Metrics.} All result tables report the accuracy of the final answer against the gold answer. BrowseComp+, Loong and Oolong use their official metrics. For FinanceBench, WikiCeleb100 and FinQ100, a GPT-4.1 judge compares it with the gold answer, giving partial credit on WikiCeleb100 and FinQ100 as in Oolong. It agrees with human labels at Cohen's $\kappa = 0.758$ (Appendix~\ref{app:eval_reliability}).

\paragraph{Baselines.}
We compare against retrieval, base-LM, and agentic baselines; full implementation details are in Appendix~\ref{app:baseline_details}.
We use \textbf{RAG}~\citep{lewis2020retrieval}, \textbf{LongRAG}~\citep{jiang2024longrag}\textbf{, GraphRAG}~\citep{edge2024graphrag} with Qwen3-4B for indexing and GPT-4.1 as the reader.
We use \textbf{DocETL}~\citep{docetl} and use the same evidence variables as ours with GPT-4.1.
We use \textbf{RLM}~\citep{zhang2025recursivelanguagemodels}, following the original RLM paper, we run RLM with GPT-5 / GPT-5-mini. 
For RLM, we give all questions of WikiCeleb100 and FinQ100 at once, so that it can build a shared state in its Python REPL and reuse it across questions. 
GraphRAG builds its graph index once, before seeing any question, and reuses it for every question.
For \system{}, GPT-4.1-mini handles evidence-variable-aware extraction and GPT-4.1 handles evidence-variable generation, reconciliation and final synthesis. 
We also report using Qwen3.5-122B-A10B.

\definecolor{ScoreGreen}{HTML}{2E7D32}
\newcommand{\score}[2]{\cellcolor{ScoreGreen!#1}#2}

\begin{figure}[t]
\centering

% ---------------- Left: table ----------------
\begin{minipage}[t]{0.66\linewidth}
\vspace{0pt}
\centering
\captionof{table}{Performance comparison on BrowseComp+ (BC+), WikiCeleb100 (WC), and FinQ100 (FQ). \$\$ means cost prohibitive.}
\label{tab:results}

\small
\resizebox{\linewidth}{!}{%
\begin{tabular}{llrrr}
\toprule
\textbf{Method} & \textbf{Worker / Reasoning LM} & \textbf{BC+} & \textbf{WC} & \textbf{FQ} \\
\midrule
RAG      & -- / GPT-4.1
    & \score{2}{20.67}  & \score{2}{31.41}  & \score{2}{5.00} \\
LongRAG  & -- / GPT-4.1
    & \score{5}{28.67}  & \score{11}{43.20} & \score{19}{28.87} \\
GraphRAG & -- / GPT-4.1
    & \$\$              & \score{16}{48.59} & 08.00\footnote{\label{fn:ten-docs}on 10 documents. For the same documents, \system{} gets a score of 64.00} \\
GraphRAG & -- / Qwen
    & \$\$              & \score{16}{48.59} & 07.00 \\
\midrule
DocETL   & GPT-4.1 / GPT-4.1
    & 70.00           & \score{22}{54.26} & 16.00\textsuperscript{\ref{fn:ten-docs}} \\
RLM      & GPT-5-mini / GPT-5
    & \underline{\score{39}{80.00}} & \score{27}{59.80} & \score{19}{28.00} \\
\midrule
\textbf{SLIDERS} & GPT-4.1-mini / GPT-4.1
    & \textbf{\score{45}{90.00}} & \textbf{\score{45}{78.91}} & \underline{\score{41}{55.22}} \\
\textbf{SLIDERS} & Qwen3.5 122B / Qwen3.5 122B
    & --                & \underline{\score{43}{76.92}} & \textbf{\score{45}{60.18}} \\
\bottomrule
\end{tabular}%
}
\end{minipage}
\hfill
% ---------------- Right: plots ----------------
\begin{minipage}[t]{0.31\linewidth}
\vspace{0pt}
\centering

\begin{tikzpicture}
\begin{groupplot}[
    group style={
        group size=1 by 2,
        vertical sep=1.15cm,
    },
    width=\linewidth,
    height=2.4cm,
    ymin=65, ymax=100,
    grid=major,
    grid style={dashed, gray!30},
    tick label style={font=\scriptsize},
    label style={font=\scriptsize},
    title style={font=\scriptsize, yshift=-1.5ex},
    legend columns=2,
    legend style={
        draw=none,
        font=\scriptsize,
        /tikz/every even column/.append style={column sep=0.3cm},
    },
]

% Top plot: token length
\nextgroupplot[
    title={\textbf{BrowseComp+: token length}},
    xlabel={Token length},
    ylabel={Accuracy (\%)},
    xmin=5.5, xmax=11,
    xtick={6, 7, 8, 9, 10, 11},
    xticklabels={6M, 7M, 8M, 9M, 10M, 11M},
    legend to name=sharedlegend,
]

\addplot[blue, thick, mark=o, mark size=1.5pt] coordinates {
    (6.15, 88.9)
    (7.17, 89.8)
    (8.50, 91.7)
    (10.28, 90.0)
};
\addlegendentry{\system{}}

\addplot[red, thick, mark=square, mark size=1.5pt] coordinates {
    (6.15, 72.2)
    (7.17, 82.7)
    (8.50, 75.0)
    (10.28, 70.0)
};
\addlegendentry{RLM}

% Bottom plot: evidence-bearing documents
\nextgroupplot[
    title={\textbf{BrowseComp+: evidence count}},
    xlabel={\# evidence-bearing documents},
    ylabel={Accuracy (\%)},
    xmin=2, xmax=11,
    xtick={2.6, 4.9, 7.9, 10.3},
    xticklabels={1--3, 4--6, 7--9, 10+},
]

\addplot[blue, thick, mark=o, mark size=1.5pt] coordinates {
    (2.6, 76.2)
    (4.9, 91.2)
    (7.9, 91.5)
    (10.3, 100.0)
};

\addplot[red, thick, mark=square, mark size=1.5pt] coordinates {
    (2.6, 76.2)
    (4.9, 72.1)
    (7.9, 85.1)
    (10.3, 71.4)
};

\end{groupplot}

% Shared legend below both rows
\node at ($(group c1r2.south) - (0,0.95cm)$) {
    \pgfplotslegendfromname{sharedlegend}
};

\end{tikzpicture}
\vspace{-4mm}
\captionof{figure}{Acc. vs token length and \# of relevant docs.}
\label{fig:browsecomp_token_and_evidence}

\vspace{-4mm}
\end{minipage}

\end{figure}

\textbf{Observation 1: SLIDERS improves evidence synthesis as evidence becomes longer and more distributed.}
BrowseComp+ evaluates whether a system can identify and synthesize evidence distributed across multiple documents for a single deep-research question. 
On this benchmark, \system{} achieves 90.00, outperforming the strongest non-\system{} baseline, RLM, by 10 points, as shown in Table~\ref{tab:results}. 
Retrieval-based methods such as RAG and LongRAG perform substantially worse, suggesting that answering these questions requires more than selecting a small set of relevant passages. 
On average, RAG and LongRAG only retrieved 10.4\% and 12.5\% of relevant documents out of all the relevant documents in the corpus. 
Moreover, the RAG and LongRAG retrievers return zero evidence docs
for 57\% and  49\% questions, respectively. 
In contrast, \system{} extracts the relevant evidence variables, groups related records, reconciles overlapping or conflicting evidence, and then answers from the resulting evidence table.

Figure~\ref{fig:browsecomp_token_and_evidence} shows that this advantage persists as the evidence becomes harder to synthesize. 
As token length increases from roughly 6M to 11M tokens, \system{} remains near 90\% accuracy, while RLM varies more substantially and ends lower. 
The gap is even clearer as the number of evidence-bearing documents increases. 
Even in the case when questions require evidence from 10 or more documents, \system{} achieves 100\% accuracy, while RLM drops to around 70\%. 
This suggests that \system{} remains effective when the answer depends on evidence distributed across many sources: the reconciled evidence table gives the answer generator a structured view of the relevant evidence, rather than forcing it to synthesize from scattered passages or independently processed chunks.

\begin{wrapfigure}{r}{0.30\textwidth}
\vspace{-1em}
\centering
\begin{tikzpicture}
\begin{axis}[
    width=\linewidth,
    height=2.5cm,
    xlabel={Number of documents},
    ylabel={Accuracy (\%)},
    xmin=10, xmax=100,
    ymin=0, ymax=100,
    xtick={20,40,60,80,100},
    grid=major,
    grid style={dashed, gray!30},
    tick label style={font=\scriptsize},
    label style={font=\scriptsize},
    title style={font=\scriptsize, yshift=-1.5ex},
    title={\textbf{SLIDERS vs. corpus size}},
    legend style={
        draw=none,
        font=\scriptsize,
        at={(0.5,-0.8)},
        anchor=north,
        legend columns=2,
    },
]

% WikiCeleb100
\addplot[cyan, thick, mark=o, mark size=1.5pt] coordinates {
    (10,70.6)
    (20,75.0)
    (30,67.4)
    (40,68.8)
    (50,69.6)
    (60,75.6)
    (70,71.4)
    (80,76.66)
    (90,71.8)
    (100,78.9)
};
\addlegendentry{WikiCeleb100}

% FinQ100
\addplot[orange] coordinates {
    (10,65.1)
    (20,54.9)
    (30,56.3)
    (40,53.9)
    (50,68.9)
    (60,58.9)
    (70,61.6)
    (80,57.7)
    (90,54.5)
    (100,55.2)
};
\addlegendentry{FinQ100}

\end{axis}
\end{tikzpicture}

\caption{SLIDERS performance as the number of documents increases on WikiCeleb100 and FinQ100.}
\label{fig:wc-finq-wrap}
\vspace{-1em}
\end{wrapfigure}

\textbf{Observation 2: Reusable evidence tables enable follow-up analysis over large document corpora. }
WikiCeleb100 and FinQ100 test the additional capability of reusing evidence collected to answer related follow-up questions. 
For these benchmarks, \system{} constructs a reconciled evidence table to answer the original research question, and then answers follow-up questions using the same table. 
RLM and GraphRAG also reuse state across questions, through RLM's shared REPL and GraphRAG's prebuilt graph, but no baseline maintains a reconciled evidence table. 

As shown in Table~\ref{tab:results}, on WikiCeleb100, \system{} reaches 78.91, compared to 59.80 for the strongest competing baseline, RLM.
This benchmark requires extracting dates of birth, debut dates, and industries across 100 Wikipedia articles, then reusing this evidence for comparisons across industries. 
The gain suggests that the evidence table is not only useful for the initial question, but also provides a reusable substrate for follow-up analysis.
FinQ100 is substantially larger, with 36M tokens of SEC filings, and stresses the same workflow at a much larger scale. 
Among baselines that complete, the best score is 28.87, whereas \system{} reaches 60.18 with Qwen3.5-122B-A10B. 
This gap indicates that follow-up analysis over large corpora requires more than rerunning retrieval for each question.

\begin{wraptable}{r}{0.35\linewidth}
    \centering
    \scriptsize
    \caption{Ablation results on BrowseComp+,WikiCeleb100,  and FinQ100. We use a separate test set for these benchmarks.}
    \begin{tabular}{lrrr}
    \toprule
    \textbf{Variants} & \textbf{BC+} & \textbf{WC} & \textbf{FQ} \\
    \midrule
    \textbf{SLIDERS}
        & \textbf{90.00} & \textbf{78.91} & \textbf{55.22} \\
    \qquad Direct table
        & 87.33 & -- & -- \\
    \qquad No Recon.
        & 87.33 & 60.50 & 35.81 \\
    \bottomrule
    \end{tabular}
    \label{tab:ablation_deepresearch}
    \vspace{-4mm}
\end{wraptable}
Finally, \system{} can also be instantiated with an open-weight reasoning model. 
Using Qwen3.5-122B-A10B for all components, \system{} outperforms all GPT-4.1-based baselines on both WikiCeleb100 and FinQ100.
We do not have the performance results on BrowseComp+ due to resource constraints.
On FinQ100, it also improves over the GPT-4.1-based \system{} variant, suggesting that the evidence-table and reconciliation framework is not tied to a specific proprietary model.

\textbf{Observation 3: Reconciliation matters most when the evidence table is large.}
As shown in Table~\ref{tab:ablation_deepresearch}, on BrowseComp+ the evidence table fits in the LLM context, allowing the \emph{Direct table} variant to pass it directly to the answer generator. It reaches 87.33, only 2.67 points below the full system, showing that evidence-table construction alone is strong when the table is small enough to consume in full. This strategy does not scale to WikiCeleb100 and FinQ100, where the tables are too large. In these settings, reconciliation is critical: removing it drops performance from 78.91 to 60.50 on WikiCeleb100 and from 55.22 to 35.81 on FinQ100. These drops show that chunk-level extraction is insufficient for large, distributed evidence; the system must group related records, reconcile overlapping or conflicting evidence, and produce an answer-ready view before generation.

\section{Evaluation on Benchmarks Within LLM Context Limits}
After demonstrating \system{}’s advantages on large-scale datasets that exceed LLM context windows, we next assess its performance on benchmarks where the entire evidence set fits within those limits.
First, we evaluate on MedEvidence~\citep{polzak2026medevidence}, a systematic-review benchmark where questions are drawn from the conclusions of 100 systematic reviews. 
Here, documents are small (upto 62k tokens), so this task does not fully leverage \system{}’s large-scale extraction and reconciliation strengths. Nevertheless, \system{} achieves a score of 71.5, outperforming one-shot GPT-5, which scores 66.2.

To broaden the comparison with techniques constrained by LLM context windows, we also evaluate \system{} on three additional “long-context QA” benchmarks:
\textbf{FinanceBench (FB)}\citep{Islam2023FinanceBench}, a single-document financial QA benchmark with 150 questions over public filings (up to 363K tokens);
\textbf{Loong}\citep{Wang2024Loong}, a multi-document QA benchmark requiring every provided document per instance ($\sim$11 documents, up to 300K tokens, across finance, law, and academic papers, in both English and Chinese);
and \textbf{Oolong}~\citep{bertsch2025oolong} (Oolong-Synth, 256K tokens), an aggregation benchmark requiring models to classify individual data points and aggregate them into a global answer.

\textbf{Baselines. } 
In addition to baselines in Sec.~\ref{sec:beyong_limits}, we also evaluate:
\textbf{Base LM}: GPT-4.1 and Qwen3.5-122B-A10B~\citep{qwen2026qwen35} (1M-token context),
\textbf{CoA}~\citep{zhang2024coa}: sequential chunk summarization with running context (GPT-4.1-mini / GPT-4.1), and 
\textbf{RLM} with GPT 4.1 and GPT 4.1-mini as LLM backbone.

\begin{table}[t]
\small
\centering
\caption{Performance comparison across FinanceBench (FB), Loong, Oolong, and Average}
\begin{tabular}{llrrrr}
\toprule
\textbf{Method} & \textbf{Worker / Reasoning LM} & \textbf{FB} & \textbf{Loong} & \textbf{Oolong} & \textbf{Avg} \\
\midrule
RAG            & -- / GPT-4.1
    & \score{0}{62.67}  & \score{0}{54.35}  & \score{0}{11.32}  & \score{0}{42.78} \\
LongRAG        & -- / GPT-4.1
    & \score{16}{72.00} & \score{9}{59.10}  & \score{8}{22.00}  & \score{11}{51.03} \\
GraphRAG       & -- / GPT-4.1
    & \score{21}{75.33} & \score{13}{61.28} & \score{8}{22.00}  & \score{13}{52.87} \\
    \midrule
BaseLM         & -- / GPT-4.1
    & \score{32}{82.00} & \underline{\score{42}{76.74}} & \score{27}{45.56} & \score{33}{68.10} \\
BaseLM         & -- / Qwen3.5 122B
    & \underline{\score{37}{84.67}} & \score{38}{74.78} & \score{11}{24.89} & \score{24}{61.45} \\
    \midrule
DocETL         & GPT-4.1 / GPT-4.1
    & \score{1}{63.33}  & \score{38}{75.03} & \score{30}{49.00} & \score{25}{62.45} \\
Chain of Agent & GPT-4.1-mini / GPT-4.1
    & \score{14}{71.30} & \score{0}{54.46}  & \score{5}{17.11}  & \score{6}{47.62} \\
RLM            & GPT-4.1-mini / GPT-4.1
    & \score{16}{72.00} & \score{11}{60.13} & \score{32}{52.00} & \score{24}{61.38} \\
RLM            & GPT-5-mini / GPT-5
    & \score{21}{75.33} & \score{34}{72.64} & \score{32}{51.42} & \score{31}{66.46} \\
    \midrule
\textbf{SLIDERS}        & GPT-4.1-mini / GPT-4.1
    & \textbf{\score{45}{89.67}} & \textbf{\score{45}{78.57}} & \underline{\score{42}{64.67}} & \textbf{\score{45}{77.64}} \\
\textbf{SLIDERS}        & Qwen3.5 122B / Qwen3.5 122B
    & \score{32}{82.10} & \score{40}{75.70} & \textbf{\score{45}{68.00}} & \underline{\score{42}{75.27}} \\
\bottomrule
\end{tabular}
\label{tab:results2}
\end{table}

\begin{figure*}[t]
\centering
\small

\begin{minipage}[t]{0.43\textwidth}
\centering
\textbf{(a) Ablation results.}

\vspace{0.5em}
\scriptsize
\begin{tabular}{lrrr}
\toprule
\textbf{Variant} & \textbf{FB} & \textbf{Loong} & \textbf{Oolong} \\
\midrule
\textbf{SLIDERS}
    & \textbf{80.00} & \underline{84.37} & \textbf{64.67} \\
\qquad Direct Table
    & 70.00 & \textbf{84.45} & 58.62 \\
\qquad No Reconciliation
    & \underline{76.70} & 82.84 & \underline{62.42} \\
\qquad No chunk extraction
    & 70.00 & 79.72 & 40.00 \\
\bottomrule
\end{tabular}
\end{minipage}
\hfill
\begin{minipage}[t]{0.53\textwidth}
\centering
\textbf{(b) Accuracy vs.\ token length.}

\vspace{0.5em}
\begin{tikzpicture}
\begin{groupplot}[
    group style={
        group size=2 by 1,
        horizontal sep=0.55cm,
        xlabels at=edge bottom,
        ylabels at=edge left,
        xticklabels at=edge bottom,
    },
    width=3.7cm,
    height=2.35cm,
    xlabel={Token Length},
    ylabel={Accuracy (\%)},
    grid=major,
    grid style={dashed, gray!30},
    tick label style={font=\scriptsize},
    label style={font=\scriptsize},
    title style={font=\scriptsize, yshift=-1.5ex},
]

% Subplot 1: Loong
\nextgroupplot[
    title={\textbf{Loong}},
    legend to name=sharedlegend,
    legend columns=4,
    legend style={
        draw=none,
        font=\scriptsize,
        /tikz/every even column/.append style={column sep=0.2cm},
    },
    ymin=40, ymax=85,
    xmin=50, xmax=300,
    xtick={64, 128, 256},
    xticklabels={64k, 128k, 256k},
]
\addplot[blue, thick, mark=o, mark size=1.4pt] coordinates {
    (50,78.4) (100,83.1) (200,73.1) (300,72.1)
};
\addplot[red, thick, mark=square, mark size=1.4pt] coordinates {
    (50,73) (100,80.1) (200,65.7) (300,49.2)
};
\addplot[green!60!black, thick, mark=triangle, mark size=1.4pt] coordinates {
    (50,77.6) (100,82.5) (200,69.4) (300,63.8)
};
\legend{\system{}, RLM, GPT-4.1}

% Subplot 2: FinanceBench
\nextgroupplot[
    title={\textbf{FinanceBench}},
    ymin=50, ymax=100,
    xmin=10, xmax=300,
    xtick={64, 128, 256},
    xticklabels={64k, 128k, 256k},
]
\addplot[blue, thick, mark=o, mark size=1.4pt] coordinates {
    (10,62.5) (50,89.5) (100,92.1) (200,92.5) (300,83.3)
};
\addplot[red, thick, mark=square, mark size=1.4pt] coordinates {
    (10,50) (50,68.4) (100,89.5) (200,73.1) (300,72.2)
};
\addplot[green!60!black, thick, mark=triangle, mark size=1.4pt] coordinates {
    (10,62.5) (50,89.5) (100,89.5) (200,80.6) (300,72.2)
};

\end{groupplot}

\node at ($(group c1r1.south)!0.5!(group c2r1.south) - (0,0.95cm)$) {
    \pgfplotslegendfromname{sharedlegend}
};
\end{tikzpicture}
\end{minipage}
\caption{Ablation on validation set and token-length analysis on long-context benchmarks. 
}
\vspace{-4mm}
\label{fig:ablation_token_length}
\end{figure*}

\textbf{Observation 1: Using Evidence Table also improves long context QA accuracy, even when the corpus already fits in the model's context window.}
As shown in Table~\ref{tab:results2}, \system{} achieves the best average performance on the three within-window benchmarks, outperforming the strongest baseline (GPT-4.1) by 9.54 points on average.
The largest within-window gain occurs on Oolong, where every question requires aggregating many local reasonings into a global answer -- the regime in which each excerpt is important or they are widely scattered through the corpus.
That the gain appears even when the input fits in the base model's window indicates that accuracy is bounded by whether the retained context supports targeted retrieval, not by raw context length.
\system{} with Qwen3.5-122B-A10B is close to its GPT-4.1 version and beats base LMs on Loong and Oolong.

\textbf{Observation 2: Each stage of SLIDERS contributes to robust evidence synthesis.}
Figure~\ref{fig:ablation_token_length}(a) shows that removing major components generally reduces performance across FinanceBench, Loong, and Oolong. 
The full system achieves the best result on FinanceBench and Oolong, with especially large gains on Oolong: removing chunk-level extraction drops performance from 64.67 to 40.00, showing that systematic evidence extraction is necessary when relevant information is distributed across long inputs. 
Removing reconciliation also lowers performance on all three benchmarks, indicating that extracted evidence must still be organized and resolved before answer generation. 
The only exception is Loong, where the Direct Table variant slightly exceeds the full system by 0.08 points. The extracted table was already small, additional reconciliation may provide less benefit. 

\textbf{Observation 3: SLIDERS is more stable than baselines as token length increases.}
Figure~\ref{fig:ablation_token_length}(b) compares accuracy across token lengths on Loong and FinanceBench. 
On Loong, SLIDERS remains above 70\% even at the longest inputs, while RLM drops sharply as the context grows, reaching below 50\% at the largest token length. 
A similar pattern appears on FinanceBench. 
SLIDERS maintains high accuracy across longer contexts and remains strongest at the largest input size. 
These results suggest that SLIDERS does not depend on directly reasoning over the entire long input.

\section{Analysis and Discussions}
\paragraph{How much reconciliation is needed?}
\label{sec:context_audit}
Here we evaluate the effect of the reconciliation agent in removing duplicates and inconsistencies, and consolidating incomplete records after extraction. As shown in Table~\ref{tab:context_funnel} (Appendix), 
reconciliation yields the largest compression on WikiCeleb100 and FinQ100, with reductions of $1.98\times$ and $4.51\times$, respectively, consistent with the measure of improvement due to reconciliation reported in 
Table~\ref{tab:ablation_deepresearch}. 

\textbf{What kind of reconciliation is needed?}
Figure~\ref{fig:operations} reports the breakdown of reconciliation issues identified by \system{}.
Records requiring \emph{deduplication}, \emph{consolidation}, and \emph{resolving conflicts}. 
Financial and legal corpora are dominated by conflict resolution, while WikiCeleb100 requires more consolidation because evidence about the same entity is distributed across sections.

\textbf{How good is automated reconciliation?}
We manually compile the reconciled evidence table for WikiCeleb100 and FinQ100 (Appendix~\ref{app:recon_correctness}) and use it to compare reconciliatin accuracy of \system{}.Reconciliation recovers 92.5\% and 90.1\% of the gold rows, with a structural F1 of 0.91 on WikiCeleb100. 
On FinQ100, precision is 0.49 because the agent keeps multiple rows for 16 entities it failed to canonicalize.
When the correct value is among the extracted candidates, the agent selects it 80\% (WikiCeleb100) and 81\% (FinQ100) of the time, and most wrong values occur because the correct record was not retained.

\textbf{Failure Analysis.}
In Fig.~\ref{fig:benchmark-breakdown}(a), on the Chinese classification subset, documents are usually under 16K tokens and the task is label prediction rather than synthesis over distributed evidence, so the benefits of using evidence variables, extraction, and reconciliation are limited. 
In Fig.~\ref{fig:benchmark-breakdown}(b), we observe that performance is lower for questions that require deep financial domain understanding.
Finally, exact counting and numerical questions remain challenging because they require not only preserving evidence, but also correctly classifying borderline cases (Fig.~\ref{fig:benchmark-breakdown}(c)), deduplicating records, normalizing values, and aggregating them. 
This limitation is especially visible in FinQ100, where many failures require precise numerical reasoning over financial evidence. 
On BrowseComp+, 8 of the 15 failures arise from incorrect reasoning over the final table, while 5 occur because \system{} does did not synthesize the correct evidence table.

\section{Conclusion}

We introduced SLIDERS, a scalable, database-centric agent for automated evidence extraction and synthesis across large document collections. By combining context-aware chunk-based extraction, autonomous database reconciliation, and source-grounded evidence management, SLIDERS addresses key limitations of existing LLM-based approaches while supporting flexible follow-up queries. Experiments across multiple benchmarks demonstrate substantial improvements over strong baselines, including gains of 19 and 27 percentage points on WikiCeleb100 and FinQ100, respectively. These results highlight the potential of database-centric agents to enable more accurate, scalable, and flexible evidence synthesis for systematic reviews.

\begin{ack}
We thank Shicheng Liu, Nikil Selvam, Aryaman Arora, Yanzhe Zhang, Nishna Reddy Aleti, Cyrus Zhou, Adit Negi, Vedant Singh, Tony Liu, Jiuding Sun, Sina Semnani and the members of the Stanford OVAL and Stanford NLP for helpful discussion about the project and comments on the manuscript. This work is supported in part by the Verdant Foundation, Hasso Plattner Institute, Ita\'u Unibanco, BMO Financial Group, and the Stanford Human-Centered Artificial Intelligence (HAI) Institute. We acknowledge the National Artificial Intelligence Research Resource (NAIRR) Pilot and Microsoft Azure for contributing to the results in this work.
\end{ack}

\bibliographystyle{unsrtnat}
\bibliography{ref}

@inproceedings{he2024never,
    title = "Never Lost in the Middle: Mastering Long-Context Question Answering with Position-Agnostic Decompositional Training",
    author = "He, Junqing  and
      Pan, Kunhao  and
      Dong, Xiaoqun  and
      Song, Zhuoyang  and
      Liu, Yibo  and
      Sun, Qianguo  and
      Liang, Yuxin  and
      Wang, Hao  and
      Zhang, Enming  and
      Zhang, Jiaxing",
    editor = "Ku, Lun-Wei  and
      Martins, Andre  and
      Srikumar, Vivek",
    booktitle = "Proceedings of the 62nd Annual Meeting of the Association for Computational Linguistics (Volume 1: Long Papers)",
    month = aug,
    year = "2024",
    address = "Bangkok, Thailand",
    publisher = "Association for Computational Linguistics",
    url = "https://aclanthology.org/2024.acl-long.736/",
    doi = "10.18653/v1/2024.acl-long.736",
    pages = "13628--13642"
}

@inproceedings{liu2025mdcure,
    title = "{MDC}ure: A Scalable Pipeline for Multi-Document Instruction-Following",
    author = "Liu, Gabrielle Kaili-May  and
      Shi, Bowen  and
      Caciularu, Avi  and
      Szpektor, Idan  and
      Cohan, Arman",
    editor = "Che, Wanxiang  and
      Nabende, Joyce  and
      Shutova, Ekaterina  and
      Pilehvar, Mohammad Taher",
    booktitle = "Proceedings of the 63rd Annual Meeting of the Association for Computational Linguistics (Volume 1: Long Papers)",
    month = jul,
    year = "2025",
    address = "Vienna, Austria",
    publisher = "Association for Computational Linguistics",
    url = "https://aclanthology.org/2025.acl-long.1418/",
    doi = "10.18653/v1/2025.acl-long.1418",
    pages = "29258--29296",
    ISBN = "979-8-89176-251-0"
}

@article{eyuboglu2025cartridges,
  title={Cartridges: Lightweight and general-purpose long context representations via self-study},
  author={Eyuboglu, Sabri and Ehrlich, Ryan and Arora, Simran and Guha, Neel and Zinsley, Dylan and Liu, Emily and Tennien, Will and Rudra, Atri and Zou, James and Mirhoseini, Azalia and others},
  journal={arXiv preprint arXiv:2506.06266},
  year={2025}
}

@article{yu2025memagent,
  title={MemAgent: Reshaping Long-Context LLM with Multi-Conv RL-based Memory Agent},
  author={Yu, Hongli and Chen, Tinghong and Feng, Jiangtao and Chen, Jiangjie and Dai, Weinan and Yu, Qiying and Zhang, Ya-Qin and Ma, Wei-Ying and Liu, Jingjing and Wang, Mingxuan and others},
  journal={arXiv preprint arXiv:2507.02259},
  year={2025}
}

@inproceedings{zhao2024longagent,
    title = "{LONGAGENT}: Achieving Question Answering for 128k-Token-Long Documents through Multi-Agent Collaboration",
    author = "Zhao, Jun  and
      Zu, Can  and
      Hao, Xu  and
      Lu, Yi  and
      He, Wei  and
      Ding, Yiwen  and
      Gui, Tao  and
      Zhang, Qi  and
      Huang, Xuanjing",
    editor = "Al-Onaizan, Yaser  and
      Bansal, Mohit  and
      Chen, Yun-Nung",
    booktitle = "Proceedings of the 2024 Conference on Empirical Methods in Natural Language Processing",
    month = nov,
    year = "2024",
    address = "Miami, Florida, USA",
    publisher = "Association for Computational Linguistics",
    url = "https://aclanthology.org/2024.emnlp-main.912/",
    doi = "10.18653/v1/2024.emnlp-main.912",
    pages = "16310--16324"
}

@article{zhang2024coa,
  title={Chain of agents: Large language models collaborating on long-context tasks},
  author={Zhang, Yusen and Sun, Ruoxi and Chen, Yanfei and Pfister, Tomas and Zhang, Rui and Arik, Sercan},
  journal={Advances in Neural Information Processing Systems},
  volume={37},
  pages={132208--132237},
  year={2024}
}

@inproceedings{li2024graphreader,
    title = "{G}raph{R}eader: Building Graph-based Agent to Enhance Long-Context Abilities of Large Language Models",
    author = "Li, Shilong  and
      He, Yancheng  and
      Guo, Hangyu  and
      Bu, Xingyuan  and
      Bai, Ge  and
      Liu, Jie  and
      Liu, Jiaheng  and
      Qu, Xingwei  and
      Li, Yangguang  and
      Ouyang, Wanli  and
      Su, Wenbo  and
      Zheng, Bo",
    editor = "Al-Onaizan, Yaser  and
      Bansal, Mohit  and
      Chen, Yun-Nung",
    booktitle = "Findings of the Association for Computational Linguistics: EMNLP 2024",
    month = nov,
    year = "2024",
    address = "Miami, Florida, USA",
    publisher = "Association for Computational Linguistics",
    url = "https://aclanthology.org/2024.findings-emnlp.746/",
    doi = "10.18653/v1/2024.findings-emnlp.746",
    pages = "12758--12786"
}

@inproceedings{narayan2025minions,
  title={Cost-efficient collaboration between on-device and cloud language models},
  author={Narayan, Avanika and Biderman, Dan and Eyuboglu, Sabri and May, Avner and Linderman, Scott and Zou, James and Re, Christopher},
  booktitle={Forty-second International Conference on Machine Learning},
  year={2025}
}

@inproceedings{zhou2025llmmapreduce,
    title = "{LLM}$\times${M}ap{R}educe: Simplified Long-Sequence Processing using Large Language Models",
    author = "Zhou, Zihan  and
      Li, Chong  and
      Chen, Xinyi  and
      Wang, Shuo  and
      Chao, Yu  and
      Li, Zhili  and
      Wang, Haoyu  and
      Shi, Qi  and
      Tan, Zhixing  and
      Han, Xu  and
      Shi, Xiaodong  and
      Liu, Zhiyuan  and
      Sun, Maosong",
    editor = "Che, Wanxiang  and
      Nabende, Joyce  and
      Shutova, Ekaterina  and
      Pilehvar, Mohammad Taher",
    booktitle = "Proceedings of the 63rd Annual Meeting of the Association for Computational Linguistics (Volume 1: Long Papers)",
    month = jul,
    year = "2025",
    address = "Vienna, Austria",
    publisher = "Association for Computational Linguistics",
    url = "https://aclanthology.org/2025.acl-long.1341/",
    doi = "10.18653/v1/2025.acl-long.1341",
    pages = "27664--27678",
    ISBN = "979-8-89176-251-0"
}

@article{gidiotis2020divide,
  title={A divide-and-conquer approach to the summarization of long documents},
  author={Gidiotis, Alexios and Tsoumakas, Grigorios},
  journal={IEEE/ACM Transactions on Audio, Speech, and Language Processing},
  volume={28},
  pages={3029--3040},
  year={2020},
  publisher={IEEE}
}

@article{docetl,
author = {Shankar, Shreya and Chambers, Tristan and Shah, Tarak and Parameswaran, Aditya G. and Wu, Eugene},
title = {DocETL: Agentic Query Rewriting and Evaluation for Complex Document Processing},
year = {2025},
issue_date = {May 2025},
publisher = {VLDB Endowment},
volume = {18},
number = {9},
issn = {2150-8097},
url = {https://doi.org/10.14778/3746405.3746426},
doi = {10.14778/3746405.3746426},
journal = {Proc. VLDB Endow.},
month = sep,
pages = {3035–3048},
numpages = {14}
}

@inproceedings{sarmah2024hybridrag,
author = {Sarmah, Bhaskarjit and Mehta, Dhagash and Hall, Benika and Rao, Rohan and Patel, Sunil and Pasquali, Stefano},
title = {HybridRAG: Integrating Knowledge Graphs and Vector Retrieval Augmented Generation for Efficient Information Extraction},
year = {2024},
isbn = {9798400710810},
publisher = {Association for Computing Machinery},
address = {New York, NY, USA},
url = {https://doi.org/10.1145/3677052.3698671},
doi = {10.1145/3677052.3698671},
booktitle = {Proceedings of the 5th ACM International Conference on AI in Finance},
pages = {608–616},
numpages = {9},
location = {Brooklyn, NY, USA},
series = {ICAIF '24}
}

@inproceedings{han2024xbrl,
author = {Han, Shijie and Kang, Haoqiang and Jin, Bo and Liu, Xiao-Yang and Yang, Steve Y},
title = {XBRL Agent: Leveraging Large Language Models for Financial Report Analysis},
year = {2024},
isbn = {9798400710810},
publisher = {Association for Computing Machinery},
address = {New York, NY, USA},
url = {https://doi.org/10.1145/3677052.3698614},
doi = {10.1145/3677052.3698614},
booktitle = {Proceedings of the 5th ACM International Conference on AI in Finance},
pages = {856–864},
numpages = {9},
location = {Brooklyn, NY, USA},
series = {ICAIF '24}
}

@article{wang2025finsage,
  title={Finsage: A multi-aspect rag system for financial filings question answering},
  author={Wang, Xinyu and Chi, Jijun and Tai, Zhenghan and Kwok, Tung Sum Thomas and Li, Muzhi and Li, Zhuhong and He, Hailin and Hua, Yuchen and Lu, Peng and Wang, Suyuchen and others},
  journal={arXiv preprint arXiv:2504.14493},
  year={2025}
}

@inproceedings{choe2025hirec,
    title = "Hierarchical Retrieval with Evidence Curation for Open-Domain Financial Question Answering on Standardized Documents",
    author = "Choe, Jaeyoung  and
      Kim, Jihoon  and
      Jung, Woohwan",
    editor = "Che, Wanxiang  and
      Nabende, Joyce  and
      Shutova, Ekaterina  and
      Pilehvar, Mohammad Taher",
    booktitle = "Findings of the Association for Computational Linguistics: ACL 2025",
    month = jul,
    year = "2025",
    address = "Vienna, Austria",
    publisher = "Association for Computational Linguistics",
    url = "https://aclanthology.org/2025.findings-acl.855/",
    doi = "10.18653/v1/2025.findings-acl.855",
    pages = "16663--16681",
    ISBN = "979-8-89176-256-5"
}

@inproceedings{Wang2024Loong,
  author    = {Minzheng Wang and Longze Chen and Cheng Fu and others},
  title     = {Leave No Document Behind: Benchmarking Long-Context LLMs with Extended Multi-Doc QA (Loong)},
  booktitle = {EMNLP},
  year      = {2024},
  note      = {arXiv:2406.17419}
}

@article{Liu2023LostMiddle,
  author  = {Nelson F. Liu and Kevin Lin and John Hewitt and others},
  title   = {Lost in the Middle: How Language Models Use Long Contexts},
  journal = {arXiv:2307.03172},
  year    = {2023}
}

@article{Islam2023FinanceBench,
  author  = {Pranab Islam and Anand Kannappan and Douwe Kiela and Rebecca Qian and Nino Scherrer and Bertie Vidgen},
  title   = {FinanceBench: A New Benchmark for Financial Question Answering},
  journal = {arXiv:2311.11944},
  year    = {2023}
}

@misc{bbc2025hallucination,
  author = {BBC},
  title = {Representation of BBC News content in AI Assistants},
  howpublished = {\url{https://www.bbc.co.uk/aboutthebbc/documents/bbc-research-into-ai-assistants.pdf}},
  year = {2025},
  month = feb,
  note = {Accessed: 2025-12-04}}

@article{liu2023evaluating,
  title={Evaluating verifiability in generative search engines},
  author={Liu, Nelson F and Zhang, Tianyi and Liang, Percy},
  journal={arXiv preprint arXiv:2304.09848},
  year={2023}
}

@article{auer2024docling,
  title={Docling technical report},
  author={Auer, Christoph and Lysak, Maksym and Nassar, Ahmed and Dolfi, Michele and Livathinos, Nikolaos and Vagenas, Panos and Ramis, Cesar Berrospi and Omenetti, Matteo and Lindlbauer, Fabian and Dinkla, Kasper and others},
  journal={arXiv preprint arXiv:2408.09869},
  year={2024}
}

@inproceedings{chu2016data,
  title={Data cleaning: Overview and emerging challenges},
  author={Chu, Xu and Ilyas, Ihab F and Krishnan, Sanjay and Wang, Jiannan},
  booktitle={Proceedings of the 2016 international conference on management of data},
  pages={2201--2206},
  year={2016}
}

@article{lewis2020retrieval,
  title={Retrieval-augmented generation for knowledge-intensive nlp tasks},
  author={Lewis, Patrick and Perez, Ethan and Piktus, Aleksandra and Petroni, Fabio and Karpukhin, Vladimir and Goyal, Naman and K{\"u}ttler, Heinrich and Lewis, Mike and Yih, Wen-tau and Rockt{\"a}schel, Tim and others},
  journal={Advances in neural information processing systems},
  volume={33},
  pages={9459--9474},
  year={2020}
}

@article{sollaci2004introduction,
  title={The introduction, methods, results, and discussion (IMRAD) structure: a fifty-year survey},
  author={Sollaci, Luciana B and Pereira, Mauricio G},
  journal={Journal of the medical library association},
  volume={92},
  number={3},
  pages={364},
  year={2004}
}

@article{mann1988rhetorical,
  title={Rhetorical structure theory: Toward a functional theory of text organization},
  author={Mann, William C and Thompson, Sandra A},
  journal={Text-interdisciplinary Journal for the Study of Discourse},
  volume={8},
  number={3},
  pages={243--281},
  year={1988},
  publisher={Walter de Gruyter, Berlin/New York Berlin, New York}
}

@article{bertsch2025oolong,
  title={Oolong: Evaluating long context reasoning and aggregation capabilities},
  author={Bertsch, Amanda and Pratapa, Adithya and Mitamura, Teruko and Neubig, Graham and Gormley, Matthew R},
  journal={arXiv preprint arXiv:2511.02817},
  year={2025}
}

@article{goldman2024really,
  title={Is it really long context if all you need is retrieval? towards genuinely difficult long context nlp},
  author={Goldman, Omer and Jacovi, Alon and Slobodkin, Aviv and Maimon, Aviya and Dagan, Ido and Tsarfaty, Reut},
  journal={arXiv preprint arXiv:2407.00402},
  year={2024}
}

@article{rekatsinas2017holoclean,
  title={Holoclean: Holistic data repairs with probabilistic inference},
  author={Rekatsinas, Theodoros and Chu, Xu and Ilyas, Ihab F and R{\'e}, Christopher},
  journal={arXiv preprint arXiv:1702.00820},
  year={2017}
}

@article{beskales2010sampling,
  title={Sampling the repairs of functional dependency violations under hard constraints},
  author={Beskales, George and Ilyas, Ihab F and Golab, Lukasz},
  journal={Proceedings of the VLDB Endowment},
  volume={3},
  number={1-2},
  pages={197--207},
  year={2010},
  publisher={VLDB Endowment}
}

@inproceedings{zhang2025data,
  title={Data cleaning using large language models},
  author={Zhang, Shuo and Huang, Zezhou and Wu, Eugene},
  booktitle={2025 IEEE 41st International Conference on Data Engineering Workshops (ICDEW)},
  pages={28--32},
  year={2025},
  organization={IEEE}
}

@article{ahmad2023retclean,
  title={Retclean: Retrieval-based data cleaning using foundation models and data lakes},
  author={Ahmad, Mohammad Shahmeer and Naeem, Zan Ahmad and Eltabakh, Mohamed and Ouzzani, Mourad and Tang, Nan},
  journal={arXiv preprint arXiv:2303.16909},
  year={2023}
}

@inproceedings{ni2024iterclean,
  title={Iterclean: An iterative data cleaning framework with large language models},
  author={Ni, Wei and Zhang, Kaihang and Miao, Xiaoye and Zhao, Xiangyu and Wu, Yangyang and Yin, Jianwei},
  booktitle={Proceedings of the ACM Turing Award Celebration Conference-China 2024},
  pages={100--105},
  year={2024}
}

@misc{zhang2025recursivelanguagemodels,
      title={Recursive Language Models}, 
      author={Alex L. Zhang and Tim Kraska and Omar Khattab},
      year={2025},
      eprint={2512.24601},
      archivePrefix={arXiv},
      primaryClass={cs.AI},
      url={https://arxiv.org/abs/2512.24601}, 
}

@article{jiang2024longrag,
  title={LongRAG: Enhancing Retrieval-Augmented Generation with Long-context LLMs},
  author={Jiang, Ziyan and Ma, Xueguang and Chen, Wenhu},
  journal={arXiv preprint arXiv:2406.15319},
  year={2024}
}

@article{edge2024graphrag,
  title={From Local to Global: A Graph RAG Approach to Query-Focused Summarization},
  author={Edge, Darren and Trinh, Ha and Cheng, Newman and Bradley, Joshua and Chao, Alex and Mody, Apurva and Truitt, Steven and Larson, Jonathan},
  journal={arXiv preprint arXiv:2404.16130},
  year={2024}
}

@misc{qwen2026qwen35,
  title        = {Qwen3.5: Towards Native Multimodal Agents},
  author       = {{Qwen Team}},
  year         = {2026},
  month        = feb,
  day          = {15},
  howpublished = {\url{https://qwen.ai/blog?id=qwen3.5}},
  note         = {Blog post}
}

@book{chandler2019cochrane,
  title={Cochrane Handbook for Systematic Reviews of Interventions},
  editor={Higgins, Julian P. T. and Thomas, James and Chandler, Jacqueline and Cumpston, Miranda and Li, Tianjing and Page, Matthew J. and Welch, Vivian A.},
  edition={2nd},
  publisher={John Wiley \& Sons},
  address={Chichester, UK},
  year={2019}
}

@article{chen2025browsecomp,
  title={Browsecomp-plus: A more fair and transparent evaluation benchmark of deep-research agent},
  author={Chen, Zijian and Ma, Xueguang and Zhuang, Shengyao and Nie, Ping and Zou, Kai and Liu, Andrew and Green, Joshua and Patel, Kshama and Meng, Ruoxi and Su, Mingyi and others},
  journal={arXiv preprint arXiv:2508.06600},
  year={2025}
}

@inproceedings{liu2024ra,
  title={Ra-isf: Learning to answer and understand from retrieval augmentation via iterative self-feedback},
  author={Liu, Yanming and Peng, Xinyue and Zhang, Xuhong and Liu, Weihao and Yin, Jianwei and Cao, Jiannan and Du, Tianyu},
  booktitle={Findings of the Association for Computational Linguistics: ACL 2024},
  pages={4730--4749},
  year={2024}
}

@inproceedings{sarthi2024raptor,
  title={Raptor: Recursive abstractive processing for tree-organized retrieval},
  author={Sarthi, Parth and Abdullah, Salman and Tuli, Aditi and Khanna, Shubh and Goldie, Anna and Manning, Christopher D},
  booktitle={The Twelfth International Conference on Learning Representations},
  year={2024}
}

@article{guo2024lightrag,
  title={Lightrag: Simple and fast retrieval-augmented generation},
  author={Guo, Zirui and Xia, Lianghao and Yu, Yanhua and Ao, Tian and Huang, Chao},
  journal={arXiv preprint arXiv:2410.05779},
  volume={2},
  number={3},
  year={2024}
}

@article{gutierrez2024hipporag,
  title={Hipporag: Neurobiologically inspired long-term memory for large language models},
  author={Guti{\'e}rrez, Bernal J and Shu, Yiheng and Gu, Yu and Yasunaga, Michihiro and Su, Yu},
  journal={Advances in neural information processing systems},
  volume={37},
  pages={59532--59569},
  year={2024}
}

@inproceedings{jiang2023active,
  title={Active retrieval augmented generation},
  author={Jiang, Zhengbao and Xu, Frank F and Gao, Luyu and Sun, Zhiqing and Liu, Qian and Dwivedi-Yu, Jane and Yang, Yiming and Callan, Jamie and Neubig, Graham},
  booktitle={Proceedings of the 2023 conference on empirical methods in natural language processing},
  pages={7969--7992},
  year={2023}
}

@inproceedings{trivedi-etal-2023-interleaving,
    title = "Interleaving Retrieval with Chain-of-Thought Reasoning for Knowledge-Intensive Multi-Step Questions",
    author = "Trivedi, Harsh  and
      Balasubramanian, Niranjan  and
      Khot, Tushar  and
      Sabharwal, Ashish",
    editor = "Rogers, Anna  and
      Boyd-Graber, Jordan  and
      Okazaki, Naoaki",
    booktitle = "Proceedings of the 61st Annual Meeting of the Association for Computational Linguistics (Volume 1: Long Papers)",
    month = jul,
    year = "2023",
    address = "Toronto, Canada",
    publisher = "Association for Computational Linguistics",
    url = "https://aclanthology.org/2023.acl-long.557/",
    doi = "10.18653/v1/2023.acl-long.557",
    pages = "10014--10037"
}

@inproceedings{asai2023self,
  title={Self-rag: Learning to retrieve, generate, and critique through self-reflection},
  author={Asai, Akari and Wu, Zeqiu and Wang, Yizhong and Sil, Avirup and Hajishirzi, Hannaneh},
  booktitle={The Twelfth International Conference on Learning Representations},
  year={2023}
}

@article{yan2024corrective,
  title={Corrective retrieval augmented generation},
  author={Yan, Shi-Qi and Gu, Jia-Chen and Zhu, Yun and Ling, Zhen-Hua},
  year={2024}
}

@inproceedings{guu2020retrieval,
  title={Retrieval augmented language model pre-training},
  author={Guu, Kelvin and Lee, Kenton and Tung, Zora and Pasupat, Panupong and Chang, Mingwei},
  booktitle={International conference on machine learning},
  pages={3929--3938},
  year={2020},
  organization={PMLR}
}

@inproceedings{karpukhin2020dense,
  title={Dense passage retrieval for open-domain question answering},
  author={Karpukhin, Vladimir and Oguz, Barlas and Min, Sewon and Lewis, Patrick and Wu, Ledell and Edunov, Sergey and Chen, Danqi and Yih, Wen-tau},
  booktitle={Proceedings of the 2020 conference on empirical methods in natural language processing (EMNLP)},
  pages={6769--6781},
  year={2020}
}

@inproceedings{izacard-grave-2021-leveraging,
    title = "Leveraging Passage Retrieval with Generative Models for Open Domain Question Answering",
    author = "Izacard, Gautier  and
      Grave, Edouard",
    editor = "Merlo, Paola  and
      Tiedemann, Jorg  and
      Tsarfaty, Reut",
    booktitle = "Proceedings of the 16th Conference of the European Chapter of the Association for Computational Linguistics: Main Volume",
    month = apr,
    year = "2021",
    address = "Online",
    publisher = "Association for Computational Linguistics",
    url = "https://aclanthology.org/2021.eacl-main.74/",
    doi = "10.18653/v1/2021.eacl-main.74",
    pages = "874--880"
}

@inproceedings{khattab2020colbert,
  title={Colbert: Efficient and effective passage search via contextualized late interaction over bert},
  author={Khattab, Omar and Zaharia, Matei},
  booktitle={Proceedings of the 43rd International ACM SIGIR conference on research and development in Information Retrieval},
  pages={39--48},
  year={2020}
}

@article{izacard2021unsupervised,
  title={Unsupervised dense information retrieval with contrastive learning},
  author={Izacard, Gautier and Caron, Mathilde and Hosseini, Lucas and Riedel, Sebastian and Bojanowski, Piotr and Joulin, Armand and Grave, Edouard},
  journal={arXiv preprint arXiv:2112.09118},
  year={2021}
}

@incollection{lasserson2024starting,
  title = {Chapter 1: Starting a review},
  author = {Lasserson, Toby J. and Thomas, James and Higgins, Julian P. T.},
  booktitle = {Cochrane Handbook for Systematic Reviews of Interventions},
  editor = {Higgins, Julian P. T. and Thomas, James and Chandler, Jacqueline and Cumpston, Miranda and Li, Tianjing and Page, Matthew J. and Welch, Vivian A.},
  edition = {Version 6.5},
  publisher = {Cochrane},
  year = {2024},
  note = {Chapter last updated August 2021},
  url = {https://www.cochrane.org/authors/handbooks-and-manuals/handbook/current/chapter-01}
}

@article{page2021prisma,
  title = {The PRISMA 2020 statement: An updated guideline for reporting systematic reviews},
  author = {Page, Matthew J. and McKenzie, Joanne E. and Bossuyt, Patrick M. and Boutron, Isabelle and Hoffmann, Tammy C. and Mulrow, Cynthia D. and Shamseer, Larissa and Tetzlaff, Jennifer M. and Akl, Elie A. and Brennan, Sue E. and Chou, Roger and Glanville, Julie and Grimshaw, Jeremy M. and Hr{\'o}bjartsson, Asbj{\o}rn and Lalu, Manoj M. and Li, Tianjing and Loder, Elizabeth W. and Mayo-Wilson, Evan and McDonald, Steve and McGuinness, Luke A. and Stewart, Lesley A. and Thomas, James and Tricco, Andrea C. and Welch, Vivian A. and Whiting, Penny and Moher, David},
  journal = {BMJ},
  volume = {372},
  pages = {n71},
  year = {2021},
  doi = {10.1136/bmj.n71},
  url = {https://www.bmj.com/content/372/bmj.n71}
}

@article{borah2017analysis,
  title={Analysis of the time and workers needed to conduct systematic reviews of medical interventions using data from the PROSPERO registry},
  author={Borah, Rohit and Brown, Andrew W and Capers, Patrice L and Kaiser, Kathryn A},
  journal={BMJ open},
  volume={7},
  number={2},
  pages={e012545},
  year={2017},
  publisher={British Medical Journal Publishing Group}
}

@book{petticrew2006systematic,
  title={Systematic Reviews in the Social Sciences: A Practical Guide},
  author={Petticrew, Mark and Roberts, Helen},
  year={2006},
  publisher={Blackwell Publishing}
}

@article{tranfield2003towards,
  title={Towards a methodology for developing evidence-informed management knowledge by means of systematic review},
  author={Tranfield, David and Denyer, David and Smart, Palminder},
  journal={British journal of management},
  volume={14},
  number={3},
  pages={207--222},
  year={2003},
  publisher={Wiley Online Library}
}

@article{quinn2025systematic,
  title={Systematic literature reviews in accounting, finance and governance},
  author={Quinn, Martin and Hiebl, Martin RW and Farrell, Michael},
  journal={Accounting, Finance and Governance Review},
  year={2025},
  publisher={Irish Accounting and Finance Association}
}

@article{baude2017making,
  title={Making doctrinal work more rigorous: Lessons from systematic reviews},
  author={Baude, William and Chilton, Adam S and Malani, Anup},
  journal={The University of Chicago Law Review},
  pages={37--58},
  year={2017},
  publisher={JSTOR}
}

@techreport{keele2007guidelines,
  title={Guidelines for Performing Systematic Literature Reviews in Software Engineering},
  author={Kitchenham, Barbara and Charters, Stuart},
  year={2007},
  number={EBSE-2007-01, version 2.3},
  institution={Keele University and Durham University}
}

@inproceedings{polzak2026medevidence,
  title={Can Large Language Models Match the Conclusions of Systematic Reviews?},
  author={Polzak, Christopher and Lozano, Alejandro and Sun, Min Woo and Burgess, James and Zhang, Yuhui and Wu, Kevin and Chiang, Chia-Chun and Nirschl, Jeffrey J and Yeung-Levy, Serena},
  booktitle={International Conference on Learning Representations},
  year={2026}
}

@misc{elicit2026systematic,
  author = {{Elicit}},
  title = {Elicit for Systematic Reviews},
  howpublished = {\url{https://elicit.com/solutions/systematic-review}},
  year = {2026},
  note = {Accessed 2026-11-02}
}

@article{wu2026can,
  title={What can we learn from 1,000 meta-analyses across 10 different disciplines?},
  author={Wu, Weilun and Duan, Jianhua and Reed, W Robert and Tipton, Elizabeth},
  journal={Research Synthesis Methods},
  volume={17},
  number={1},
  pages={123--156},
  year={2026},
  publisher={Cambridge University Press}
}

@article{li2019randomized,
  title={A randomized trial provided new evidence on the accuracy and efficiency of traditional vs. electronically annotated abstraction approaches in systematic reviews},
  author={Li, Tianjing and Saldanha, Ian J and Jap, Jens and Smith, Bryant T and Canner, Joseph and Hutfless, Susan M and Branch, Vernal and Carini, Simona and Chan, Wiley and de Bruijn, Berry and others},
  journal={Journal of Clinical Epidemiology},
  volume={115},
  pages={77--89},
  year={2019},
  publisher={Elsevier}
}

@article{gartlehner2024data,
  title={Data extraction for evidence synthesis using a large language model: A proof-of-concept study},
  author={Gartlehner, Gerald and Kahwati, Leila and Hilscher, Rainer and Thomas, Ian and Kugley, Shannon and Crotty, Karen and Viswanathan, Meera and Nussbaumer-Streit, Barbara and Booth, Graham and Erskine, Nathaniel and others},
  journal={Research synthesis methods},
  volume={15},
  number={4},
  pages={576--589},
  year={2024},
  publisher={Wiley Online Library}
}

@article{riis2025chatgpt,
  title={ChatGPT-4o can serve as the second rater for data extraction in systematic reviews},
  author={Riis, Johannes and Kristjansen, Karoline Assifuah and Andersen, Stig and Okubo, Yoshiro and J{\o}rgensen, Martin Gr{\o}nbech},
  year={2025}
}

@article{smyth2026quantifying,
  title={Quantifying overclaiming propensity in frontier LLM agents},
  author={Smyth, Nolan and Mantilla-Ramos, Yorguin-Jose and Notsawo, Pascal Jr Tikeng and Helbling, Saskia and Tosato, Alberto and Merzouk, Mohamed Amine and Dziri, Nouha and Gidel, Gauthier and Tosato, Tommaso},
  journal={arXiv preprint arXiv:2609.20812},
  year={2026}
}

@article{liang2025surveyx,
  title={Surveyx: Academic survey automation via large language models},
  author={Liang, Xun and Yang, Jiawei and Wang, Yezhaohui and Tang, Chen and Zheng, Zifan and Song, Shichao and Lin, Zehao and Yang, Yebin and Niu, Simin and Wang, Hanyu and others},
  journal={arXiv preprint arXiv:2502.14776},
  year={2025}
}

@inproceedings{wang-etal-2026-arxiv2table,
    title = "ar{X}iv2{T}able: Toward Realistic Benchmarking and Evaluation for {LLM}-Based Literature-Review Table Generation",
    author = "Wang, Weiqi  and
      Ou, Jiefu  and
      Song, Yangqiu  and
      Van Durme, Benjamin  and
      Khashabi, Daniel",
    editor = "Liakata, Maria  and
      Moreira, Viviane P.  and
      Zhang, Jiajun  and
      Jurgens, David",
    booktitle = "Proceedings of the 64th Annual Meeting of the {A}ssociation for {C}omputational {L}inguistics (Volume 1: Long Papers)",
    month = jul,
    year = "2026",
    address = "San Diego, California, United States",
    publisher = "Association for Computational Linguistics",
    url = "https://aclanthology.org/2026.acl-long.346/",
    doi = "10.18653/v1/2026.acl-long.346",
    pages = "7602--7624",
    ISBN = "979-8-89176-390-6"
}

%%%%%%%%%%%%%%%%%%%%%%%%%%%%%%%%%%%%%%%%%%%%%%%%%%%%%%%%%%%%%%%%%%%%%%%%%%%%%%%
%%%%%%%%%%%%%%%%%%%%%%%%%%%%%%%%%%%%%%%%%%%%%%%%%%%%%%%%%%%%%%%%%%%%%%%%%%%%%%%
% APPENDIX
%%%%%%%%%%%%%%%%%%%%%%%%%%%%%%%%%%%%%%%%%%%%%%%%%%%%%%%%%%%%%%%%%%%%%%%%%%%%%%%
%%%%%%%%%%%%%%%%%%%%%%%%%%%%%%%%%%%%%%%%%%%%%%%%%%%%%%%%%%%%%%%%%%%%%%%%%%%%%%%
\newpage
\appendix
\onecolumn
\renewcommand{\thepart}{}
\renewcommand{\partname}{}
\noptcrule
\setcounter{parttocdepth}{3}
\part{Appendix}
\parttoc
\newpage

\section{Method Details}
\label{app:method}

\subsection{Question Decomposition}
\label{app:question_decomposition}
The original question posed by the user is not directly suitable for all stages of the pipeline.
In particular, extraction and reconciliation are intermediate steps that impose different requirements on task formulation.
Extraction should enumerate all relevant candidate evidence from each document chunk, while reconciliation should combine these candidates into a globally consistent database.
Conditioning either stage on the original question can cause them to prematurely answer the question rather than constructing the intermediate representation.

Consider the query: ``Return the second poem about the Great Wall of China.'' 
Suppose the first chunk contains one relevant poem (P1), and the second chunk contains two relevant poems (P2, P3). 
If the extractor is conditioned on the original question, it may incorrectly interpret the ordinal constraint ``second'' at the chunk level:
it extracts nothing from the first chunk (since no second poem exists locally) and extracts P3 from the second chunk (as the second poem within that chunk).
This results in an incorrect global outcome, since the true second poem across the entire document collection is P2. 
Similar issues arise for other global constraints such as superlatives, rankings, and aggregations.

To prevent such premature answer-oriented behavior, \system{} decomposes the input question 
$q$ into component-specific queries. 
We generate an extraction query $q_e$ that specifies what information should be identified from each document chunk, and a reconciliation query $q_r$ that defines how partial extractions should be combined and cleaned to form a coherent database. 
Schema induction and final answer synthesis, in contrast, operate directly on the original question $q$, as these stages are explicitly responsible for capturing the full semantic intent of the user's request and producing the final answer.

\subsection{Context-Aware Chunking}
\label{app:chunking}

Each chunk retains the raw text as well as its structural metadata, including the current heading path (e.g., Header 1 $\rightarrow$ Header 1.1 $\rightarrow$ Header 1.1.3), the chunk index, the document title, and the document description.
This ensures that every chunk is self-contained, carrying both global and local context needed for faithful extraction.

\subsection{Schema Induction Templates}
\label{app:schema-guidelines}
Here we provide the schema library we provide \system{}.

\begin{lstlisting}[
  basicstyle=\ttfamily\tiny,
  breaklines=true,
  frame=single,
  xleftmargin=0.3cm,
  xrightmargin=0.3cm
]
# Question Type Guidelines

## Ordering and Retrieval Questions
- The schema should not contain the index of the item since extraction will be done at chunk level, and won't have the global index.

## Multiple Choice Questions
- The schema should contain reasons for and against each option.
- Include fields such as 'option_label' (e.g., 'option A name', 'option B name', 'option C name'...), 'option_text', and 'support_label'.
- You can use enums as a field which says: 'supported', 'not_supported', 'insufficient_evidence', 'contradicted'.

# Document Type Guidelines

## Dataset Documents
- You should include any given identifier so that datapoints can be aligned across pages/chunks.
- Include at least one label field (e.g., 'label') plus any other attributes needed to answer or aggregate for the question.

## Story/Narrative Documents
- Define an 'Entity' table to capture the relationship 'entity $\leftrightarrow$ properties mentioned on this page'.
- Include fields such as 'entity_name', 'entity_type', and a field summarizing important properties on this page (e.g., 'entity_information').
- If the question depends on actions or plot, define an 'Event' (or similar) table representing 'event $\leftrightarrow$ participants/time/location', with one row per event.
- Canonicalize the names across chunks when possible so that information about the same entity/event can be merged across chunks.

## Policy/Whitepaper Documents
- You should have fields like stakeholders, implementation steps, conditions, etc
- If it talks about recommendations, then you make sure to add fields like target_area/domain, intended outcome, advantages, disadvantages, etc
\end{lstlisting}

\subsection{SLIDERS Implementation Details}
\label{app:implementation}

\paragraph{Metadata Extraction} To identify headers, tables and figures in a document, we use DocLing~\cite{auer2024docling}, an off-the-shelf pdf to markdown converter and parse the markdown to identify the structural metadata.
In scenarios where the original pdf is not available, we preprocess the raw text to identify, tables and section headers using LLMs.

\textbf{Compute Usage. } For Open AI models (GPT 4.1, GPT 4.1-mini, GPT 5, GPT 5-mini), we use Microsoft Azure service. For Qwen3-4B embedding model and Qwen3.5 122B-A10, we run inference on a GPU cluster, using A100, H100 and H200 Nvidia GPUs. For Qwen3.5 baseLM, we used 6 H100 GPUs at a time. For SLIDERS we used 4 H100 GPUs. Finally for Qwen3-4B we use 1 A100.

\textbf{Models. }For \system{}, GPT-4.1-mini handles evidence variable aware extraction and GPT-4.1 handles evidence-variable generation, evidence reconciliation, and final evidence-grounded synthesis; we additionally report a fully open-source variant using Qwen3.5-122B-A10B for all components.

\subsection{Sample Configuration}
\label{app:config}

We provide a representative configuration file for the Loong Finance (EN) benchmark.

\begin{lstlisting}[
  language={},
  basicstyle=\ttfamily\tiny,
  breaklines=true,
  frame=single,
  xleftmargin=0.3cm,
  xrightmargin=0.3cm,
]
experiment: loong
system: sliders
system_config:
  generate_task_guidelines: false
  rephrase_question:
    enabled: true
    prompt_file: sliders/rephrase_question_component.prompt
    library_of_guidelines_path: sliders_taxonomy.json
  generate_schema:
    add_extra_information_class: false
    add_document_text: true
    generate_schema_type: library_based
    library_of_guidelines_path: sliders_taxonomy.json
  extract_schema:
    decompose_fields: false
    dedupe_merged_rows: false
    num_samples_per_chunk: 1
    is_relevant_chunk: true
    extract_quotes: true
  merge_tables:
    merge_strategy: recon_v2
    reconciliation:
      debug_mode: false
      primary_key_selection:
        version: v1
        max_candidates: 10
        max_inspections_per_candidate: 10
        max_retries: 3
      canonicalization:
        mode: two_pass
        max_cycles: 20
        max_retries_per_cycle: 3
        max_inspection_history: 3
        inspections_per_field: 50
        null_handling:
          max_inspections: 5
          placeholder_text: UNKNOWN
          verification:
            enable: false
            max_inspections: 5
      controller_executor_loop:
        max_iterations: 5
        max_controller_inspections: 5
        max_executor_inspections: 5
        max_sql_attempts: 3
        verification:
          enable: true
          max_inspections: 5
        context_generator:
          enable: true
          model: gpt-4.1
          max_rows: 20
      non_pk_canonicalization:
        enable: true
        inspections_per_field: 5
        column_selector:
          max_inspections: 5
          excluded_columns:
            - row_id
            - page_number
            - __reconciliation_context__
            - number_instances
          excluded_patterns:
            - _quote
            - _rationale
            - _confidence
      statistics:
        enable: true
        filename: reconciliation_stats.json
  inspect_answer:
    enable_citation_generation: true
    enable_reconciliation_stats_verbalization: true
  check_if_merge_needed: false
  force_sql: false
  answer_strategy: sql_inspect
  models:
    answer:
      model: gpt-4.1
      max_tokens: 8192
      temperature: 0.0
    answer_no_table:
      model: gpt-4.1-mini
      max_tokens: 8192
      temperature: 0.0
    answer_tool_output:
      model: gpt-4.1
      max_tokens: 8192
      temperature: 0.0
    extract_schema:
      model: gpt-4.1-mini
      max_tokens: 32000
      temperature: 0.0
    generate_schema:
      model: gpt-4.1
      max_tokens: 8192
      temperature: 0.0
    merge_tables:
      model: gpt-4.1
      max_tokens: 32000
      temperature: 0.0
    task_guidelines:
      model: gpt-4.1
      max_tokens: 8192
      temperature: 0.0
    check_objective_necessity:
      model: gpt-4.1
      max_tokens: 8192
      temperature: 0.0
    rephrase_question:
      model: gpt-4.1
      max_tokens: 8192
      temperature: 0.0
    direct_answer:
      model: gpt-4.1
      max_tokens: 8192
      temperature: 0.0
    force_answer:
      model: gpt-4.1
      max_tokens: 8192
      temperature: 0.0
    is_relevant_chunk:
      model: gpt-4.1-mini
      max_tokens: 8192
      temperature: 0.0
    check_if_merge_needed:
      model: gpt-4.1
      max_tokens: 8192
      temperature: 0.0
    select_guidelines_for_schema:
      model: gpt-4.1-mini
      max_tokens: 2000
      temperature: 0.0
    select_primary_key:
      model: gpt-4.1
      max_tokens: 8192
      temperature: 0.0
    canonicalize_fields:
      model: gpt-4.1
      max_tokens: 8192
      temperature: 0.0
    column_selector:
      model: gpt-4.1
      max_tokens: 8192
      temperature: 0.0
    context_generator:
      model: gpt-4.1
      max_tokens: 8192
      temperature: 0.0
    handle_null_pks:
      model: gpt-4.1
      max_tokens: 8192
      temperature: 0.0
experiment_config:
  benchmark_path: /data1/hypothesis_dataset/loong/loong_processed.jsonl
  gpt_results_path: None
  files_dir: /data1/hypothesis_dataset/loong/doc/
  specific_ids_csv: sample_data/50_sample_ids_finance_en_final.csv
  soft_evaluator_model: gpt-4.1
  hard_evaluator_model: gpt-4.1
  num_questions: null
  random_state: 42
  document_config:
    chunk_size: 16000
    overlap_size: 0
  filter_by_type: "financial"
output_file: loong_sliders_finance_en_26jan.json

\end{lstlisting}

\section{Benchmarks}
\label{app:benchmark_descriptions}

\subsection{Evaluation Metrics}
For BrowseComp+ we use the official evaluation method, reporting binary accuracy for answering questions. For WikiCeleb100 and FinQ100 we use llm as a judge and partial scoring similar to Oolong's numeric metric. 
For FinanceBench, we use LLM-as-a-judge evaluation and use the official metrics for Loong and Oolong.

\subsection{Benchmark Statistics}
\label{app:benchmark_statistics}
As shown in Table~\ref{tab:benchmark_stats}, the benchmark suite covers a range of document scales, data sources, and reasoning requirements, from single-document extraction to large-corpus multi-hop retrieval. FinanceBench contains one document per question and focuses on realistic financial extraction and arithmetic over 150 questions. Loong and Oolong evaluate mid-scale settings: Loong uses roughly 11 real documents per question and emphasizes retrieval and aggregation, while Oolong uses synthetic single-document inputs for classification and aggregation. The remaining benchmarks stress larger evidence sets. WikiCeleb100 and FinQ100 each contain 100 documents, requiring aggregation, comparison, and arithmetic over real-world corpora. BrowseComp+ is the largest setting, with 1,000 documents and 150 questions designed to test multi-hop retrieval and reasoning across a substantially broader context.

\begin{table}[t]
\centering
\small
\caption{Benchmark statistics.}
\label{tab:benchmark_stats}
\begin{tabular}{lrrcl}
\toprule
\textbf{Benchmark} & \textbf{\# Docs} & \textbf{\# Questions} & \textbf{Real/Synth} & \textbf{Task Type} \\
\midrule
FinanceBench & 1 per Q & 150 & Real & Extraction, Arithmetic \\
Loong & $\sim$11 per Q & 200 & Real & Retrieval, Aggregation \\
Oolong & 1 per Q & 192 & Synthetic & Classification, Aggregation \\
WikiCeleb100 & 100 & 22 & Real & Aggregation, Comparison \\
FinQ100 & 100 & 25 & Real & Aggregation, Arithmetic \\
BrowseComp+ & 1000 & 150 & Real & Multi-hop \\
\bottomrule
\end{tabular}
\end{table}

\subsection{Benchmark Construction}
\label{app:benchmark_construction}

Question construction begins from a seed query: for WikiCeleb100, ``Which artist made their debut at the youngest age across Film, Music, Content Creation, and Other industries?''; and for FinQ100, ``Which company has the lowest long-term borrowing?'' We derive follow-up questions by reusing and recombining the extracted information, including temporal cohorts (e.g., artists debuting in the 1950s or born in the 1970s) and aggregate financial properties (e.g., companies with no long-term borrowing or borrowing exceeding \$80M).

To annotate gold answers, we use evidence extraction task to extract an initial structured representation, then manually verify all extracted values against the original Wikipedia articles and financial filings. Automated extraction serves only as a starting point to accelerate the annotation process; the final ground-truth was determined entirely through manual verification and correction. We then manually reconcile the resulting tables to obtain a consolidated database representation of each document set, and manually author SQL queries for each question. For each benchmark, we additionally retain five questions that cannot be answered using SQL alone.

WikiCeleb100 and FinQ100 are primarily follow-up analysis benchmarks: they test whether an evidence table built for an initial research question can support related downstream questions without reprocessing the full corpus. They also function as large-corpus QA benchmarks, but their main purpose is to evaluate reusable evidence-table construction.

We acknowledge that these benchmarks favor systems capable of evidence extraction and reuse. That is intentional: this is the capability being evaluated. However, the evaluation does not give SLIDERS privileged access to the gold table, and competing systems are free to extract, aggregate, or build intermediate representations. The fact that removing reconciliation substantially degrades performance on both WikiCeleb100 and FinQ100 shows that the task is not solved by extraction alone; large-scale grouping, deduplication, conflict handling, and answer-time use of the table are also required.

\textbf{Evaluation. }SLIDERS currently assumes that follow-up questions are answerable from the evidence variables induced for the original research question. If a follow-up requires a missing variable, the current implementation may fail. A natural extension is to detect missing variables, regenerate or extend the schema, and run incremental extraction only for the newly required fields.

\textbf{License}
Wikipedia content is generally copyrighted by its contributors and made available under the Creative Commons Attribution–ShareAlike license, meaning it may be reused with proper attribution and, for adapted text, compatible share-alike licensing. Images and media on Wikipedia must be checked individually because their licenses vary. SEC materials created by the U.S. Securities and Exchange Commission as part of official government work are generally not protected by U.S. copyright in the United States, but SEC logos, trademarks, and third-party content may still be restricted. SEC EDGAR filings are publicly accessible, but because they are submitted by companies and individuals rather than authored by the SEC, their factual data is generally reusable while the expressive text, exhibits, or third-party materials may still be subject to copyright or other reuse limits.

\subsection{Evaluation Reliability}
\label{app:eval_reliability}
\textbf{LLM-as-a-judge variance.} We quantify evaluation variance by running GPT-4.1 (temperature 0.7) three times per instance:
FinanceBench std = 0.47, Loong std = 0.31, Oolong std = 1.02, WikiCeleb std = 0.21, FinQ100 std = 0.38.
Low variance across benchmarks indicates stable evaluation.

\begin{wraptable}{r}{0.50\textwidth}
\centering
\small
\caption{Manual evaluation on 50 questions each. FN/FP denote
false negatives/positives relative to human labels.}
\label{tab:manual_eval}
\begin{tabular}{llccc}
\toprule
\textbf{Dataset} & \textbf{System} & \textbf{Cohen's $\kappa$} & \textbf{FN} & \textbf{FP} \\
\midrule
FinanceBench & SLIDERS  & 0.769 & 4 & 0 \\
FinanceBench & GPT-4.1  & 0.646 & 3 & 2 \\
Oolong       & SLIDERS  & 0.855 & 1 & 1 \\
\midrule
\multicolumn{2}{l}{Combined} & 0.758 & --- & --- \\
\bottomrule
\end{tabular}
\end{wraptable}

\textbf{Human validation of LLM-as-a-judge.}
To validate automatic evaluation, we manually annotated 50
questions each from FinanceBench (for both SLIDERS and GPT-4.1)
and Oolong (for SLIDERS), yielding a combined Cohen's
$\kappa = 0.758$ (substantial agreement).
Table~\ref{tab:manual_eval} breaks down agreement by condition.
Notably, SLIDERS errors on FinanceBench are exclusively false
negatives (4 FN, 0 FP), indicating that our reported scores
are conservative estimates of true accuracy.

\subsection{Loong Ablation Subset IDs}
\label{app:loong_ablation_subset_ids}

For the Loong ablation experiments, we use fixed validation subsets
from the four Loong domains: Finance-EN, Finance-ZH, Papers, and Legal.
To make the ablation setting reproducible, we list the exact question
identifiers used in each subset below.

\paragraph{Finance-EN.}
\begin{lstlisting}[
  basicstyle=\ttfamily\tiny,
  breaklines=true,
  frame=single,
  xleftmargin=0.3cm,
  xrightmargin=0.3cm
]
7182fb1c-7a74-4532-9a2b-0ec0b2508b87
83cac46c-dd52-4583-88a8-a3187841d0ff
ff3aa0ab-8dff-4115-8365-0b734655b779
f606061e-dbf5-4fa9-8a90-08b326bd327e
d1710ae4-0478-460f-aed7-8b9b28fe4304
f7c48b53-aa43-41e5-8ce9-2dc3f88e8d2a
1559effb-ac3b-474a-87d2-b3d31895421f
3a27046e-41ea-4680-9a7f-1c52e2d2117b
e51a4edb-f35b-443f-9e7c-3d8ef9a9a72f
260808aa-4f73-4544-ba29-4c0d1275c13d
45c88b97-dd49-4684-8b29-29b2415c8770
039a5d13-0bb9-4fff-afcf-be51e4c96b99
64e38c31-e0d1-4ea8-8fac-72913e2f6e79
c0c4d8ea-cfb7-4b39-9b8c-b3d1d367e09c
140fcb0e-9e63-4beb-9f5e-1c299c4e5cc2
2ddb83fa-e02a-4e7b-b7c6-ae8ed682833c
96942bc4-573c-4ed9-bd67-f547584bf9fb
8ee4e095-725f-4d7b-bbfb-bfd4209af4a6
0f9a3498-a531-4986-8991-7301785cd398
f64e969a-0cad-4b00-a810-9648de758d5a
7b470b0f-b2aa-4066-825e-8fa2d7d26d77
4f57910f-a512-449c-9845-a3835d26c907
a783c691-7900-465f-8213-d484de803848
4404b6e2-a5f8-4d64-a782-57af3986a3a9
7596de64-a9b7-43ca-b18c-5594c7d9dd68
42496b0c-7b6b-40c6-9c6e-c01ed69421a2
96bada0f-4b00-4312-96d2-39c1521ec277
bec8be58-45b1-4219-948c-bd410342197b
197ead69-2756-4db1-895c-3d4f03926081
70d2f3da-d575-4e98-b43f-ee59f035e856
f05faf41-a05d-4f58-9f21-324c37d45a0c
c3f0bc88-3b62-4162-9bb9-76537c41e590
709984a4-8546-425b-9d49-e4317508a590
30142ca9-9a2f-4ffd-851e-fd05ea818238
238d1b8e-a02a-4d4f-8600-a1b31ef44672
836bddf8-dd83-4e8b-8a4e-0ff992a4ac74
da9fd050-450b-4486-8047-4214ac156127
870d8cb3-6bdd-4428-9871-bbbd6b14d831
9aa8e75a-6804-4d8f-933c-6539ea79168e
6c3c979f-2a58-4e90-b0e1-9adc1c61d327
25d28410-ba8d-4475-a305-f6b9cb3b3bb2
9fd8f0e0-2be6-4b92-9c41-871aa11e0a90
d29b8b36-ba2b-4c61-9f32-6bb29b519bc7
5a035018-d80c-4131-925a-7705466d55b2
4e4a7586-8901-4fdf-bb7e-71d4d95e6a1e
9206d973-c2cf-4f14-8651-3f0fc28ba096
147b7898-1256-45ed-9e92-a6fd396aeeff
0e76dcd2-b0fa-44f8-8b98-3ec18e014720
17461024-fc6a-4f61-a8d8-62294462bd0f
a3c15a25-920d-4247-85e4-044580dd4631
\end{lstlisting}

\paragraph{Finance-ZH.}
\begin{lstlisting}[
  basicstyle=\ttfamily\tiny,
  breaklines=true,
  frame=single,
  xleftmargin=0.3cm,
  xrightmargin=0.3cm
]
360f0e5c-416e-4c4a-afa3-e10518b3b3a2
cae27e9d-deba-4ae8-b55b-430c0eb50e52
0523b4e1-a368-4a23-857e-1fdb2f566444
747f5c6d-add3-4d47-b383-f98671743f78
7f81435c-b7d9-4610-a797-9f94932b358e
9f416db5-2766-4ce0-ae5f-cf54b50df416
94281743-7e98-4786-a466-82205e170839
6c4cab6e-e220-40c3-8e8b-fb649cee0e8b
231d0127-15d7-4425-bfaa-6a2659ed9f23
aa78da15-1854-4667-9e29-e01347e7cee3
e0007015-e0b3-4a25-a3c2-38715f07f430
836626b6-2fdb-4591-8bc8-90be5da23fd7
198f9f86-ce3b-4b20-9495-051394400da1
3e0ed776-9822-4a8a-9c19-051ab8300413
e34df903-e992-4a1c-908e-1b79fbfc9a51
e83edd85-2e17-4b22-9f89-909c6a1811a0
cbd31589-9532-4b79-a9e9-790a832e1a3a
1441c822-56ca-4e5d-a138-d74864fcfbf9
dc466ed9-e56f-47b5-b011-15836308a965
1d88d2b5-b9bf-49bb-80fc-b9e384aa8b10
0f992dd5-5c3d-4bec-997b-fdf7a6696d7f
5bd2b85f-f380-46e8-9832-c4211e1a6c04
3ea9864e-1b26-4726-9fc3-b38ae570442c
589ac24d-3755-4e9d-acf3-31fa3d45e230
de6cdade-8399-4878-9867-d46d6f5ccdcd
79159f9f-97e9-4250-8ca3-0df95478c3f2
9b9202d5-06d9-4692-9fc5-05eb2660f94f
ec62feb9-5817-4ca7-a0a7-fd25b6977b8d
dfa7bce1-33b5-4121-b7f4-efd0fbb07ca2
89e5aec8-f61b-458d-89a3-d00d08cb775f
1937ef3a-c8ec-42d5-a91c-6952c4676a21
2c534637-035b-49f0-b2c5-7984db256ddf
30540dcc-bae5-49f3-8986-95e0b981e8e6
882e85d8-bc46-4ac8-9d26-2b6e80ca8153
bec203d9-7506-42fb-945a-0956b7ca521f
e8fd487d-1981-447c-b927-8ca7ba71486b
9c659026-aa57-45b4-bff7-ee73ada3268c
1b8667b8-e6a3-475d-aaa4-02102804ce4d
9cecf073-04e2-43b3-8174-6eeac8126146
34487085-a4dd-4b9f-a8a0-890865a0c446
aa62cc72-10cb-41d5-8a27-5b8b5e642a06
fa2887f1-7b3f-4899-bdf6-6c8740950af2
9c98b550-383e-460e-b694-974fba81dc6b
8582c4cc-4d10-4126-aa69-a5fae868ecd5
6cd1f216-8ef9-486f-8755-08531eb907dc
c9495554-23fd-4667-a519-7e9139b5d17c
424d60a8-5b9d-475c-bf84-0778fe71950c
78054239-28d8-416c-ad49-c9470f3298dc
631705a6-816c-4f2c-a9b1-bbc844515212
73a7aa5e-0f2a-477f-ac7d-66898ca44427
\end{lstlisting}

\paragraph{Papers.}
\begin{lstlisting}[
  basicstyle=\ttfamily\tiny,
  breaklines=true,
  frame=single,
  xleftmargin=0.3cm,
  xrightmargin=0.3cm
]
949f3450-9b30-4556-be5f-ca8afe9e072f
ad9ca28e-c5db-4686-9200-39efaaf4e0e5
74df278b-9320-46dc-8c71-0aea2419aafa
c4a4e642-fa00-4377-a756-40f2f7ea7a74
42872573-11e0-4406-855c-ad0d83ce0bc4
ffb3aa9f-9152-4a91-9abe-37f09798ba43
3e4ee654-dfd8-4ffd-a825-79e0997961c9
f947493a-511c-4cce-bd2d-7bb7c3d51ddd
1df3da1b-87c8-4692-9ec7-83d1b0e9caa3
90d84852-ab82-4261-ad17-683cd8db3fba
a05907b0-87b2-4776-8be9-2b3ece704493
b84f9183-b962-473b-9fe5-231c0a51398d
f6671f47-58a6-4852-9dd9-c324d97c89b7
aea761bc-1a20-402a-9b4d-bd4118947fd5
29c9920f-3ea3-41e7-b334-fd95a23e6731
0acccefc-0b0e-4758-adcb-dc5d18627ac3
a356842c-b55b-454b-a7ba-a3efc1d020bd
a62d977d-2b61-4bc7-ba0e-22ed1ae176d1
69c49d27-71ee-4acb-8334-6af1be3196f5
1ee9e393-74b2-45e0-a02e-d715a7fec886
b04564ab-bfe7-474d-a2fa-450ea4005ce9
8f43c67e-e96d-4c9e-a54a-04e1217cccce
05d742e6-2804-43a0-ad34-733335dedca5
84bd6e6f-ad9e-4566-a349-501a5854e9ca
2941580d-7018-4c0f-aa7b-97aaa2d73dcb
66bef80e-4e0d-462a-b0bf-873fd6b5b8cf
3925c247-7d83-44bb-bc30-0c2b751e092a
7cb9fcc8-9c3f-4518-bff9-61e85238ab5c
7c0e4a47-b1d9-451a-9cd6-17fd723f187a
0bf88a67-cbf2-49ac-b170-84b506392356
96c4dd20-1260-4541-89d5-f4f199ba7ef2
6622b8f8-aec4-4daa-8fb5-4a7b224ac093
39b03fab-e31e-4159-b7ea-7314b361f0d9
7df5cf53-1601-43c2-9688-6dc6667ef7b3
31ee8539-2e9e-40e2-adc3-082c872698f7
59fc6ed6-6750-4b52-916d-cd466b856922
d652b46e-4fc8-4cdc-ac7f-36aaaba74439
51c625a5-a791-483a-a690-16b959115841
aa9707c4-9506-4e51-899d-9213ad10a324
4e20a21e-b90c-455c-abbf-c62583e2d4c4
cddab93c-c55f-4803-9df4-d28ade76c3d3
dfcd1962-3307-4022-a6ed-e4fd2b023de5
8449c961-673e-4bb9-a1c6-538761165167
b2dc9c73-47fd-416b-ad81-2abc14df6ef0
f1600139-fdb1-4b1f-8658-8d2529e24f9d
4ed5f412-6e70-4d47-ad88-1ffede8b744c
119c5961-858a-42f9-83eb-0a9458d0b5f5
762cf793-7d61-4b42-9273-dcafe1636df0
e853197a-cf5c-457b-b4f2-f851da4971c3
b2842feb-448a-45f1-89b0-d6143407de5b
\end{lstlisting}

\paragraph{Legal.}
\begin{lstlisting}[
  basicstyle=\ttfamily\tiny,
  breaklines=true,
  frame=single,
  xleftmargin=0.3cm,
  xrightmargin=0.3cm
]
786fdcee-3928-481e-a724-4fda67796b2f
8b823edb-0c14-4d63-af75-8af76f58a4de
5a6d86f0-9cdf-43be-9dea-71fb5b6e808b
ff4a9bc4-3742-40cc-9f3a-7e6fe79ddd1b
0812e68a-b2df-43d8-9ee1-777ecfe2c520
70d9e9c5-5c1b-4fe5-bcd3-a0bf9c46bd7f
5c7b562a-f603-49a4-93c4-bd4a1611030d
514713f8-0e3f-49f7-a994-c85fe96f2c9d
86498cea-b26a-4c7b-8bb9-e295135e6676
6d49003b-6922-4644-afba-c6b63e76ecef
02149212-63eb-4c91-a50d-ef8c36948eaf
38273dab-0811-48e0-94a7-636ab1acb552
9e4b4c5b-937c-4aed-ad70-65fb68281c70
dde1534f-b1e5-4b2c-8438-67c4e4e8a5fe
d1e10064-4c0f-4141-8f2d-95a9d481f600
1b36de5b-425b-4f1e-bc0d-8c51177a6895
ec687abf-1121-4716-946d-82b7fa166529
f6a7760f-c09c-42c0-9f72-02f5840a200e
fafed09b-94ef-4b5f-8b36-5252e1e9a136
7273463f-ffee-47da-bb8f-5a10b1b74bd6
8cc235c3-4f58-49d5-bc38-790f804a37a9
de1b3c88-6d88-4d7d-a5cf-e65d225562fa
fb4b1c14-5e23-4e2c-8559-ad0ced3244f2
41f524d8-fa6f-4a42-a25b-4830b92a60bf
ecca0eee-fd47-49e7-bb4f-1ac220cd9420
2022d060-5803-4986-a08c-38f482bfb8e7
b45b29d1-9b91-45f1-b168-f3df01428bda
cd1d4c0a-b3e9-4a10-8f46-eee4d31c7580
b6e0f2b3-33d2-49ce-8593-984f6741a042
3336e301-8970-4ac5-af4b-5386a8b9306f
b749e2d7-7e89-499f-a45d-e7cc3f0e3574
85082e73-c0a9-47ec-a7c3-c32708580638
375c0bba-a360-4414-9ab8-d2e39f9dddb1
19c55985-7496-4338-88a3-df16e2397536
00980600-cf5d-4696-8f2b-2e3a8eaad4ed
a3a1029d-23d6-4614-9886-cb20d43319c2
81dc8271-a7b8-40a3-84ad-84c52ef5c9d0
4d6cd4f8-57d2-4a28-9522-6e4cd8bfbbc7
d2f86999-fd03-46f1-a8c4-8ebc849b055e
d9113b08-52ef-492c-9e2d-61f78b4d2787
02dd37de-1a1c-4aec-aa00-5e07275105b1
88081c3d-cfc6-4fc5-ac8d-6ae77395f2bb
e23e6b38-37f3-483f-a75c-fe459a60dd5c
67e5b0df-9c44-4719-bc69-b5fddac2fe63
cd38232d-1fa9-4f41-af67-6ee9fd3408db
016dd010-e7dd-4c85-b302-57ede3428464
b5cdfc83-775a-4163-a075-92d93853eeb8
d04e9b50-7b7d-4bbf-b266-6e19acf5bbf2
40b47bd1-abb2-4fd8-bb06-da3c72ac5737
257aaac9-041b-4643-9769-8c0c85da7736
\end{lstlisting}

\subsection{Questions and Answers}
\label{app:questions}

We provide all questions from both benchmarks.

\subsubsection{WikiCeleb100}

\begin{enumerate}[leftmargin=*, itemsep=0.2em]
  \item \textit{Which industry has the most artist debuts - Music, Film, or Content Creation?} \\ \textbf{Answer:} Music
  \item \textit{What are the top 3 countries of birth for artists in the dataset?} \\ \textbf{Answer:} United States, India, England
  \item \textit{Which artist has the earliest debut year in the dataset?} \\ \textbf{Answer:} Vera-Ellen
  \item \textit{Are there more artists who debuted in Music or Film?} \\ \textbf{Answer:} Music
  \item \textit{Which decade saw the most artist debuts?} \\ \textbf{Answer:} 2010.0
  \item \textit{What is the distribution of artists across different industries?} \\ \textbf{Answer:} Music, Film, Other, Content Creation
  \item \textit{Show the trend of artist debuts from earliest to latest decade - is it increasing, decreasing, or stable?} \\ \textbf{Answer:} increasing
  \item \textit{Which industry tends to have artists debuting earlier on average - Music or Film?} \\ \textbf{Answer:} Film
  \item \textit{What are the top 2 countries for artist births?} \\ \textbf{Answer:} United States, India
  \item \textit{Among artists born outside the US, which countries contribute the most?} \\ \textbf{Answer:} India, England, South Korea, Malaysia, Sweden
  \item \textit{Do more artists debut in the first half of a decade (years 0-4) or the second half (years 5-9)?} \\ \textbf{Answer:} First half
  \item \textit{For each of the top 5 countries, which industry is most dominant among their artists?} \\ \textbf{Answer:} United States, India, England, British India, South Korea
  \item \textit{Which artists have crossed over into multiple industries?} \\ \textbf{Answer:} Lesley Hornby, Olajide Olayinka Williams Olatunji, Wee Meng Chee, Zendaya Maree Stoermer Coleman, Aliaune Damala Bouga Time Puru Nacka Lu Lu Lu Badara Akon Thiam, ...
  \item \textit{Is the Music industry growing or declining over time compared to Film? Compare debuts by decade.} \\ \textbf{Answer:} declined marginally till 1980 and then grew steadily
  \item \textit{Among artists who work in multiple industries, is it more common to start in Music then move to Film, or vice versa?} \\ \textbf{Answer:} more
  \item \textit{What is the typical gap in years between an artist's Music debut and their Film debut?} \\ \textbf{Answer:} 4.6
  \item \textit{Which birth places have produced artists in the most diverse range of industries?} \\ \textbf{Answer:} Atlanta, Georgia, U.S., Muar, Johor, Malaysia, Oakland, California, Neasden, Middlesex, England, London, ...
  \item \textit{What proportion of artists from India work in Film compared to artists from the United States?} \\ \textbf{Answer:} more
  \item \textit{Who are the artists with the most recent debuts, and what industries are they in?} \\ \textbf{Answer:} Xavier James Trudeau, Ciara Mary-Alice Thompson, HorsegiirL, Darren Jason Watkins Jr., Victoria Beverley Walker
  \item \textit{Is there a trend showing Content Creation becoming more common in recent decades compared to traditional industries?} \\ \textbf{Answer:} Relatively more common in recent decades but still not as common as music or film
  \item \textit{Which artist debuted at the youngest age?} \\ \textbf{Answer:} Raven-Symon\'e Christina Pearman-Maday
  \item \textit{Which artist debuted at the youngest age in each industry?} \\ \textbf{Answer:} Film, Music, Content Creation, Other
\end{enumerate}

\subsubsection{FinQ100}

\begin{enumerate}[leftmargin=*, itemsep=0.2em]
  \item \textit{Which company has the highest long-term borrowings?} \\ \textbf{Answer:} CARMAX INC
  \item \textit{What are the top 5 companies by long-term borrowings?} \\ \textbf{Answer:} CARMAX INC, MICRON TECHNOLOGY INC, AUTOZONE INC, Salesforce, Inc., COSTCO WHOLESALE CORP NEW
  \item \textit{Are there more companies with zero long-term debt or with debt over 1 billion?} \\ \textbf{Answer:} debt\_category: Zero debt, more\_or\_less: more
  \item \textit{Which reporting period has the most company filings?} \\ \textbf{Answer:} 2025-10-31
  \item \textit{What are the companies with zero long-term debt?} \\ \textbf{Answer:} Apex Treasury Corp, BRB Foods Inc., CALERES INC, Cantor Equity Partners V, Inc., Dynamix Corp III, ...
  \item \textit{What is the distribution of companies across different debt ranges (0, 0-100M, 100M-1B, 1B+)?} \\ \textbf{Answer:} 1. Zero, 2. Under 100M, 3. 100M - 1B, 4. Over 1B
  \item \textit{How does the median long-term borrowings compare to the mean? Is the distribution skewed?} \\ \textbf{Answer:} Mean is more than median. Right skewed.
  \item \textit{What percentage of total long-term borrowings is held by the top 5 companies?} \\ \textbf{Answer:} 49.2
  \item \textit{Which companies have reported borrowings for multiple periods?} \\ \textbf{Answer:} IIOT-OXYS, Inc., TJX COMPANIES INC /DE/, VAIL RESORTS INC
  \item \textit{What is the range of long-term borrowings (min to max) and how spread out is the data?} \\ \textbf{Answer:} "min\_borrowings": 0.0, "max\_borrowings": 16586.77, "range\_spread": 16586.77, "std\_deviation": 2645.37
  \item \textit{Are companies reporting in October vs November showing different average borrowing levels?} \\ \textbf{Answer:} Borrowing levels in November are lower
  \item \textit{What is the ratio of companies with high debt ($>$1B) to companies with low/no debt ($<$100M)?} \\ \textbf{Answer:} 0.44
  \item \textit{If we exclude the top 3 largest borrowers, how does the average borrowing change?} \\ \textbf{Answer:} Decreases by 395.4
  \item \textit{Which companies have borrowings closest to the median value (within 25\% of median)?} \\ \textbf{Answer:} KESTRA MEDICAL TECHNOLOGIES, LTD., Oil-Dri Corp of America, SPORTSMAN'S WAREHOUSE HOLDINGS, INC.
  \item \textit{What is the cumulative percentage of total borrowings as we go from smallest to largest companies?} \\ \textbf{Answer:} CARMAX INC, MICRON TECHNOLOGY INC, AUTOZONE INC, Salesforce, Inc., COSTCO WHOLESALE CORP /NEW, ...
  \item \textit{Among companies with CIK numbers, which sectors (based on CIK ranges) have higher average borrowings?} \\ \textbf{Answer:} Mid registrants
  \item \textit{What is the total long-term borrowings across all companies in the dataset?} \\ \textbf{Answer:} 102590.76
  \item \textit{For TJX Companies which has multiple periods, did their long-term borrowings increase or decrease over time?} \\ \textbf{Answer:} decreased
  \item \textit{What percentage of companies have long-term borrowings under 50 million?} \\ \textbf{Answer:} 53.7
  \item \textit{How concentrated is the debt? What percentage of companies hold 80\% of the total borrowings?} \\ \textbf{Answer:} 15
  \item \textit{Are companies in the 'Recent registrants' CIK range (1M-1.5M) more likely to have zero debt compared to 'Mid registrants' (500K-1M)?} \\ \textbf{Answer:} Recent
  \item \textit{Among companies with borrowings over 1 billion, which has the lowest borrowings?} \\ \textbf{Answer:} ROSS STORES, INC.
  \item \textit{Do more companies have borrowings above or below 100 million?} \\ \textbf{Answer:} Below 100M
  \item \textit{What are the bottom 5 companies by long-term borrowings (excluding zero debt companies)?} \\ \textbf{Answer:} LIBERTY STAR URANIUM \& METALS CORP., OFF THE HOOK YS INC., USA OPPORTUNITY INCOME ONE, INC., Rise Gold Corp., IIOT-OXYS, Inc.
  \item \textit{Which company has the lowest non-zero long-term borrowings?} \\ \textbf{Answer:} LIBERTY STAR URANIUM \& METALS CORP.
\end{enumerate}

\subsection{SQL Program Complexity}
\label{app:sql_complexity}
\textbf{Complexity of writing SQL programs.} 
The SQL programs written by the Question Answering agent are non-trivial as shown in~\autoref{lst:amd-answer}.

\lstset{
  basicstyle=\ttfamily\tiny,
  frame=single,
  breaklines=true,
  columns=fullflexible
}
\begin{lstlisting}[language=SQL, caption={Computing depreciation-to-revenue margin for AMD},     label={lst:amd-answer}
]
SELECT
    d.depreciation_amortization_expense,
    t.total_revenue,
    ROUND(
        100.0 * d.depreciation_amortization_expense /
        t.total_revenue,
        2
    ) AS da_margin_percent
FROM DepreciationAmortizationExpense AS d
JOIN TotalRevenue AS t
    ON d.company_name = t.company_name
   AND d.fiscal_year  = t.fiscal_year
WHERE d.company_name = 'AMD'
  AND d.fiscal_year  = 2015
  AND d.depreciation_amortization_expense IS NOT NULL
  AND t.total_revenue IS NOT NULL
LIMIT 1;

\end{lstlisting}

\begin{lstlisting}[language=SQL, caption={An entity resolution SQL for inspection},     label={lst:entity-res}
]
SELECT
    product_or_service_name,
    COUNT(*) AS cnt
FROM CompanyProductOrService_2063d536_AMD_2022_10K
GROUP BY
    product_or_service_name
ORDER BY
    product_or_service_name,
    cnt DESC;
\end{lstlisting}

\section{Baselines}
\label{app:baseline_details}

\subsection{Baseline Implementations}
\label{app:baseline_implementations}

We compare against three families of baselines: retrieval-augmented methods, base LMs operating directly on the full corpus, and agentic scaffolding methods. All baselines use the same evaluation protocol (LLM-as-a-judge with partial scoring) as \system{}. Models referred to below are listed in Table~\ref{tab:results} and Table~\ref{tab:results2} of the main paper.

\paragraph{RAG (dense retrieval).}
Standard chunk-level dense retrieval. Documents are split into fixed-size chunks of 4k tokens and indexed with Qwen3-4B as the embedding model. At inference, the question is embedded and the top-$k$ chunks are retrieved by cosine similarity and concatenated as context for the GPT-4.1 reader, which generates the answer. 

\paragraph{LongRAG.}
Follows the original LongRAG recipe~\citep{jiang2024longrag}: longer chunks (16k tokens) are used so that fewer retrievals are needed and each retrieved unit carries more local context. Embedder Qwen3-4B; reader GPT-4.1. 

\paragraph{GraphRAG.}
Microsoft's graph-structured RAG~\citep{edge2024graphrag}. The corpus is processed through entity-relation extraction to build a knowledge graph, then community summaries are produced bottom-up. At query time, the system selects communities relevant to the question and uses their summaries as context. Embedder Qwen3-4B; reader GPT-4.1.

\paragraph{Base LMs.}
We pass the full corpus directly to the model in a single prompt. We test GPT-4.1 (1M-token context) and Qwen3.5-122B-A10B (1M-token context). No retrieval, summarization, or chunking is performed.

\paragraph{DocETL.}
Map-reduce extraction with per-question schemas~\citep{docetl}. The system generates a plan that maps over chunks (16k tokens) to extract structured fields, then reduces by aggregating across chunks. Worker and reasoning model: GPT-4.1.

\paragraph{Chain of Agents (CoA).}
Sequential chunk-by-chunk (16k tokens) summarization with a running-context buffer~\citep{zhang2024coa}. Each agent ingests one chunk plus the running summary from the previous agent and produces an updated summary; a final manager agent answers the question from the last summary. Worker model: GPT-4.1-mini; reasoning model: GPT-4.1 (manager).

\paragraph{RLM (Recursive Language Models).}
Recursive task decomposition~\citep{zhang2025recursivelanguagemodels}. The reasoning LM (GPT-5) decides at each step whether to read a sub-document directly or to delegate to a sub-call (its own recursive instance) on a smaller portion of the corpus; sub-calls return summarized findings to the parent. Worker model: GPT-5-mini. We use the authors' default recursion-depth and step-budget settings. We run RLM for maximum 30 iterations.

\subsection{RLM with GPT-4.1}
\label{app:rlm-gpt41}
To justify the model choice for RLM (GPT~5 / GPT~5-mini), we also ran RLM with GPT~4.1 (main) and GPT~4.1-mini (sub-LLM), matching SLIDERS' configuration. As shown in \autoref{tab:rlm_gpt41}, RLM performs worse with GPT~4.1, confirming that our main comparison gives RLM an advantage. SLIDERS with GPT~4.1 still outperforms RLM with GPT~5 across all benchmarks.

\begin{table}[h]
\centering
\caption{RLM performance with different base models.}
\label{tab:rlm_gpt41}
\begin{tabular}{lrr}
\toprule
\textbf{System} & \textbf{Loong} & \textbf{FinanceBench} \\
\midrule
RLM (GPT~5 \& GPT~5-mini) & 72.64 & 75.33 \\
RLM (GPT~4.1 \& GPT~4.1-mini) & 60.13 & 72.00 \\
\textbf{SLIDERS} (GPT~4.1 \& GPT~4.1-mini) & \textbf{78.57} & \textbf{89.67} \\
\bottomrule
\end{tabular}
\end{table}

\section{Additional Results and Analysis}
\label{app:additional_results}

\subsection{Cost Analysis}
\label{app:cost}

\autoref{tab:cost_breakdown} presents the average cost per question in US dollars for \system{} across different benchmarks. Costs vary based on document length, complexity, and the number of reconciliation operations required.

\begin{table}[h]
\centering
\caption{Average cost per question (in USD) across all benchmarks.}
\label{tab:cost_breakdown}
\begin{tabular}{lr}
\toprule
\textbf{Benchmark} & \textbf{Avg. Cost (\$)} \\
\midrule
Oolong            & 1.30 \\
Loong Legal       & 0.65 \\
Loong Finance (ZH) & 0.79 \\
Loong Finance (EN) & 0.60 \\
Loong Papers      & 0.37 \\
FinanceBench      & 0.63 \\
WikiCeleb100     & 0.60 \\
FinQ100          & 1.39 \\
\bottomrule
\end{tabular}
\end{table}

OoLong has the highest per-question cost due to its longer input contexts, while Loong Papers has the lowest cost with shorter academic papers requiring fewer reconciliation steps.

\begin{wraptable}{r
}{0.35\textwidth}
\vspace{-4mm}
\centering
\caption{WikiCeleb100 latency.}
\label{tab:graphrag_comparison}
\scriptsize
\begin{tabular}{lccc}
\toprule
 & \textbf{Cost} & \textbf{Time} & \textbf{Acc.} \\
\midrule
GraphRAG & \$182 & 2.3 hr & 48.59 \\
\system{} & \$13.10 & $\sim$16 min & 78.91 \\
\bottomrule
\end{tabular}
\vspace{-4mm}
\end{wraptable}
\paragraph{Cost and Latency Analysis.}

Across benchmarks, compared to RLM, an agent-scaffolding baseline, \system{} achieves equal or lower cost, averaging \$0.79 per question, and its cost is further amortized when the evidence table is used for followup questions. 
Most of the cost is due to canonicalization of evidence variables for reconciling the evidence table.
On FinQ100, reusing the evidence table substantially lowers marginal query cost and latency, the answer generator takes $\sim$25 seconds per question.
Table~\ref{tab:graphrag_comparison} shows the same trend on WikiCeleb100, where \system{} completes the extraction and reconciliation in $\sim$16 minutes at \$13.10, compared to 2.3 hours and \$182 for GraphRAG, while also achieving higher accuracy.
We compare to GraphRAG because it also prebuilds a graph for the given document corpus.

% ============================================================================
% SLIDERS System Prompts Appendix
% Auto-generated by generate_prompts_appendix.py
% ============================================================================

\begin{wraptable}{r}{0.50\textwidth}
\centering
\small
\vspace{-2em}
\caption{Accuracy across schema-induction models.
\textbf{Bold} indicates best per dataset; $\Delta$ is the range
(max $-$ min).}
\label{tab:schema_robustness}
\begin{tabular}{lcccc}
\toprule
\textbf{Dataset} & \textbf{4.1} & \textbf{4.1-mini} & \textbf{5} & $\Delta$ \\
\midrule
Loong Papers     & \textbf{91.30} & 89.96 & 88.00 & 3.30 \\
Loong Legal      & 64.12 & \textbf{68.34} & 61.26 & 7.08 \\
Loong Finance EN & \textbf{74.50} & 68.10 & 73.10 & 6.40 \\
Loong Finance ZH & \textbf{93.96} & 90.46 & 93.20 & 3.50 \\
\midrule
Loong Avg        & \textbf{80.97} & 79.22 & 78.89 & 2.08 \\
\midrule
FinanceBench     & 76.71  & \textbf{80.00}  & \textbf{80.00}  & 3.29  \\
\bottomrule
\end{tabular}
\end{wraptable}

\subsection{Reconciliation Correctness}
\label{app:recon_correctness}
WikiCeleb100 and FinQ100 have manually reconciled gold tables (Appendix~\ref{app:benchmark_construction}). We align the agent's final table with the gold table, matching rows that describe the same entity. Each row is a match (present in both tables), an extra row (only in the agent's table, a spurious row or an unmerged duplicate), or a missing row (only in the gold table, dropped or over-merged). Structural precision is the share of the agent's rows that match gold rows, structural recall is the share of gold rows the agent produces, and structural F1 combines the two. Value accuracy is the share of matched rows whose key value is exactly correct. Pick error is the share of matched rows where the correct value was among the extracted candidates but the agent chose another one.

\begin{table}[h]
\centering
\small
\caption{Correctness of reconciliation decisions against the gold reconciled tables.}
\label{tab:recon_correctness}
\begin{tabular}{lcc}
\toprule
\textbf{Metric} & \textbf{WikiCeleb100} & \textbf{FinQ100} \\
\midrule
Structural precision & 0.898 & 0.490 \\
Structural recall & 0.925 & 0.901 \\
Structural F1 & 0.911 & 0.635 \\
Value accuracy on matched rows (strict) & 49.6\% (61/123) & 65.0\% (65/100) \\
Pick error ($\downarrow$) & 12.2\% (15/123) & 15.0\% (15/100) \\
\bottomrule
\end{tabular}
\end{table}

The low precision on FinQ100 comes from imperfect canonicalization, where the agent keeps multiple rows for 16 entities. The low strict value accuracy on WikiCeleb100 comes from canonicalization errors between names and alternate names. In the remaining matched rows with a wrong value (47 of 123 on WikiCeleb100 and 20 of 100 on FinQ100), the correct value was not among the extracted candidates.

\subsection{Schema Induction Robustness}
\label{app:schema_robustness}
A natural concern is whether SLIDERS is sensitive to the choice of
schema-induction model. We generate schemas with three models of
varying capability (GPT-4.1-mini, GPT-4.1, GPT-5) on Loong and
FinanceBench. The resulting schemas differ substantially: GPT-4.1
averages 1.0 tables and 3.3 fields per question vs.\ GPT-5 with
1.54 tables and 13.3 fields, a $4\times$ complexity gap. Yet
downstream accuracy remains stable
(Table~\ref{tab:schema_robustness}): Loong average accuracy spans
just 2.1 points across all three models, and FinanceBench stays
within 3.3 points, demonstrating that schema induction is not a
fragile bottleneck. One caveat is that more complex schemas can
increase reconciliation difficulty, post-reconciliation ranges
widen modestly on Loong, suggesting diminishing returns from
overly detailed schemas.

\begin{figure}[t]
\centering
\begin{minipage}[b]{0.32\linewidth}
\centering
\begin{tikzpicture}
\begin{axis}[
    width=\linewidth,
    height=4cm,
    xbar stacked,
    bar width=4pt,
    xlabel={\% Reconciliation Errors},
    symbolic y coords={Legal,Papers,WC,FB,Fin-ZH,Fin-EN,FQ,BC+},
    ytick=data,
    enlarge y limits=0.12,
    xmajorgrids=true,
    xmin=0,
    xmax=100,
    xtick={0,50,100},
    xticklabel={\pgfmathprintnumber{\tick}\%},
    tick label style={font=\scriptsize},
    label style={font=\scriptsize},
    legend style={
        font=\tiny,
        at={(0.5, 1.12)},
        anchor=south,
        legend columns=3,
        draw=none,
        /tikz/every even column/.append style={column sep=2pt}
    }
]
\addplot[fill=opBlue, draw=opBlue!70!black] coordinates {
    (20.00,Legal)
    (100.00,Papers)
    (20.83,WC)
    (47.30,FB)
    (39.24,Fin-ZH)
    (36.67,Fin-EN)
    (32.29,FQ)
    (56.35,BC+)
};
\addplot[fill=opGreen, draw=opGreen!70!black] coordinates {
    (0.00,Legal)
    (0.00,Papers)
    (29.17,WC)
    (5.41,FB)
    (0.00,Fin-ZH)
    (0.00,Fin-EN)
    (0.00,FQ)
    (1.15,BC+)
};
\addplot[fill=opRed, draw=opRed!70!black] coordinates {
    (80.00,Legal)
    (0.00,Papers)
    (50.00,WC)
    (47.30,FB)
    (60.76,Fin-ZH)
    (63.33,Fin-EN)
    (67.71,FQ)
    (42.50,BC+)
};
\legend{Dedup., Consol., Resolve Conflict}
\end{axis}
\end{tikzpicture}
\captionsetup{justification=raggedright}
\caption{Reconciliation issues found by \system{}.}
\label{fig:operations}
\end{minipage}
\hfill
\begin{minipage}[b]{0.65\linewidth}
\centering
\begin{subfigure}[b]{0.46\linewidth}
\centering
\begin{tikzpicture}
\begin{axis}[
    title style={font=\large, yshift=0.5ex},
    ybar,
    single ybar legend, 
    bar width=0.08cm,
    width=\linewidth,
    height=3cm,
    ylabel={Acc (\%)},
    ylabel style={font=\scriptsize},
    ymin=0, ymax=100,
    ytick={0, 25, 50, 75, 100},
    xtick={1, 2, 3, 4},
    xticklabels={{Spotlight}, Comparison, Clustering, {Chain of Reas.}},
    x tick label style={font=\tiny, rotate=35, anchor=north east, align=right, inner sep=1pt},
    xmin=0.5, xmax=4.5,
    ymajorgrids=true,
    grid style={gray!30},
    tick style={draw=none},
    legend style={
        at={(0.5, 1.12)},
        anchor=south,
        legend columns=4,
        font=\tiny,
        draw=none,
        /tikz/every even column/.append style={column sep=0.1cm},
    },
    legend cell align={left},
]
\addplot[fill=blue!70, draw=none] coordinates {
    (1, 100) (2, 76) (3, 61) (4, 39)
};
\addplot[fill=red!70, draw=none] coordinates {
    (1, 100) (2, 83) (3, 88) (4, 86)
};
\addplot[fill=green!60!black, draw=none] coordinates {
    (1, 50) (2, 58) (3, 73) (4, 57)
};
\addplot[fill=orange!80, draw=none] coordinates {
    (1, 0) (2, 0) (3, 90) (4, 95)
};
\legend{Fin-EN, Fin-ZH, Legal, Papers}
\end{axis}
\end{tikzpicture}
\vspace{-2mm}
\caption{Loong}
\label{fig:loong}
\end{subfigure}
\hfill
\begin{subfigure}[b]{0.26\linewidth}
\centering
\begin{tikzpicture}
\begin{axis}[
    title style={font=\large, yshift=0.5ex},
    ybar,
    bar width=0.18cm,
    width=\linewidth,
    height=3.5cm,
    ylabel={Acc (\%)},
    ylabel style={font=\scriptsize},
    ymin=0, ymax=100,
    ytick={0, 25, 50, 75, 100},
    xtick={1, 2, 3},
    xticklabels={Domain-Rel., Metrics-Gen., Novel-Gen.},
    x tick label style={font=\tiny, rotate=45, anchor=north east, inner sep=1pt},
    xmin=0.4, xmax=3.6,
    ymajorgrids=true,
    grid style={gray!30},
    tick style={draw=none},
]
\addplot[fill=sliderscolor, draw=none] coordinates {
    (1, 86) (2, 96) (3, 86)
};
\end{axis}
\end{tikzpicture}
\caption{FinanceBench}
\label{fig:financebench}
\end{subfigure}
\hfill
\begin{subfigure}[b]{0.26\linewidth}
\centering
\begin{tikzpicture}
\begin{axis}[
    title style={font=\large, yshift=0.5ex},
    ybar,
    bar width=0.18cm,
    width=\linewidth,
    height=3.5cm,
    ylabel={Acc (\%)},
    ylabel style={font=\scriptsize},
    ymin=0, ymax=100,
    ytick={0, 25, 50, 75, 100},
    xtick={1, 2, 3},
    xticklabels={Counting, User, Timeline},
    x tick label style={font=\tiny, rotate=45, anchor=north east, inner sep=1pt},
    xmin=0.4, xmax=3.6,
    ymajorgrids=true,
    grid style={gray!30},
    tick style={draw=none},
]
\addplot[fill=sliderscolor, draw=none] coordinates {
    (1, 60.3) (2, 80.0) (3, 100.0)
};
\end{axis}
\end{tikzpicture}
\caption{Oolong}
\label{fig:oolong}
\end{subfigure}
\caption{\system{}'s accuracy across difficulty levels, domains and question types on different benchmarks.}
\label{fig:benchmark-breakdown}
\end{minipage}\par
\vspace{-4mm}
\end{figure}

\subsection{Relevance Gate Analysis}
\label{app:relevance_gate}
We evaluate the relevance gate on 20 sampled incorrect predictions across benchmarks. Across 516 chunks, 282 were rejected by the relevance gate, with only 1 false negative, yielding a false-negative rate of 0.4\%. We note that this estimate is derived from a limited sample of pipeline-level failures and may not be fully representative of the gate's overall accuracy across the corpora and leave it for future analysis. This indicates that the relevance gate is not a major source of error; most failures arise downstream (e.g., schema mismatch or reconciliation).

\subsection{Multi-Document Summarization Case Study}
\label{app:multidoc_summarization}

To demonstrate generality beyond QA, we ran \system{} with the prompt ``Summarise the given research papers'' on three ML papers from the Loong benchmark.
The system induced a 5-table schema: PaperSummary (title, authors, abstract), PaperContributions, PaperMethods, PaperFindings, and PaperConclusions, each with paper\_title as linking key and free-text fields.
It extracted 117 records across 3 papers, then aggregated them into coherent per-paper summaries.
Because extraction runs per-chunk rather than over the full corpus, it avoids information loss from context-window pressure seen in direct LLM baselines.
This case study suggests that \system may extend beyond factoid or tabular QA to more open-ended synthesis tasks.

\subsection{Issues Discovered by \system{}}
\label{app:errors}
During error analysis, we identified cases where the ground truth annotations in FinanceBench and Loong are incorrect.
Because \system{} traces provenance, we can verify whether a predicted answer is correct or the ground truth itself is erroneous.
We identified two types of errors: (1)~answers that are spread across multiple pages, and (2)~incorrect or incomplete annotations.
For FinanceBench, which assumes answers are localized to a single page, we found more complete answers spanning multiple pages for question IDs \texttt{financebench\_id\_02416, financebench\_id\_00283}.
For Loong in the financial domain, we found ground truth entries with missing units and incorrect values; for example, for a question the annotation says the accounts payable is 1,740 instead of the correct value of 1,663 thousand. 
These findings demonstrate the auditability and interpretability of \system{}.

\begin{table}[h]
\centering
\small
\caption{Effect of reconciliation agent on the number of rows per primary key group of evidence table.}
\label{tab:context_funnel}
\begin{tabular}{lrrrr}
\toprule
\textbf{Benchmark} & \textbf{Raw tokens} & \textbf{Pre} & \textbf{Post} & \textbf{Comp.} \\
\midrule
FinanceBench & $\le$363K & 1.72 & 1.04 & 1.65$\times$ \\
Loong (avg)  & $\le$300K & 1.41 & 1.04 & 1.36$\times$ \\
Oolong       & 256k      & 1.02 & 1.01 & 1.00$\times$ \\
WikiCeleb100 & 3.9M      & 2.04 & 1.03 & 1.98$\times$ \\
BrowseComp+  & 6M--11M   & 2.66 & 1.10 & 1.32$\times$ \\
FinQ100      & 36M       & 5.36 & 1.19 & 4.51$\times$ \\
\bottomrule
\end{tabular}
\label{tab:context-funnel}
\end{table}

\section{Limitations}
\label{app:limitations}

\textbf{Evidence variable dependence.} The induced evidence variable determines what evidence is recoverable at answer time; questions whose evidence does not factor cleanly into typed fields fall back on free-text columns. The variables are induced from a sample of documents. If the sample misses a kind of document, the variables can be incomplete, and evidence the question needs may not be extracted. 

\textbf{Extraction completeness.} The relevance gate trades a small recall loss for a large precision gain, but rare or implicit evidence the gate underweights could be silently dropped. 

\textbf{Reconciliation errors.} Aggressive canonicalization can collapse genuinely distinct entities, particularly in domains with high-cardinality keys or near-duplicate naming conventions.

\textbf{Latency.} The pipeline requires multiple sequential LLM calls, resulting in higher latency (2-3 min end-to-end) than single-call baselines; this is acceptable for accuracy-critical workflows but not for real-time applications.

\textbf{Evaluation. }Evaluation relies on LLM-as-a-judge, which we validate with human annotations (Cohen's $\kappa = 0.758$), but residual noise remains.
\system{} achieves 55\% on FinQ100, insufficient for full automation in high-stakes financial analysis.
For safety-critical deployments, we recommend human-in-the-loop verification: \system{}'s provenance tracking enables reviewers to efficiently validate each extraction against source documents. 

\textbf{Provenance correctness. } When provenance is incorrect (e.g., wrong text span), the reconciliation agent may make faulty decisions; our verification of 410 facts found 99.03\% correct, and users can inspect source quotes directly.

\textbf{Human in the Loop. }SLIDERS is an implementation of the evidence synthesis pipeline, not a full interactive review platform. The current system produces auditable tables and provenance, but does not yet provide a user interface for editing variables or reconciliation decisions. Building such a human-in-the-loop interface is future work.

\section{Prompts}
\label{app:prompts}

\subsection{System Prompts}
\label{app:prompts_system}

We provide all prompts used in the \system{} pipeline, ordered by their execution sequence. 
Prompts are displayed in a monospace font with syntax highlighting disabled for readability.

\subsubsection{Schema Generation}

\paragraph{Schema Induction}
Generates a structured schema (tables and fields) based on the question and document descriptions.

\begin{lstlisting}[
  basicstyle=\ttfamily\tiny,
  breaklines=true,
  breakatwhitespace=false,
  frame=single,
  xleftmargin=0.3cm,
  xrightmargin=0.3cm
]
# instruction
Given description of documents, sample document content, and a question, define the tables and fields (including intermediate/supporting fields) that, once extracted at the page level and merged later, will let us answer the question.

Table Modeling rules:
- One table = one relationship. If the relationship logically decomposes (e.g., case summary vs per-counsel arguments), define multiple tables.
- No nested types. When you would need a list/array, model it as multiple rows in the appropriate table.
- Include supporting fields. Add any inputs required for downstream computation/validation (e.g., numerators/denominators, period markers, currency).
- Normalize by design. Each numeric field must declare a single unit and scale; extractors will normalize values into that target.

Field Specs (required keys per field):
- name: string (concise, machine-friendly).
- data_type: one of str | int | float | decimal | bool | date | enum.
- Prefer decimal for money/precise quantities; date for dates.
- enum_values: list of strings (required iff data_type == "enum").
- unit: string or None (e.g., USD, %, shares; use None if not applicable).
- scale: one of none | thousands | millions | billions | trillions | basis_points | log10 | ln.
    - Values will be stored in this scale.
    - Example: $1.23B with scale="millions" $\rightarrow$ store 1230, unit="USD".
- description: self-contained extraction instruction (what the value represents; common surface forms/synonyms).
{% if enable_extraction_guidelines %}- extraction_guideline: string or None (optional but recommended). A concrete instruction for the extractor on HOW to extract this field.
    - Use this to clarify extraction logic that isn't obvious from the description alone.
    - Specify the SOURCE of the value (e.g., which document/entity the field refers to).
    - Clarify SCOPE constraints (e.g., what to include/exclude, how to handle ambiguity).
    - Provide FORMAT guidance (e.g., normalization rules, handling incomplete data).
    - Keep it concise (1-2 sentences max), specific to this field, and actionable.
{% endif %}- normalization (object; optional keys as needed):
    - currency: target code (e.g., "USD") or "as_is".
    - percent: "0_100" or "0_1".
    - date_format: e.g., "YYYY-MM-DD".
- required: boolean (expected presence; still None if missing on a page).

General Rules:
- Each table represents a relationship. If modeling the relationship requires multiple relations, define multiple tables (e.g., a "CaseSummary" table for case-level facts and a "CounselArguments" table for per-counsel arguments).
- No nested types. If the extraction needs multiple items (e.g., many arguments), the extractor will create multiple rows in the appropriate relationship table.
- Include any linking fields needed to relate tables later (e.g., case_title, docket_number, entity_name, period_end).
- Normalize all values as specified below so independently run extractors produce standardized outputs.
- If a value is missing/unknown, set the value to None (not 0, empty string, or placeholder text).
- Page/section metadata is precomputed; do not include it in fields.
- The schema must describe only document-grounded relationships and fields, not the question itself. Do not add fields that simply restate the question.

Output:
- Emit JSON.
- For each table: include name, description, and a fields array.
- Do not include page metadata (it's precomputed).

# input
# Question
{{ question }}

# Documents description
{{ document_description }}
{% if document_text %}

# Sample Document Content
{{ document_text }}
{% endif %}

{% if guidelines %}
{{guidelines}}
{% endif %}

These are reserved keywords for fields. DO NOT USE THEM: `row_id`, `document_name`, `metadata`.
\end{lstlisting}

\subsubsection{Information Extraction}

\paragraph{Chunk Relevance Filtering}
Determines whether a document chunk contains information relevant to answering the question.

\begin{lstlisting}[
  basicstyle=\ttfamily\tiny,
  breaklines=true,
  breakatwhitespace=false,
  frame=single,
  xleftmargin=0.3cm,
  xrightmargin=0.3cm
]
# instruction
You will be given a single page from a larger document along with a question. The extraction will be done on this page.

Your task is to determine whether this page contains ANY information that could be useful for answering the question once this page is combined with other pages later.

Treat the page as RELEVANT if it satisfies ANY of the following:
- It contains a direct or partial answer to the question.
- It contains definitions, conditions, steps, list items, examples, or descriptions that relate to the question.
- It narrows down possibilities or provides context that a reconciliation agent could use (e.g., "the first principle is...", "see Section 3 for...", "Step 2: ...").

Treat the page as IRRELEVANT only if:
- None of the content on the page has any meaningful connection to the question, and
- It only contains metadata values, and doesn't explictly mention useful fields. For example, if the page only contains the title, company name, then ignore this page.
- There is no text, that could help answer the question in any way during later aggregation.

First, briefly explain your reasoning in natural language. Then say if its relevant or not.

# input
# Question
{{question}}

# Document page
{{document}}
\end{lstlisting}

\paragraph{Table Extraction}
Extracts structured data from document chunks according to the generated schema.

\begin{lstlisting}[
  basicstyle=\ttfamily\tiny,
  breaklines=true,
  breakatwhitespace=false,
  frame=single,
  xleftmargin=0.3cm,
  xrightmargin=0.3cm
]
# instruction
Given a schema and a single page, emit rows per table using the schema's field names/types---and per-field evidence:
- value (normalized to the field's unit/scale). If scale is None and the page content says 1.3 billion, then write the value as 1300000000.
{% if extract_quotes %}
- quote (largest contiguous exact text; list of strings or None)
{% else %}
- quote (always null; do not extract snippets when quotes are disabled)
{% endif %}
- rationale (why this value; include table headers/context if from a table)
- is_explicit (boolean)
- confidence (string; one of "Very High", "High", "Medium", "Low", "Very Low")

Confidence Rubric:
- Very High
  - The value is explicitly stated verbatim or via a trivial transformation (e.g., unit normalization) in the current text span, unambiguously refers to the target entity, and fully satisfies the schema field definition.
  - Surrounding context confirms the interpretation without inference.
- High
  - The value is explicitly stated, but either (a) requires light interpretation to map to the schema (e.g., paraphrase, implicit unit), or (b) could plausibly appear elsewhere in the document, even though the current span provides sufficient evidence.
  - Entity reference and schema alignment are still clear.
- Medium
  - The value is not explicitly stated but can be reasonably inferred from the text through interpretation, aggregation, or implicit assumptions.
  - The match to the schema is partial or approximate, and alternative interpretations are plausible.
- Low
  - The value is a speculative guess derived from weak textual cues, heuristics, or world knowledge.
  - Evidence is indirect, incomplete, or ambiguous, and the extractor believes the value is likely incorrect.
  - For entity relationships: Inferred from abbreviated references (acronyms, short names, citation numbers) without the full entity name appearing on the page.
- Very Low
  - No supporting evidence exists in the text.
  - The value is a placeholder, default, or fabrication inserted to satisfy schema completeness and should not be trusted or used downstream.

{% if enable_verbalization_instructions %}
## Critical: Exhaustive List Processing

**APPLIES TO:** Any section containing a numbered or bulleted list of related entities, including but not limited to:
- Sections titled "References", "Bibliography", "Citations", "Works Cited", "Related Work", "Dependencies", "Components", "Exhibits", "Appendices"
- Dependency lists, package catalogs, component inventories, related entity lists
- Any enumerated list where entities might be mentioned

**REQUIRED PROCESS:**
1. **In `verbalization` field**: List ALL target entities from the question you're looking for
   - **CRITICAL**: Include EVERY entity mentioned ANYWHERE in the question---this includes the primary entity being analyzed AND all comparison/related entities
   - Example: If the question asks about relationships involving entities X, Y, and Z, then your target entities list should be [X, Y, Z]---NOT just a subset
   - When processing ANY document, check for ALL entities from the question, regardless of which document you're currently analyzing
2. **Identify the list boundaries** - Find where the list starts and ends
3. **Process EVERY entry sequentially** - Go through each numbered/bulleted item one by one (e.g., [1], [2], [3]... to the last entry)
4. **In `verbalization` field**: For EACH AND EVERY entry, you MUST document:
   - **EXPLICITLY LIST EACH ENTRY** - Write: "Entry [N] 'identifier/name found here' - MATCH to <target entity name>" OR "Entry [N] 'identifier/name found here' - no match"
   - **DO NOT SUMMARIZE** - List each entry individually even if there are 100+ entries
   - **CRITICAL**: Even if you don't see matches initially, you MUST list every single entry. Saying "Checked entries [1]-[92], none match" or "The section lists many entries but none match" is WRONG - you must list each entry explicitly
5. **In `tables` field**: Extract a row immediately when you find a match (High/Very High confidence, with concise rationale)
   - **CRITICAL**: For EVERY entry you marked as "MATCH" in your verbalization, you MUST create a corresponding extraction row. The verbalization is your working notes; the tables section is where you record the actual data. DO NOT identify matches in verbalization and then fail to extract them.
6. **Continue to the end** - Do not stop after finding matches; process every remaining entry
7. **In `verbalization` field**: Confirm you reached and checked the final entry (e.g., "Completed all 92 entries")
8. **CRITICAL - Understanding Entity Relationships in Lists**: When processing a source entity's list of related entities:
   - Finding target entity Y in source entity X's list means "X relates to Y" - EXTRACT THIS
   - Whether Y appears in the question is IRRELEVANT to whether this is a valid relationship
   - ONLY skip extraction when source equals target (i.e., X's list contains X itself - a self-loop, likely a data error)
   - [OK] Correct: Processing "Entity Alpha"'s related items list, found "Entity Beta" (where Beta is mentioned in the question) $\rightarrow$ Extract "Alpha relates to Beta"
   - [OK] Correct: Processing "Component A"'s dependency list, found "Component B" (from question) $\rightarrow$ Extract the dependency relationship
   - [OK] Correct: Processing "Site 001"'s supplier catalog, found "Supplier Corp" (from question) $\rightarrow$ Extract the supplier relationship
   - [X] Incorrect: Processing "Item X"'s related items list, found "Item X" itself $\rightarrow$ Skip (self-loop)
   - **WRONG reasoning**: "The found entity is mentioned in my question, therefore this is self-referential" - NO. Self-reference means the source entity points to itself, not that the target appears in the question.

Examples: 
- If you see a section with 50 entries (e.g., a list with [1] through [50]), and the question asks about 5 specific entities:
  - [OK] Correct in `verbalization`: "Target entities: [Entity A, Entity B, Entity C, Entity D, Entity E]. Entry [1] 'Author et al., 2020. Some description about topic X' - no match. Entry [2] 'Source, 2021. Entity A: A comprehensive analysis' - MATCH to Entity A. Entry [3] 'Creator, 2019. Another description' - no match. ... Entry [50] 'Final Author, 2023. Final description' - no match. Completed all 50 entries."
  - [OK] Correct in `tables`: Extract rows for all matches found
- If you see an exhibits list with 15 items and question asks about 3 companies:
  - [OK] Correct in `verbalization`: "Target entities: [Acme Corp, Widget Inc, Gadget LLC]. Entry [1] 'Exhibit 1.1: Lease Agreement with Property Owner' - no match. Entry [2] 'Exhibit 2.1: Supply Contract with Acme Corp' - MATCH to Acme Corp. ... Entry [15] 'Exhibit 5.4: Insurance Policy' - no match. Completed all 15 entries."
- [X] Wrong: "The section lists many entries, but none of the target entities appear explicitly" (This is a SUMMARY without listing entries)
- [X] Wrong: "Checked entries [1]-[50]. Found matches for Entity A and Entity B." (This is a SUMMARY, not entry-by-entry verbalization)
- [X] Wrong: Stop after finding 2 matches in first 10 entries
- [X] Wrong: "List entries [1]-[92]. None match." (Must list each entry explicitly: "[1] ... - no match. [2] ... - no match.")
{% endif %}

Output guidelines:
- Produce one row per real item visible on the page (e.g., per counsel-argument, per entity-period-metric).
- Strictly follow the schema, do not add any new field/key.
- If the page contains multiple distinct items, output multiple rows.
{% if extract_quotes %}
- For any field, if the given page does not have any relevant information, set the value and quote to null.
{% else %}
- For any field with no relevant information, set value to null. Quote extraction is disabled, so leave `"quote": null` for every field.
{% endif %}
- Consider the summary if provided to contextualize the extraction.
- IMPORTANT: If the question below contains a "FOCUS:" directive specifying a particular row or item to extract, you MUST extract ONLY that specific row/item. Ignore all other rows/items in the page. This focused extraction is used when processing information-dense pages to ensure accuracy and prevent confusion between similar items.

Extraction and Normalization Rules:
- Page-local only. Use only the current page's content.
- Rationale.
    - Start with rationale, and reason about which parts of the text correspond to the field 
    - 1--3 sentences: how you located/parsed it; if from a table, include headers and relevant table context.
    - Note any normalization you performed (e.g., "converted $1.23B to 1230 millions USD").
- Normalization is mandatory.
    - Convert magnitudes into the field's target scale (e.g., target millions: 1,200,000$\rightarrow$1.2; 1.23B$\rightarrow$1230).
    - Convert unit (e.g., currency) to the target if specified; else keep as_is.
    - Normalize percent per the field's rule (0_100 vs 0_1) and date per date_format.
- Type safety. Coerce to the declared data_type; on failure, set value = null.
{% if extract_quotes %}
- Quote (evidence).
    - Provide a list with the largest contiguous exact substring that supports the value (often a single element).
    - Preserve punctuation/case; no ellipses or paraphrase.
    - If truly not present on the page $\rightarrow$ quote = null.
    - Remember if quote is not present, then the value should also be None.
{% else %}
- Quote (evidence) is disabled. Always emit `"quote": null` for every field; do not copy supporting text into the output.
{% endif %}
- is_explicit.
    - True if the underlying value appears explicitly on the page (including trivial deterministic normalization like unit/scale conversion or parentheses-to-negative).
    - False if it requires arithmetic beyond normalization (e.g., sums, differences, ratios) or approximations/inference.
- Duplicates vs distinct items.
    - Multiple mentions of the same item: choose the most precise/authoritative instance on that page (e.g., structured table over prose).
    - Distinct items: separate rows.
- Relationship schemas with main entity:
    - If the schema has a primary entity field (e.g., "entity_name", "source_identifier"), populate it consistently in every row.
    - For relationship fields: never leave the primary/source entity field null. Only set secondary/target fields to null when no such relationship is present.
    {% if enable_verbalization_instructions %}
    - **ACTION REQUIRED for structured list sections**: When you encounter enumerated lists containing entity relationships (such as dependency catalogs, related entity lists, component inventories, reference lists, or bibliographic entries), follow the "Critical: Exhaustive List Processing" instructions above. This includes listing all target entities, processing every entry individually without summarization, and understanding that finding a question entity in another document's list is a valid relationship to extract.{% endif %}

# Example
Suppose schema has metric_value (decimal, unit=USD, scale=None) and the page says: "Revenue was $1.23 billion in 2023."

```JSON
{
  "verbalization": "Not a references section. Extracting direct field values from page content.",
  "tables": [
    {
      "name": "Financials",
      "rows": [
        {
          "fields": {
            "entity_name": {
              "rationale": "The page header mentions the information hence I can use it to extract the entity name. Taken from page header.",
              "value": "Acme Inc.",
              "quote": ["ACME INC."],
              "is_explicit": true,
              "confidence": "Very High"
            },
            "period_end": {
              "rationale": "The period end is mentioned but it is not in the required format. I need to normalized to YYYY-MM-DD. The period was mentioned in the page title",
              "value": "2023-12-31",
              "quote": ["For the year ended December 31, 2023"],
              "is_explicit": true,
              "confidence": "Very High"
            },
            "metric_name": {
              "rationale": "Table 2 mentions the metrics. The Metric label in the caption of the table talking about balance sheet.",
              "value": "revenue",
              "quote": ["Revenue"],
              "is_explicit": true,
              "confidence": "Very High"
            },
            "metric_forecast": {
              rationale: "The text mentions the current and past revenue. However, there is no mention of future revenue. Hence I should set the value and quote to None.,
              "value": null,
              "quote": null,
              "is_explicit": false,
              "confidence": "Very Low"
            },
            "metric_value": {
              "rationale": "The revenue is mentioned in in section Commentry. However, it says $1.23 billion. Since the scale is None, I should make it 1230000000. Converted $1.23B to 1230000000 with scale='None', unit='USD'.",
              "value": 1230,
              "quote": ["Revenue was $1.23 billion in 2023"],
              "is_explicit": true,
              "confidence": "Very High"
            }
          }
        }
      ]
    }
  ]
}
```

# input
The following question has been asked by the user over the whole document set:
<Question Start>
{{question}}
<Question End>

The Relationship Schema for this question has already been defined.
Your task is to populate this schema using only the information available in the document. The extracted data will later be aggregated with information from other pages to generate the final answer.

{% if previous_chunk_summary %}
# Summary of the pages till now
{{previous_chunk_summary}}
{% endif %}

<Document Start>
# The name of the CURRENT DOCUMENT you are extracting from is:
{{document_name}}

# Document Descriptions
{{document_description}}

# Page Content
{{document}}
<Document End>

# Relationship Schema
{{schema}}

# Field Extraction Guidelines
{% if extraction_guidelines %}
The schema includes specific extraction guidelines for certain fields. Follow these instructions when extracting:

{{extraction_guidelines}}
{% endif %}

# Output Format

Your output should be a JSON object with these fields:
{% if enable_verbalization_instructions %}
**IMPORTANT**: Verbalization is your working notes to show your reasoning. The `tables` array is where you record your actual extractions. Any match you identify in verbalization MUST have a corresponding row in tables.

1. **verbalization** (optional string): For sections with enumerated lists (such as dependency lists, related entity lists, "References", "Bibliography", "Citations", component inventories, or any numbered/bulleted entity listings), describe your systematic checking process:
   - List ALL entities mentioned ANYWHERE in the question---include the primary entity AND all comparison/related entities (e.g., if the question asks about relationships involving entities X, Y, and Z, list [X, Y, Z])
   - For each entry in the list: entry number/identifier, name/title found, match determination (which target entity it matches, or "no match")
   - **DO NOT SUMMARIZE**: You must list each entry individually, even if none match. Saying "checked all entries, none match" is insufficient.
   - Confirm you reached the final entry
   
   **IMPORTANT for inline references (non-list sections):**
   - When processing abbreviated inline identifiers (e.g., short codes, numeric references), DO NOT verbalize speculative mappings to full entity names.
   - Only report matches when the entity name itself appears on the page.
   - [X] WRONG: "The document mentions identifier X in context Y, which corresponds to 'Full Entity Name Z'"
   - [OK] CORRECT: "The document mentions identifier X but does not include the full entity name - no match"
   
   Examples: 
   - "Target entities: ['Entity A Title', 'Entity B Title', 'Entity C Title']. Reference list section entries [1]-[25]. Entry [1] 'ID-001. Item X description' - no match. Entry [2] 'ID-002. Entity A Title: A comprehensive analysis' - MATCH to 'Entity A Title'. Entry [3] 'ID-003. Another item description' - no match. ... Entry [25] 'ID-025. Final item' - no match. Completed all 25 entries."
   - "Target entities: ['Supplier A', 'Supplier B', 'Supplier C']. Dependencies section entries [1]-[18]. Entry [1] 'Component X from Vendor Alpha' - no match. Entry [2] 'Raw materials from Supplier A' - MATCH to 'Supplier A'. Entry [3] 'Packaging from Provider Beta' - no match. ... Entry [18] 'Logistics from Carrier Gamma' - no match. Completed all 18 entries."

2. **tables** (required array): The extracted data following the schema above, with concise rationales per field.
   
   **CRITICAL**: If you identified ANY matches (marked as "MATCH") in your verbalization, you MUST create a corresponding extraction row in the tables array for EACH match. The verbalization is your working notes---the tables section is where you record the actual extractions. Do NOT skip creating rows for matches you identified during verbalization.
{% else %}
1. **tables** (required array): The extracted data following the schema above, with concise rationales per field.
{% endif %}

Fill out the schema above based on the Document content.
\end{lstlisting}

\subsubsection{Data Reconciliation}
\ref{app:data_reconciliation_prompt}

\paragraph{Primary Key Selection}
Selects primary key fields which are used to group data rows for further reconciliation.

\begin{lstlisting}[
  basicstyle=\ttfamily\tiny,
  breaklines=true,
  breakatwhitespace=false,
  frame=single,
  xleftmargin=0.3cm,
  xrightmargin=0.3cm
]
# instruction
You are selecting **primary keys** for database tables that have been automatically extracted from documents. 

## Purpose
Your selected primary key will be used for **downstream data consolidation**:
1. **Deduplication**: Rows with the same primary key will be identified as duplicates
2. **Conflict Resolution**: When duplicates exist, we'll merge them by selecting the best value for each field
3. **Table Consolidation**: The final cleaned table will have one row per unique primary key

## Context
- The data was automatically extracted from unstructured documents (PDFs, text, etc.)
- Extraction errors create duplicate rows for the same real-world observation
- Multiple extractions of the same data point may have conflicting values
- Your primary key defines which rows represent the "same thing" and should be merged

## Your Task
Select the **conceptual primary key** that identifies unique observations in the ideal cleaned table. When two rows have the same primary key, they should represent the same real-world observation and be consolidated into one row.

## Key Principles

### The Key Defines What Gets Consolidated
Your primary key determines which rows will be merged together:
- Rows with **identical primary key values** $\rightarrow$ consolidated into ONE row
- Rows with **different primary key values** $\rightarrow$ kept as separate rows
- During consolidation, we merge conflicting field values by selecting the most reliable extraction

### Separate Identifiers from Values
**Identifiers** define WHAT is being observed - these form the primary key:
- Entity identifiers: company, user, product, document
- Temporal identifiers: date, period, timestamp
- Categorical identifiers: metric_type, event_type, category
- Relationship identifiers: source_entity, target_entity

**Values** are the observed data - these get consolidated during conflict resolution:
- Measurements: amounts, counts, percentages, scores
- Content: text, descriptions, labels, classifications
- Attributes: status, flags, properties

**Rule**: Primary keys should contain ONLY identifiers, never measured values.

Examples:
- Metrics over time: Key = entity + time_period (metric values get consolidated)
- Events with content: Key = entity + timestamp (content fields get consolidated)
- Entity relationships: Key = source + target (the relationship itself IS the identifier)

### Field Preferences
**STRONGLY PREFER business/domain identifier fields** as primary keys.

**Use `row_id` or `document_name` as the primary key in the scenarios where other fields are allowed to have duplicates as in text that represents dataset.**

## Analysis Guidelines

### 1. Understand the Data Domain
Ask yourself: What is this table measuring or recording?
- Financial data: Observations about companies at points in time
- Event logs: Actions by users at timestamps
- Relationships: Connections between entities
- Classifications: Labels applied to items

### 2. Identify the Identifier Fields
Which fields define WHAT is being measured (not the measurement itself)?
- Entity identifiers: company_name, user_id, paper_title, product_id
- Temporal identifiers: date, period_end, timestamp, fiscal_year
- Categorical identifiers: metric_name, event_type, category
- Relationship identifiers: source_entity, target_entity

Exclude value fields from the key: amounts, counts, text content, computed metrics, labels

### 3. Build Composite Keys
Most tables need multiple dimensions to identify observations:
- Financial: entity + time (+ metric_name if multiple metrics per row)
- Events: entity + time (+ event_type if multiple events per entity-time)
- Relationships: source + target
- Time series: entity + time + measure_type

### 4. Validate with Consolidation Logic
Ask yourself:
- If two rows have the same key, should they be merged into one row? If yes, good key.
- If two rows have different values but the same key, is one a duplicate/error? If yes, good key.
- Are you including measured values in the key? If yes, reconsider - those should be consolidated, not part of the key.
- Does the key define the "grain" of the consolidated table? If yes, good key.

### 5. Use Statistics as Supporting Evidence
The table statistics (null counts, distinct values) can help validate your choice:
- Key fields should have low null rates
- Key combinations should have reasonable cardinality
- But don't let statistics override semantic meaning

### 6. Expect Duplicates - That's Why We Need the Key
The current extracted data will have duplicates with the correct key - this is by design:
- Multiple extractions of the same observation create duplicate rows
- These duplicates will be consolidated using your primary key
- Text variations (spacing, punctuation) will exist in both keys and values
- Your job: identify the conceptual key that groups rows for consolidation, not ensure perfect uniqueness now

## SQL Query Capabilities
You have access to SQL queries to inspect the data. Use queries to:
- Check uniqueness: `SELECT field, COUNT(*) as cnt FROM table GROUP BY field HAVING cnt > 1`
- Check null rates: `SELECT COUNT(*) - COUNT(field) as null_count FROM table`
- Check cardinality: `SELECT COUNT(DISTINCT field) FROM table`
- Check composite uniqueness: `SELECT field1, field2, COUNT(*) FROM table GROUP BY field1, field2 HAVING COUNT(*) > 1`
- Sample values: `SELECT field FROM table LIMIT 10`

Focus queries on business fields first. Only query system fields if you're considering them as a fallback.

## Query Budget
You have a budget of {{max_queries}} SQL queries. Use them strategically:
- First query: Get overview statistics for all fields
- Subsequent queries: Drill down on promising primary key candidates
- Final queries: Validate your proposed primary key choice

## Input
**Question:** {{question}}

**Table Name:** {{table_name}}

**Schema:**
{{schema}}

**Table Statistics:**
{{table_stats}}

{% if query_history %}
**Previous Query Results:**
{{query_history}}
{% else %}
**Previous Query Results:**
No queries executed yet
{% endif %}

{% if sql_error_feedback %}
**SQL Error Feedback:**
{{sql_error_feedback}}
{% endif %}

## Output Format
Based on your analysis, provide your decision:

**Action**: Choose either "query" (run another SQL query to gather more information) or "finalize" (you have enough information to recommend a primary key)

**Reasoning**: Explain your thinking (2-3 sentences)
- If action is "query": Why do you need this information? What will it tell you about the consolidation key?
- If action is "finalize": Why is this the best primary key for consolidation? Explain what defines a unique observation and what fields will be consolidated during conflict resolution.

**SQL** (only if action is "query"): The SQL query to run

**Primary Key** (only if action is "finalize"): List of field names that form the primary key, e.g., ["company_name", "period_end_date"] or ["citing_paper_title", "cited_paper_title"]. Use null if no semantic primary key exists.

## Examples

### Example 1: Single Metric Per Row
```
Action: finalize
Reasoning: This table records a single metric for entities over time. The consolidation key is entity + time_period. Multiple extractions of the same entity-period will have the same key and be merged into one row. The metric_value field will be consolidated during conflict resolution (selecting the most reliable extracted value). Primary key excludes the value field, as values are what get consolidated, not what defines uniqueness.
Primary Key: ["entity_name", "time_period"]
```

### Example 2: Evidence / Dataset Table
```
Action: finalize  
Reasoning: This table represents a dataset of extracted mentions, not canonical facts. Each row is a distinct evidence instance that should be preserved even if it refers to the same underlying real-world observation. Consolidating rows would destroy provenance and reduce the dataset to one representative mention, which is not the goal.  
Primary Key: ["row_id"]
```


### Example 3: Multiple Metrics Per Row
```
Action: finalize
Reasoning: This table has multiple metric types per entity-period. The metric_type field IS an identifier (which measurement) that defines uniqueness along with entity and time. During consolidation, rows with the same entity-period-metric_type will be merged, with the amount field being consolidated via conflict resolution. Without metric_type in the key, we'd incorrectly merge different metrics (revenue vs. expenses) into one row.
Primary Key: ["entity_name", "time_period", "metric_type"]
```

### Example 4: Relationship/Association Data
```
Action: finalize
Reasoning: This table records relationships between entities. The consolidation key is source + target, as each unique relationship should appear once in the final table. Multiple extractions of the same relationship will be merged. Any additional fields (relationship_strength, relationship_date, etc.) are attributes of the relationship that get consolidated during conflict resolution, not part of the identifier.
Primary Key: ["source_entity", "target_entity"]
```

### Example 5: Event/Transaction Data
```
Action: finalize  
Reasoning: Events are identified by entity + timestamp. Multiple extractions of the same event will have the same key and be consolidated into one row. Fields like event_description, amount, and status are event attributes that will be resolved during conflict resolution. If truly multiple distinct events can occur at the exact same timestamp for the same entity, and there's no event_id or event_type to distinguish them, then entity + timestamp is still the semantic consolidation key.
Primary Key: ["entity_id", "timestamp"]
```

Now, provide your decision.

\end{lstlisting}

\paragraph{Document-Level Canonicalization}
Canonicalizes fields within each document.

\begin{lstlisting}[
  basicstyle=\ttfamily\tiny,
  breaklines=true,
  breakatwhitespace=false,
  frame=single,
  xleftmargin=0.3cm,
  xrightmargin=0.3cm
]
# instruction
You are canonicalizing the field **{{field_name}}** in rows from a single document.

## Context
- **Document**: {{document_name}}
- **All rows are from the SAME document** - variations are likely extraction artifacts, not real distinctions
- Use OTHER COLUMNS to determine if different values refer to the same entity

## Key Principle: Check All Columns for Entity Relationships

**Before deciding if two values are different entities, inspect ALL columns.** Other columns may contain information that reveals two seemingly different values actually refer to the same entity. Look for any column where:
- One value appears as an attribute of another
- The source text (quotes) mentions both forms together
- Context suggests equivalence

**Always start by examining the full row data:**
```sql
SELECT * FROM {{table_name}} ORDER BY {{field_name}};
```

If you see that value A appears in another column for rows with value B (or vice versa), they likely refer to the same entity and should be canonicalized.

## Workflow

1. **First inspection**: Check ALL columns to find entity relationships
2. **Look for cross-references**: Does value X appear in any column of rows where {{field_name}} = Y?
3. **If values appear related**, canonicalize them to one form
4. **Apply normalization** (case, punctuation, whitespace) for remaining variations
5. **Inspect Again to verify**: Check if the table still contains rows to be canonicalized.
5. **Stop** when no more relationships or variations exist

## Skip If
- Free-text / natural language content
- High cardinality with unique values
- Long verbose strings (>50 chars average)

## Canonicalization SQL Format (CRITICAL)

**You must REPLACE the column, not add a new one.** Do NOT use `SELECT *` for canonicalization.

**CORRECT** - explicitly list columns and keep the same column name:
```sql
SELECT
  row_id,
  CASE WHEN {{field_name}} = 'ValueA' THEN 'ValueB' ELSE {{field_name}} END AS {{field_name}},
  other_col1,
  other_col2,
  ...
FROM {{table_name}}
ORDER BY row_id;
```

**WRONG** - DO NOT do this:
```sql
SELECT *, CASE ... END AS {{field_name}}_canonicalized  -- WRONG: adds new column
SELECT *, CASE ... END AS {{field_name}}  -- WRONG: duplicates column
```

Requirements:
- List ALL columns explicitly (no `SELECT *`)
- The CASE expression must be named exactly `{{field_name}}` (same as original)
- End with `ORDER BY row_id`
- DuckDB syntax (no `|>` operators)
- Escape quotes: `''` not `\'`

## Input

**Table**: {{table_name}} | **Field**: {{field_name}} | **Document**: {{document_name}}

**Schema:**
{{schema}}

**Table Statistics:**
{{table_stats}}

{% if inspection_history %}
**Inspection History:**
{{inspection_history}}
{% endif %}

**Canonicalization History:** {{canonicalization_history}}

{% if canonicalization_error_feedback %}
**Error:** {{canonicalization_error_feedback}}
{% endif %}

**Inspections remaining:** {{inspections_remaining}}

## Actions

**inspect**: Run SQL to check column relationships
```
action: "inspect"
sql: "<SQL>"
reasoning: "<what you're checking>"
```

**canonicalize**: Merge related values (MUST replace column, not add new one)
```
action: "canonicalize"
sql: "<SELECT with explicit columns, CASE for {{field_name}}, ORDER BY row_id>"
reasoning: "<which values are the same entity>"
```

**stop**: After canonicalization has been performed and you have done inspection again: No more variations found
```
action: "stop"
reasoning: "<confirm all values are now consistent>"
```

**skip**: Field is not categorical
```
action: "skip"
reasoning: "<why>"
```

\end{lstlisting}

\paragraph{Error Detection Agent}
Agent that inspects data and decides which reconciliation operation to perform (deduplicate, aggregate, resolve conflicts, canonicalize).

\begin{lstlisting}[
  basicstyle=\ttfamily\tiny,
  breaklines=true,
  breakatwhitespace=false,
  frame=single,
  xleftmargin=0.3cm,
  xrightmargin=0.3cm
]
# instruction
You are a data quality controller for a single primary key group during table reconciliation.

Your role is to analyze a subset of rows sharing the same primary key value and decide which reconciliation operation to perform next.

You are processing rows for **one specific primary key value only**. This is a focused reconciliation task for a single entity.

The table contains values extracted from different pages of documents. Each page was processed independently by extraction workers.

## Pipeline Context

**Preprocessing completed**:
1. [OK] **Primary Key Selected**: A semantic primary key has been identified based on the schema and data domain
2. [OK] **Primary Key Fields Canonicalized**: Values in the primary key fields have been standardized (case normalization, whitespace trimming, format consistency)
3. [OK] **Table Split by Primary Key**: Rows have been grouped by primary key values

**Primary Key for this table**: {{primary_key}}

**Your specific primary key value**: {{pk_value}}

You are seeing ONLY the rows with this specific primary key value (typically 2-5 rows, occasionally more).

## Your Position in the Loop

You are in iteration {{iteration}} of {{max_iterations}} for this primary key group.

**Operations performed so far**: {{operations_history}}

After each operation you chose, an executor agent:
- Inspected the rows
- Generated and executed SQL to perform the operation
- Returned updated rows (could be 0, 1, or all rows depending on the operation)

You now see the **current state** of rows for this primary key after those operations.

## Your Goal

**Primary objective**: Reduce this primary key group to **exactly 1 row** through reconciliation operations.

Multiple rows with the same primary key indicate extraction redundancy, complementary information, or conflicts that need resolution. The ideal end state is a single consolidated row containing all valid information for this entity.

**When to stop before reaching 1 row**:
- Only if the executor has attempted an operation and determined, after deep analysis of the data, that the remaining rows represent genuinely distinct observations that should NOT be merged
- The executor's refusal must be based on semantic evidence (e.g., different temporal contexts, distinct aspects that shouldn't be combined)
- Simply noticing that rows "look different" is NOT sufficient reason to stop - differences are exactly what reconciliation operations are designed to handle

**When executor does not reduce to 1 row, diagnose the reason**:
1. **Legitimate refusal**: The executor reasoned that rows represent distinct entities/events that should not be merged (e.g., multiple separate occurrences, different aspects that shouldn't be combined) $\rightarrow$ route to **stop**
2. **Wrong operation chosen**: The executor tried to apply the operation but the current state actually requires a different operation (e.g., routed to consolidate but there are conflicts to resolve, or routed to deduplicate but there's complementary information to aggregate) $\rightarrow$ route to the **appropriate operation** that addresses the actual data pattern

## Your Decision Process

You iterate: **inspect** $\rightarrow$ **inspect** $\rightarrow$ ... $\rightarrow$ **route**

At each call, you decide:
1. **inspect**: Run a SQL query to understand the current state better
2. **route**: Route to a specific reconciliation operation (or stop)

## Available Route Actions

**Important**: When deciding which operation to route to, focus on **main data columns** (measured values like names, dates, amounts, descriptions). Use metadata columns (`*_quote`, `*_rationale`, `_confidence`, `_is_explicit`) to understand the context and semantics of the main data. However, do NOT route to an operation if the issue exists ONLY in metadata columns while the main data columns are already clean. Route based on problems in main data columns; the executor will then handle corresponding metadata columns appropriately.

1. **deduplicate**
   - Trigger when: Multiple rows with redundant information in non-primary-key fields
   - Purpose: Remove redundant rows, keeping the best representative
   - Executor will select the best row based on confidence, completeness, and data quality
   - Note: If a key column has non-redundant information (different values, NULLs to fill, or conflicts), route to aggregate, consolidate, or resolve_conflicts instead
   
2. **aggregate**
   - Trigger when: The field semantics CLEARLY indicate multiple values should coexist (e.g., list of all products, sum across categories)
   - Use sparingly: Only when you're confident the field is meant to hold multiple values, not when you're trying to avoid making a choice
   - Purpose: Combine multiple valid values into a unified representation (arrays, lists, sums, ranges)
   - Executor will apply: SUM (additive quantities across categories), ARRAY_AGG (multiple items in a collection), MIN/MAX (range bounds), COALESCE (filling different fields)
   - Examples: Multiple phone numbers $\rightarrow$ array, revenue across product lines $\rightarrow$ sum, multiple categories $\rightarrow$ array
   
3. **consolidate**
   - Trigger when: Information is scattered across rows with incomplete data that can be filled by copying values from other rows
   - Purpose: Broadcast complete information to fill gaps where data is missing
   - Executor will propagate consistent values across rows without combining or transforming them
   - Note: Returns multiple rows; use when information should be copied, not merged
   
4. **resolve_conflicts**
   - Trigger when: Different values exist for a field, and only ONE is correct or most appropriate
   - Default choice: When uncertain whether to aggregate or resolve, choose resolve_conflicts
   - Purpose: Select the single best value based on evidence, source quality, semantic fit, and specificity
   - Executor will evaluate which value to keep and discard the others
   - Examples: Different dates for same event, different names, different statuses, different measurements, conflicting attributes
   
5. **canonicalize**
   - Trigger when: Non-primary-key fields have different formats or representations
   - Example: "5M" vs "5000000", "USA" vs "United States"
   - Purpose: Standardize non-primary-key column values to a canonical form
   - Note: Primary key fields are already canonicalized. NULL values should generally be preserved.
   
6. **stop**
   - Trigger when: 
     * **Ideal case**: Successfully reduced to 1 row
     * **Acceptable case**: After diagnosing the most recent executor result, you determine the executor legitimately refused to merge because rows represent genuinely distinct observations (not because the wrong operation was chosen)
   - Before stopping with >1 row remaining: Analyze the executor's reasoning from the last operation. Did it refuse because rows are truly distinct, or because the operation chosen wasn't suitable for the actual data pattern?
   - If the operation was wrong (e.g., tried to deduplicate when should aggregate, or tried to consolidate when should resolve conflicts), route to the correct operation instead
   - Do NOT stop just because: Rows look different, have some NULL values, or seem "mostly done" - these are exactly what reconciliation addresses
   - Purpose: End the reconciliation loop for this primary key group

## Current State

Question to answer: {{question}}

Table name: {{table_name}}

Primary key value: {{pk_value}}

Current row count: {{current_row_count}}

## Schema
{{schema}}

## Table Statistics (for your specific PK group)
{{table_stats}}

## Your Inspection History (current controller call)
{{inspection_history}}

## Inspection Budget
{% if must_route_now %}
[WARNING] WARNING: No more inspections allowed. You MUST route to an action now (action: "route").
{% else %}
You can inspect {{inspections_remaining}} more time(s) before you must route.
{% endif %}

# Task

Decide: should you inspect more, or are you ready to route to an action?

**If you need more information**, output:
- action: "inspect"
- sql: Your inspection SQL query (e.g., check row differences, compare confidence scores)
- reasoning: Why you need this inspection

**Example inspections:**
```sql
-- View all rows for this PK
SELECT * FROM {{table_name}};

-- Compare measured values across rows
SELECT user_id, message_text, date, _confidence FROM {{table_name}};

-- Check metadata
SELECT _quote, _rationale, _is_explicit FROM {{table_name}};
```

**If you're ready to route**, output:
- action: "route"
- route_to: One of [deduplicate, aggregate, consolidate, resolve_conflicts, canonicalize, stop]
- reasoning: Why you're routing to this action based on the current state

Think about:
- **Target**: Are we at 1 row yet? If not, what operation will move us closer?
- Are rows redundant (same information), complementary (different columns populated), or conflicting (same column with different values)?
- **For conflicting values**: Default to **resolve_conflicts** to pick the best value. Only use **aggregate** if the field clearly needs to hold multiple values
- Are there format inconsistencies to canonicalize?
- Can consolidate or deduplicate reduce row count? Can resolve_conflicts eliminate value conflicts?
- **If last operation didn't reduce to 1 row**: Was it a legitimate refusal (rows are truly distinct) or was the wrong operation applied? If wrong operation, route to the correct one.
- **Only stop early if**: Executor has legitimately refused to merge after analysis because rows represent distinct entities, NOT because the chosen operation wasn't suitable for the data pattern

**Important**: 
- You're only seeing rows for ONE primary key value
- After you route, an executor will perform that ONE operation
- You'll be called again to decide the next operation (up to {{max_iterations}} total iterations)
- Each operation should make meaningful progress toward a clean, consolidated result

\end{lstlisting}

\paragraph{Reconciliation agent}
Agent that generates and executes SQL to perform the reconciliation operation chosen by the controller.

\begin{lstlisting}[
  basicstyle=\ttfamily\tiny,
  breaklines=true,
  breakatwhitespace=false,
  frame=single,
  xleftmargin=0.3cm,
  xrightmargin=0.3cm
]
# instruction
You are an executor agent for reconciling rows that share the same primary key values.

You are processing a small subset of the full table - specifically, **all rows that share a specific primary key value**.
{% if objective == "consolidate" %}Your task is to fill NULL values by broadcasting non-NULL values across these rows, returning **multiple rows** with gaps filled.{% elif objective == "aggregate" %}Your task is to aggregate values across these rows where appropriate, returning **multiple rows** with aggregated values broadcasted.{% elif objective == "deduplicate" %}Your task is to identify and remove redundant rows, returning **1 or more rows** depending on redundancy patterns.{% else %}Your task is to consolidate these rows into **1 output row** that best represents this entity.{% endif %}

The table contains values extracted from a document based on the provided schema. Each row was extracted independently by workers processing different document pages, which may have had partial context.

## Pipeline Context

**Preprocessing completed**:
1. [OK] **Primary Key Selected**: A semantic primary key has been identified
2. [OK] **Primary Key Fields Canonicalized**: Primary key values have been standardized

**Primary Key for this table**: {{primary_key}}
**Your specific primary key value**: {{pk_value}}

You are only seeing rows with this specific primary key value. Typically 2-5 rows, but occasionally more.

## Your Task

You have been assigned the objective: **{{objective}}**

You iterate: inspect $\rightarrow$ inspect $\rightarrow$ ... $\rightarrow$ generate_merge_sql

{% if objective == "consolidate" %}Your merge SQL should return **multiple rows** (all input rows with NULLs filled where appropriate).{% elif objective == "aggregate" %}Your merge SQL should return **multiple rows** (all input rows with aggregated values broadcasted where appropriate).{% elif objective == "deduplicate" %}Your merge SQL should return **1 or more rows** (deduplicated set where redundant rows are reduced to representatives, distinct rows are preserved).{% else %}Your merge SQL should return **1 row** that consolidates all the rows in your subset.{% endif %} However:
- If you determine that the rows in your group are NOT actually duplicates/conflicts but represent distinct entities or events, return **all original rows** by using `SELECT * FROM {{table_name}}` to preserve them.

**Note on metadata fields**: Columns like _rationale, _quote are metadata about the extraction process, not measured values.

{% if objective == "deduplicate" %}
## Objective: Deduplicate

**Goal**: Identify which rows contain redundant information in main data columns and deduplicate them, while preserving rows that contain distinct information. Return the deduplicated set.

### Phase 1: Inspect and Identify Redundancy Groups

**Inspection strategy**:
- Always examine all rows first: `SELECT * FROM {{table_name}}`
- Read the _quote and _rationale fields to understand what each row represents
- Compare ALL non-primary-key measured fields (date, amount, description, etc.)
- Consider the semantic context: Is it possible for this entity type to have multiple instances?

**Critical Decision: Can this event/entity occur multiple times in reality?**

**Step 1: Identify redundancy patterns**

Deduplication applies when rows have the same values across ALL (or nearly all) main data columns. You need to identify which specific rows share redundant information:

1. **All rows redundant**: All rows have the same values across main data columns
   - Example: 5 rows all describing the same entity with the same attributes (Company X, founded 2010, headquarters Chicago, CEO John Smith) extracted from 5 different pages
   - All 5 rows form one redundancy group because they contain the same complete set of facts
   - **Action needed**: Select 1 best row from this group

2. **Partial redundancy**: Some rows share the same complete set of main data values (forming redundant subsets), while other rows differ in at least one main data column
   - Example: 5 rows describing purchases where 3 have identical values (Item A, $50, 2024-01-10, Store X) and 2 have different values (Item B, $30, 2024-01-11, Store Y) and (Item A, $50, 2024-01-12, Store X)
   - The 3 identical rows form one redundancy group
   - The other 2 rows are distinct (each differs in date or item from the redundancy group)
   - **Action needed**: Select 1 best row from the redundancy group, keep the 2 distinct rows unchanged

3. **No redundancy**: Each row has different values in at least one main data column
   - Example: 3 rows describing different product variants (Color: Red, Size: M), (Color: Blue, Size: L), (Color: Red, Size: L)
   - No redundancy groups exist because each row has a unique combination of values
   - **Action needed**: Return all rows unchanged using `SELECT * FROM {{table_name}}`

**Why grouping matters**: Redundant rows (those with the same complete set of main data values) should be reduced to one representative per group, but distinct rows (those with at least one differing main data value) must be preserved. Your SQL needs to handle both simultaneously.

**Step 2: For each redundancy group identified**, determine the semantic nature:

**Important note on distinguishing metadata**: The absence of differing metadata (timestamps, quotes, rationale) between duplicate rows does NOT automatically indicate extraction redundancy. It may simply mean the extraction process was coarse-grained and didn't capture fine-grained distinguishing details. Focus on the semantic nature of what the data represents.

**Case A: Multiple real-world occurrences are possible**
- The entity/relationship represents actions, events, or transactions that can legitimately happen multiple times
- Characteristics of multiple occurrences:
  - Behavioral or transactional data (purchases, interactions, activities)
  - Events occurring across time that can repeat
  - Primary key includes temporal dimensions without fine-grained precision (date without timestamp)
  - Data describes "what happened" rather than "what is"
- These rows represent DISTINCT REAL-WORLD EVENTS that share the same entity identifier
- **Action**: Keep 1 best row, set `number_instances = <total_row_count>`
- **Interpretation**: This event/transaction occurred N times in the real world

**Case B: Only one instance exists in reality**
- The fact/attribute is an intrinsic property or one-time event that cannot change or repeat
- Characteristics of singular facts:
  - Definitional attributes (birth date, founding year, inherent properties)
  - One-time historical events that cannot recur
  - Data describes "what is" rather than "what happened"
  - Primary key is purely entity-based without temporal dimensions
- These rows are EXTRACTION REDUNDANCY: The same singular fact captured multiple times from different sources
- **Action**: Keep 1 best row, set `number_instances = 1`
- **Interpretation**: This fact exists once but was extracted N times

### Phase 2: Generate Merge SQL

After identifying redundancy groups and their semantic nature, generate SQL to:
1. From each redundancy group, select the best representative row (most complete, best quote)
2. Set `number_instances` appropriately for each representative
3. Include all rows that are not part of any redundancy group (distinct rows)
4. Return the deduplicated set

**Example 1: All rows redundant** (5 rows all contain same artist birth date extracted from different pages)

Inspection reveals: All 5 rows have identical main data (artist_name, date_of_birth, etc.), only differ in metadata
Decision: Extraction redundancy (birth date is a singular fact)
Action: Return 1 row with number_instances = 1

```sql
-- After inspection, chose row_id = 102 as best (highest confidence, clearest quote)
SELECT 
    * EXCLUDE(number_instances),
    1 AS number_instances  -- Singular fact extracted 5 times
FROM {{table_name}}
WHERE row_id = 102
```

**Example 2: Subset redundancy** (5 rows: 3 contain redundant alternate name "Name A", 2 contain distinct alternate names "Name B" and "Name C")

Inspection reveals:
- Rows 201, 202, 203: all have alternate_name = "Name A" $\rightarrow$ redundancy group 1
- Row 204: has alternate_name = "Name B" $\rightarrow$ distinct
- Row 205: has alternate_name = "Name C" $\rightarrow$ distinct

Decision: 
- "Name A" appears 3 times (extraction redundancy, same name extracted from 3 pages)
- "Name B" and "Name C" are distinct values, should be preserved

Action: Keep 1 row from the "Name A" group, plus the 2 distinct rows $\rightarrow$ return 3 rows total

```sql
-- After inspection, chose row_id = 201 from the "Name A" group as best representative
SELECT 
    * EXCLUDE(number_instances),
    1 AS number_instances  -- "Name A" is singular, extracted 3 times
FROM {{table_name}}
WHERE row_id = 201  -- Best of the "Name A" redundancy group

UNION ALL

-- Preserve distinct rows unchanged
SELECT * FROM {{table_name}}
WHERE row_id IN (204, 205)  -- "Name B" and "Name C" are distinct
```

**Alternative pattern for partial deduplication**:
```sql
-- Use CASE to mark which rows to keep, then filter
WITH ranked AS (
    SELECT *,
        CASE 
            -- Redundancy group: rows with similar main data
            WHEN main_col_a = 'duplicate_value' AND main_col_b IS NULL 
            THEN ROW_NUMBER() OVER (PARTITION BY main_col_a, main_col_b ORDER BY CASE _confidence WHEN 'Very High' THEN 5 WHEN 'High' THEN 4 WHEN 'Medium' THEN 3 WHEN 'Low' THEN 2 WHEN 'Very Low' THEN 1 ELSE 0 END DESC, row_id)
            -- All other rows are distinct, keep them
            ELSE 1
        END AS keep_rank
    FROM {{table_name}}
)
SELECT 
    * EXCLUDE(keep_rank, number_instances),
    CASE 
        WHEN <condition for redundancy group> THEN <1 or count>
        ELSE number_instances  -- Preserve for distinct rows
    END AS number_instances
FROM ranked
WHERE keep_rank = 1
```

**Key points**:
- `number_instances` value depends on your semantic analysis
- Return 1 or more rows depending on redundancy structure
- Each returned row should have appropriate `number_instances` value
- Preserve rows with distinct main data values
- Only deduplicate rows that are truly redundant in main data columns
- Use inspection results to justify your choice
- Document your reasoning clearly
{% endif %}

{% if objective == "aggregate" %}
## Objective: Aggregate

**Goal**: For columns containing multiple distinct values that should be combined, compute an aggregated representation (sum, collection, concatenation) and broadcast it to all rows.

**Inspection strategy**:
- Examine all rows: `SELECT * FROM {{table_name}}`
- For each column, identify if it has multiple distinct non-NULL values
- Read _quote and _rationale to verify these values are complementary and should be combined
- Determine the appropriate aggregation method for each column

**Decision logic for each column**:
For columns with multiple distinct non-NULL values:
- **Complementary values that should combine** $\rightarrow$ apply aggregation across rows and broadcast result to all rows
  - Multiple valid instances of a multi-valued attribute: ARRAY_AGG, STRING_AGG
  - Numeric quantities to sum: SUM
  - Range boundaries: MIN/MAX pairs
- **Conflicting values** (mutually exclusive, only one correct) $\rightarrow$ leave unchanged, will be handled by resolve_conflicts
- **Single value or all NULLs** $\rightarrow$ leave unchanged

**Key principle**: Aggregation combines semantically compatible values that should coexist. Return all rows with aggregated values broadcasted where appropriate.

**Merge SQL strategy**:
1. **First inspect**: Determine which columns have multiple values that should be combined
2. **Then generate SQL**: Use window functions to compute aggregated values and broadcast to all rows
3. Columns that don't need aggregation remain unchanged
4. Return ALL rows with aggregations applied where needed

**SQL Pattern using window functions**:
```sql
-- Apply aggregation to specific columns that need combining, broadcast result to all rows
-- Use * EXCLUDE to avoid listing all columns
SELECT 
  * EXCLUDE(col_a, col_a_quote, col_b),
  -- Aggregate multi-valued column across all rows in this PK group
  STRING_AGG(DISTINCT col_a, ' | ') OVER (PARTITION BY pk_col1, pk_col2) AS col_a,
  STRING_AGG(DISTINCT col_a_quote, ' | ') OVER (PARTITION BY pk_col1, pk_col2) AS col_a_quote,
  -- Sum numeric values across rows
  SUM(col_b) OVER (PARTITION BY pk_col1, pk_col2) AS col_b
FROM {{table_name}}
WHERE pk_col1 = '...' AND pk_col2 = '...'
```

**Alternative pattern using subqueries**:
```sql
-- Compute aggregations in subqueries, join back to preserve all rows
-- Use * EXCLUDE for cleaner SQL
SELECT 
  t.* EXCLUDE(col_a, col_a_quote),
  -- Replace col_a with aggregated value across all rows
  (SELECT STRING_AGG(DISTINCT col_a, ' | ') FROM {{table_name}} WHERE pk_col1 = t.pk_col1 AND pk_col2 = t.pk_col2) AS col_a,
  (SELECT STRING_AGG(DISTINCT col_a_quote, ' | ') FROM {{table_name}} WHERE pk_col1 = t.pk_col1 AND pk_col2 = t.pk_col2) AS col_a_quote
FROM {{table_name}} t
WHERE t.pk_col1 = '...' AND t.pk_col2 = '...'
```

**Key points**:
- Return ALL input rows with the same schema (no columns added or dropped)
- Only aggregate columns where combining values makes semantic sense
- Broadcast aggregated values to all rows in the group
- Columns with conflicts or single values remain untouched for later iterations
{% endif %}

{% if objective == "consolidate" %}
## Objective: Consolidate

**Goal**: Fill NULL values in non-primary-key columns by broadcasting non-NULL values from other rows in the same column, where semantically appropriate.

**Inspection strategy**:
- Examine all rows: `SELECT * FROM {{table_name}}`
- For each column, identify if it has NULLs in some rows and non-NULL values in others
- Read _quote and _rationale to determine if broadcasting makes semantic sense
- Check if non-NULL values in the same column are consistent (not conflicting)

**Decision logic for each column**:
For each column with mixed NULL/non-NULL values:
- **If there is exactly ONE distinct non-NULL value** $\rightarrow$ broadcast it to fill NULLs in that column
- **If there are MULTIPLE distinct non-NULL values** $\rightarrow$ do NOT consolidate this column; leave it unchanged for other operations (aggregate or resolve_conflicts) to handle
- If semantic context suggests NULLs should remain (e.g., truly missing data) $\rightarrow$ preserve NULLs

**Key principle**: Consolidation is a focused operation that only fills gaps where information is unambiguous. When a column has multiple distinct non-NULL values, those values need different handling (combining via aggregation or selecting via conflict resolution) which will be addressed in subsequent iterations.

**Merge SQL strategy**:
1. **First inspect**: Examine all rows to understand which row has the best value for each column
2. **Then generate SQL**: For each consolidatable column, choose the specific row_id with the most reliable value
3. Use subqueries or window functions to broadcast from those chosen rows
4. Return ALL input rows with NULLs filled ONLY in columns that meet consolidation criteria
5. For columns with multiple distinct non-NULL values, preserve them exactly as-is

**Pattern 1: Selective broadcasting with subqueries**
  ```sql
  -- After inspecting, substitute actual row_ids you identified as having the best values
  -- Use * EXCLUDE for cleaner SQL that avoids listing all columns
  SELECT 
    t.* EXCLUDE(col_a, col_a_quote, col_b, col_b_quote),
    -- Broadcast col_a from the row you determined has the best value (e.g., row 117)
    COALESCE(t.col_a, (SELECT col_a FROM {{table_name}} WHERE row_id = <row_with_best_col_a>)) AS col_a,
    COALESCE(t.col_a_quote, (SELECT col_a_quote FROM {{table_name}} WHERE row_id = <row_with_best_col_a>)) AS col_a_quote,
    -- Broadcast col_b from a different row you chose (e.g., row 118)
    COALESCE(t.col_b, (SELECT col_b FROM {{table_name}} WHERE row_id = <row_with_best_col_b>)) AS col_b,
    COALESCE(t.col_b_quote, (SELECT col_b_quote FROM {{table_name}} WHERE row_id = <row_with_best_col_b>)) AS col_b_quote
  FROM {{table_name}} t
  ```

**Pattern 2: Using window functions with specific row selection**
  ```sql
  -- Use * EXCLUDE for simpler SQL
  SELECT 
    * EXCLUDE(col_a, col_b),
    -- Consolidate col_a by broadcasting from the row with best metadata
    COALESCE(col_a, MAX(CASE WHEN row_id = <chosen_row_id> THEN col_a END) OVER (PARTITION BY pk_col1, pk_col2)) AS col_a,
    -- Consolidate col_b from a different source row
    COALESCE(col_b, MAX(CASE WHEN row_id = <other_row_id> THEN col_b END) OVER (PARTITION BY pk_col1, pk_col2)) AS col_b
  FROM {{table_name}}
  ```

**Key approach**: After inspecting the rows, determine for each consolidatable column which specific row has the most reliable non-NULL value (based on quote quality, rationale clarity, or completeness), then broadcast that specific value.

- Keep metadata columns (_quote, _rationale) from original rows unchanged

**Key points**:
- Return multiple rows (all input rows with selective NULLs filled)
- Only modify columns where broadcasting a single value is appropriate
- Leave columns with multiple distinct non-NULL values untouched for subsequent operations
- The reconciliation loop will handle remaining issues in later iterations
{% endif %}

{% if objective == "resolve_conflicts" %}
## Objective: Resolve Conflicts

**Goal**: When rows have conflicting measured values, determine if it's a true conflict to resolve or separate legitimate occurrences.

**Inspection strategy**:
- Examine all rows: `SELECT * FROM {{table_name}}`
- Identify which non-primary-key fields have conflicting (different distinct) values
- Read _quote and _rationale fields to understand what each row represents

**Important decision**:
- **If conflicting field is a temporal/event field** (date, timestamp) and both values are equally valid $\rightarrow$ these represent separate events, keep all rows: `SELECT * FROM {{table_name}}`
- **If conflict is due to measurement error** (different amounts, descriptions for same event) $\rightarrow$ resolve by selecting most reliable value

**Merge SQL strategy for true conflicts**:
- Use metadata to decide: Compare _quote and _rationale for supporting evidence
- **Decision process**: For each competing row, explicitly evaluate it against each criterion below. Document your assessment for each criterion before making the final selection.
- Selection criteria:
  1. **Field definition alignment**: Prefer values that match the semantic scope of the field
     - Choose values that directly answer what the field asks for, not broader or narrower variants
     - Match the specificity: if the field has modifiers or constraints, select values that align with those constraints
     - When the field name corresponds to a standard reporting item, the value extracted from where that item is formally reported should be strongly preferred over values described or explained elsewhere, even if explanatory sections provide additional context
  2. **Extraction location relevance**: Prefer values extracted from locations that directly serve the field's purpose
     - Consider which document sections or structures are authoritative for this type of information
     - Prioritize locations where this field would naturally appear in its primary context
     - When available, examine text_header or similar metadata to understand extraction context
  3. **Source authority**:
     - Direct measurements or final values over constituent parts or supplementary details
     - Structured presentations (tables, statements, forms, labeled fields) over descriptive text (discussions, notes, commentary)
     - Primary data sections over explanatory sections
     - Explicit field labels over derived mentions
  4. **Quote quality**: Prefer direct quotes over inferred values
  5. **Rationale clarity**: Prefer clear, specific rationale
- **Preferred approach**: After inspection, identify the specific row_id you want to keep and filter directly: `SELECT * FROM {{table_name}} WHERE row_id = <chosen_row_id>`
- Alternative: Use ROW_NUMBER() with ORDER BY, but be careful with tiebreakers
{% endif %}

# Input

{% if verification_mode %}
**Phase**: Verification

Primary key value you were processing: {{pk_value}}

## Initial Table Schema (before your SQL)
{{initial_schema}}

## Final Table Schema (after your SQL)
{{final_schema}}

### Your Generated SQL
```sql
{{generated_sql}}
```

### Your Reasoning When Generating
{{generation_reasoning}}

### Available Tables for Inspection
Two tables are registered in DuckDB:
- **`initial_table`**: {{initial_row_count}} rows BEFORE your SQL was applied
- **`final_table`**: {{final_row_count}} rows AFTER your SQL was applied

**IMPORTANT**: You do NOT have the actual table contents yet. You must inspect both tables using SQL to verify the transformation was executed correctly.

### Your Verification Inspection History
{{verification_inspection_history}}

{% else %}
Question to answer: {{question}}

Table name: {{table_name}}

Primary key value you're processing: {{pk_value}}

## Schema
{{schema}}

## Table Statistics (for your subset)
{{table_stats}}

## Controller's Reasoning
The controller routed to this objective with the following reasoning:
{{controller_reasoning}}

## Your Inspection History
{{inspection_history}}

## Inspection Errors (if any)
{{inspection_error_feedback}}

## SQL Generation Errors (if any - for retry)
{{sql_error_feedback}}

## Status

{% if inspections_remaining > 0 %}
**Phase**: Inspection (optional)
**Budget**: You can inspect {{inspections_remaining}} more time(s), or choose to generate merge SQL now.

After inspection phase ends, you'll enter the SQL generation phase with 3 attempts to generate valid SQL.
{% else %}
**Phase**: SQL Generation
[WARNING] You are now in the SQL generation phase. You can no longer inspect.
Generate the merge SQL. If it fails, you'll get error feedback and up to 2 more attempts (3 total).
{% endif %}
{% endif %}

# Task

{% if verification_mode %}
Your merge SQL has been executed. Verify that the SQL correctly transformed initial_table $\rightarrow$ final_table as intended by your reasoning.

**VERIFICATION SCOPE:**
You are checking if your SQL performed the transformation correctly, NOT whether the reconciliation approach itself was correct. Focus on: Did the SQL execute the logic described in your reasoning?

**VERIFICATION WORKFLOW:**

1. **First, you MUST inspect** (especially if you haven't inspected yet):
   - action: "inspect"
   - sql: Query to compare initial_table and final_table (e.g., SELECT * from both, or JOIN them)
   - reasoning: What transformation aspects you're verifying (row count changes, column transformations, etc.)

2. **After inspecting and verbalizing what you observed, then decide:**

   **Option A: Approve** - The SQL correctly executed the intended transformation
   - action: "approve"
   - reasoning: Confirm the transformation matches your generation reasoning:
     * Does initial_table $\rightarrow$ final_table match the expected transformation?
     * Are row counts, aggregations, and column values as described in your reasoning?

   **Option B: Regenerate** - The SQL did not execute the intended transformation correctly
   - action: "regenerate"  
   - sql: Corrected SQL that properly implements your original reasoning
   - reasoning: What transformation error did you observe? How does the new SQL fix it?

**Remaining Inspections**: {{remaining_inspections}}

**Critical Rules:**
- Inspect the actual data before deciding
- Compare initial_table vs final_table to verify the transformation
- Approve if the SQL correctly implemented your reasoning
- Regenerate if the SQL failed to execute the intended transformation
- Regenerated SQL is auto-accepted (no second verification)

{% else %}
Decide: should you inspect more, or are you ready to generate the merge SQL?

If you need more information, output:
- action: "inspect"
- sql: Your inspection SQL query (SELECT statement to understand your subset)
- reasoning: Why you need this inspection

If you're ready to generate the merge, output:
- action: "generate_merge_sql"
- sql: Your merge SQL query that {% if objective == "consolidate" %}returns **multiple rows** with NULLs filled{% elif objective == "aggregate" %}returns **multiple rows** with aggregated values broadcasted{% elif objective == "deduplicate" %}returns **1 or more rows** with redundancy removed{% else %}consolidates all rows in your subset to **exactly 1 row**{% endif %}
- reasoning: How your SQL accomplishes the objective

**Important**: 
- **Merge SQL format**: Must be a SELECT statement{% if objective not in ["consolidate", "aggregate", "deduplicate"] %} that returns exactly 1 row{% endif %} (no CREATE TABLE, INSERT, or DELETE)
- **Schema preservation**: Include ALL original columns (row_id, page_number, *_quote, *_rationale)
- **number_instances column**: {% if objective == "deduplicate" %}Set this to the total row count (if true duplicates) or 1 (if semantically distinct instances) for each returned row{% elif objective in ["consolidate", "aggregate"] %}Preserve the existing value for each row{% else %}This column is not visible in your table; it will be preserved automatically{% endif %}
- **Quote/rationale columns**: When merging, update these columns accordingly (concatenate sources or keep the best)
{% if objective in ["consolidate", "aggregate"] %}- **Multiple row output**: Your SQL should return all input rows with transformations applied where appropriate{% elif objective == "deduplicate" %}- **Flexible row output**: Return 1 or more rows depending on redundancy structure (1 representative per redundancy group + all distinct rows){% else %}- **Single row output**: Your SQL MUST return exactly 1 row. Use LIMIT 1, ROW_NUMBER() = 1, or GROUP BY as appropriate{% endif %}
- **Result limits**: Inspection SQL should use LIMIT to avoid large results
{% endif %}

\end{lstlisting}

\subsubsection{Answer Generation}

\paragraph{SQL Query Generator}
Iteratively generates SQL queries to inspect the reconciled data and gather information for answering.

\begin{lstlisting}[
  basicstyle=\ttfamily\tiny,
  breaklines=true,
  breakatwhitespace=false,
  frame=single,
  xleftmargin=0.3cm,
  xrightmargin=0.3cm
]
# instruction
You are a SQL query generator for answering questions about data in DuckDB tables.
{% if citation_mode %}

You have completed the query generation phase. Your queries have been used by the final answer generator to produce the verbalized answer.

Now you need to generate a single SQL query that selects the rows used to answer the question, including all provenance columns (quote, rationale, text_headers). This will enable proper citation of the source data.
{% else %}
You cannot see full tables - only schema and statistics. Use SQL queries to retrieve information needed to answer the question.
{% endif %}

{% if not citation_mode %}
## Strategy
{% if strategy == "full_table" %}
Table is small (<=100 rows). Start by sampling the full table with `SELECT * FROM table LIMIT {{row_limit}}` to see all data, then use focused queries if needed.
{% else %}
Table is large (>100 rows). Use focused SQL queries with filters and aggregations. Results are limited to {{row_limit}} rows. Consider sampling first if you need to understand the structure.
{% endif %}

## Guidelines
- You have {{queries_remaining}} quer{{ "y" if queries_remaining == 1 else "ies" }} remaining
{% if is_first_query and strategy != "sql_focused" %}
- **First query**: Sample the table (`SELECT * FROM table LIMIT {{row_limit}}`) to see actual data
{% endif %}
- Use focused queries: filters, aggregations, ordering
- Verify results through multiple approaches (sample, then aggregate, or cross-check with different filters)
- **[Important] Empty results**: If a query returns no rows, verify this with alternative queries before drawing conclusions---empty results often indicate SQL issues rather than missing data
- Check data types and sample values before filtering
- Each query result is truncated to {{row_limit}} rows
- For calculations, use SQL functions like ROUND() to ensure precision
- Default ordering: use ORDER BY row_id ASC to preserve order in the document unless the question explicitly requests a different sort (e.g., by date).
- Only the values might be wrong (schema/columns are assumed correct). Use rationale/evidence columns to vet rows; prefer rows with clear rationales that match the question and drop rows with weak or conflicting rationale. Feel free to write cleaning SQL (filters, trims, normalizations) before using the data.

# Working with dates
- If a column is DATE/TIMESTAMP, do not use LIKE, SUBSTR, LEFT, RIGHT, ILIKE, or regex on it.
- Use date-native predicates only: EXTRACT, DATE_TRUNC, strftime, or range filters with DATE 'YYYY-MM-DD'.


{% if sql_error_feedback != "No previous SQL errors" %}
## Previous SQL Errors
{{sql_error_feedback}}
Fix the SQL and retry.
{% endif %}
{% endif %}

# input

## Question
{{question}}

## Schema
{{schema}}

{% if not citation_mode %}
## Table Statistics
{{table_stats}}

## Previous Queries
{{query_history}}

## Status
{% if queries_remaining > 0 %}
{{queries_remaining}} quer{{ "y" if queries_remaining == 1 else "ies" }} remaining. Either:
- `action: "query"` with SQL to retrieve more information
- `action: "finalize"` if you have sufficient information
{% else %}
No queries remaining. Must finalize now (`action: "finalize"`).
{% endif %}
{% endif %}

{% if citation_mode %}
# Citation SQL Generation

## Previous Investigation History
{{query_history}}

## Finalization Reasoning
{{finalization_reasoning}}

## Final Answer Generated
{{final_answer}}

## Your Citation SQL Attempts
{{citation_attempts_history}}

## Task
Generate a SQL query that selects the rows used to answer the question. Include:
- All primary key columns
- All data columns referenced in the answer
- **Provenance columns**: Any columns with names containing "quote", "rationale", or "text_header" (case-insensitive)
- Filter to only the specific rows that contributed to the answer

**Workflow:**
1. **If you haven't tried yet OR need to refine**: 
   - action: "execute"
   - sql: Your citation SQL query
   - reasoning: Which rows you're selecting and why

2. **If your previous SQL result looks good**:
   - action: "finalize"
   - sql: "" (leave empty)
   - reasoning: Confirm that the previous result contains the right citation rows

**Attempts remaining**: {{attempts_remaining}}

Provide:
- `reasoning`: Explain your decision
- `action`: "execute" (try SQL) or "finalize" (accept previous result)
- `sql`: SQL query (required if action="execute", leave empty if action="finalize")
{% else %}
Provide:
- `reasoning`: What you need or why you're finalizing
- `action`: "query" or "finalize"
- `sql`: SQL query (required if action is "query")
{% endif %}

\end{lstlisting}

\paragraph{Answer Verbalization}
Verbalizes SQL query results into a natural language answer.

\begin{lstlisting}[
  basicstyle=\ttfamily\tiny,
  breaklines=true,
  breakatwhitespace=false,
  frame=single,
  xleftmargin=0.3cm,
  xrightmargin=0.3cm
]
# instruction
{% if citation_mode %}
Generate a citation paragraph based on the source data rows below.

You have already answered the question. Now you need to generate a citation paragraph that explains which source documents and specific data points were used. This paragraph will be appended to your answer to provide proper attribution.

**DO NOT re-answer the question.** Only generate the citation paragraph.
{% else %}
Answer the question using the SQL query results below.

You executed one or more SQL queries to gather information from the database. Based on the query results, provide a comprehensive answer to the question.
{% endif %}

# input
# Question
{{question}}

# Schema
{{classes}}

{% if citation_mode %}
# Your Generated Answer
{{generated_answer}}

# Citation SQL Query
The following SQL query was executed to retrieve source data rows with provenance information:
```sql
{{citation_sql}}
```

# Source Data Rows (with Provenance)
{{citation_data}}

# Task
Generate a concise citation paragraph (2-4 sentences) that:
- Mentions the specific source documents or data entries used
- References key provenance information (quotes, rationales, text headers) if available
- Explains how the source data supports the answer

This citation paragraph will be appended to your answer above. Format it to flow naturally after the answer.
{% else %}
# SQL Queries Executed
{{tool_call}}

# Query Results
{{tool_output}}

Based on these SQL query results, provide a clear and accurate answer to the question. Reference specific values from the query results to support your answer.
{% endif %}


\end{lstlisting}

\subsection{Evaluation Prompts}
\label{app:eval-prompts}

\paragraph{Hard Evaluator}

\begin{lstlisting}[
  basicstyle=\ttfamily\tiny,
  breaklines=true,
  frame=single,
  xleftmargin=0.3cm,
  xrightmargin=0.3cm
]
# instruction
Evaluate the predicted answer against the gold answer. The predicted answer should match the gold answer.

Use the following criteria to evaluate the predicted answer:
- If there are rounding errors, its incorrect.
- If final framing doesn't match, its incorrect.
- If the justification doesn't match, its incorrect.
- If the only issue is gold being in decimal and the predicted answer being in fraction or percentage, its correct.

# input
# Question
{{question}}

# Gold Answer
{{gold_answer}}

# Predicted Answer
{{predicted_answer}}
\end{lstlisting}

\paragraph{Soft Evaluator}

\begin{lstlisting}[
  basicstyle=\ttfamily\tiny,
  breaklines=true,
  frame=single,
  xleftmargin=0.3cm,
  xrightmargin=0.3cm
]
# instruction
Evaluate the predicted answer against the gold answer. The predicted answer should match the gold answer.

# input
# Question
{{question}}

# Gold Answer
{{gold_answer}}

# Predicted Answer
{{predicted_answer}}
\end{lstlisting}

\paragraph{Loong Evaluator}

\begin{lstlisting}[
  basicstyle=\ttfamily\tiny,
  breaklines=true,
  frame=single,
  xleftmargin=0.3cm,
  xrightmargin=0.3cm
]
We would like to request your feedback on the performance of the AI assistant in response to the user question displayed below according to the gold answer. Please use the following listed aspects and their descriptions as evaluation criteria:
    - Accuracy and Hallucinations: The assistant's answer is semantically consistent with the gold answer; The numerical value and order need to be accurate, and there should be no hallucinations.
    - Completeness: Referring to the reference answers, the assistant's answer should contain all the key points needed to answer the user's question; further elaboration on these key points can be omitted.
Please rate whether this answer is suitable for the question. Please note that the gold answer can be considered as a correct answer to the question.

The assistant receives an overall score on a scale of 1 to 100, where a higher score indicates better overall performance.
Please note that if the assistant's answer and the gold answer fully meet the above criteria, its overall rating should be the full marks (100).
Please first provide a comprehensive explanation of your evaluation, avoiding any potential bias.
Then, output a line indicating the score of the Assistant.

PLEASE OUTPUT THE SCORE ON A SCALE OF 1 TO 100.

# input
# Question
{{question}}

# Gold Answer
{{gold_answer}}

# Predicted Answer
{{predicted_answer}}

[Question]
{{question}}

[Gold Answer]
{{gold_answer}}

[The Start of Assistant's Predicted Answer]
{{predicted_answer}}
[The End of Assistant's Predicted Answer]

[System]
We would like to request your feedback on the performance of the AI assistant in response to the user question displayed above according to the gold answer. Please use the following listed aspects and their descriptions as evaluation criteria:
    - Accuracy and Hallucinations: The assistant's answer is semantically consistent with the gold answer; The numerical value and order need to be accurate, and there should be no hallucinations.
    - Completeness: Referring to the reference answers, the assistant's answer should contain all the key points needed to answer the user's question; further elaboration on these key points can be omitted.
Please rate whether this answer is suitable for the question. Please note that the gold answer can be considered as a correct answer to the question.

The assistant receives an overall score on a scale of 1 to 100, where a higher score indicates better overall performance.
Please note that if the assistant's answer and the gold answer fully meet the above criteria, its overall rating should be the full marks (100).
Please first provide a comprehensive explanation of your evaluation, avoiding any potential bias.
Then, output a line indicating the score of the Assistant.

PLEASE OUTPUT WITH:
- "EXPLANATION" IS THE REASON BEHIND THE CORRECT SCORE GIVEN TO THE PREDICTED ANSWER.
- "CORRECT" IS A SCORE ON A RANGE OF 1 TO 100.

\end{lstlisting}

\paragraph{Numeric Extractor}

\begin{lstlisting}[
  basicstyle=\ttfamily\tiny,
  breaklines=true,
  frame=single,
  xleftmargin=0.3cm,
  xrightmargin=0.3cm
]
# instruction
Given the question and the assistant's full answer, extract only the final numeric answer. Return digits with an optional leading minus sign; remove commas, units, and explanation. If no numeric answer is present, return null.

# input
# Question
{{question}}

# Predicted Answer
{{predicted_answer}}

\end{lstlisting}

\paragraph{Multi-Numeric Extractor}

\begin{lstlisting}[
  basicstyle=\ttfamily\tiny,
  breaklines=true,
  frame=single,
  xleftmargin=0.3cm,
  xrightmargin=0.3cm
]
# instruction
Given a question, the ground truth answer, and the generated answer, first determine if the answers contain numeric values that should be compared.

If the answers are purely textual/qualitative (e.g., lists of names, descriptions, yes/no answers without numbers), set has_numbers to False and return empty lists.

If the answers contain numeric values (counts, percentages, amounts, rankings, etc.), set has_numbers to True and extract all numeric values from both answers.

For each numeric value, identify:
1. A key/identifier (e.g., country name, industry name, or "value" for single numbers)
2. The numeric value

Return two lists: one for ground truth and one for generated answer.

For single numeric answers, use "value" as the key.
For multiple values like rankings or distributions, use the entity name (country, industry, etc.) as the key.

SPECIAL HANDLING FOR RANKINGS:
- If rankings have associated counts/values (e.g., "Artist A: 39 songs, Artist B: 25 songs"), extract those counts
- If rankings are purely ordinal without counts (e.g., "1. Artist A, 2. Artist B, 3. Artist C"), extract the ordinal position as the value:
  - First/1st item $\rightarrow$ value: 1
  - Second/2nd item $\rightarrow$ value: 2
  - Third/3rd item $\rightarrow$ value: 3, etc.
- This allows comparison of ranking order even when no explicit counts are given

IMPORTANT: Canonicalize the keys to ensure matching between ground truth and generated answer.
- Use the same canonical key name for entities that refer to the same thing in both answers
- Normalize trivial differences such as:
  - Capitalization (e.g., "Music" and "music" should use the same key)
  - Abbreviations (e.g., "USA" and "United States" should use the same key)
  - Punctuation and formatting (e.g., "Film/TV" and "Film & TV" should use the same key)
  - Singular/plural forms (e.g., "artist" and "artists" should use the same key)
- Choose the clearest, most standard form as the canonical key
- Apply the same canonicalization to both ground truth and generated values

# input
# Question
{{question}}

# Ground Truth Answer
{{gold_answer}}

# Generated Answer
{{predicted_answer}}

\end{lstlisting}

\section{Example Execution Trajectories}
\label{app:trajectories}

We provide representative execution traces from the \system{} pipeline to illustrate key
components in action. These logs show the reasoning and SQL operations performed by
different agents during processing.

\subsection{Reconciliation Loop}

Example of the reconciliation agent (controller) inspecting data, routing to an operation, and the operations (executor) performing conflict resolution.

\begin{lstlisting}[
  basicstyle=\ttfamily\tiny,
  breaklines=true,
  frame=single,
  xleftmargin=0.3cm,
  xrightmargin=0.3cm
]
INFO     ================================================================================
INFO     [PK ('3M', '2022-12-31')] Processing: 8 rows, max 5 iterations
INFO     ================================================================================
INFO     [PK ('3M', '2022-12-31')] Controller (iter=0): Inspection 1 SQL
INFO     [PK ('3M', '2022-12-31')] SQL: SELECT row_id, page_number, operating_income, net_sales,
         operating_margin, operating_income_quote, net_sales_quote, operating_margin_quote FROM
         OperatingMarginMetrics_69ffa681 ORDER BY row_id;
INFO     [PK ('3M', '2022-12-31')] Controller (iter=0): Inspection 2 SQL
INFO     [PK ('3M', '2022-12-31')] SQL: SELECT row_id, page_number, operating_income, net_sales,
         operating_margin, operating_income_rationale, net_sales_rationale, operating_margin_rationale FROM
         OperatingMarginMetrics_69ffa681 ORDER BY row_id;
INFO     [PK ('3M', '2022-12-31')] Controller (iter=0): Routing to 'resolve_conflicts'
INFO     [PK ('3M', '2022-12-31')] Reasoning: There are 8 rows for the same PK ('3M', '2022-12-31'), all in the
         OperatingMarginMetrics table. The main data columns (operating_income, net_sales, operating_margin) show
         several repeated values for total company (operating_income=6539, net_sales=34229,
         operating_margin~19.1), but also several rows for business segments (Safety and Industrial,
         Transportation and Electronics, Health Care, Consumer) with distinct values. Some rows have calculated
         margins, some have explicit margins, and some have missing values. The segment rows are not redundant
         with the total company rows---they represent different aspects. However, among the total company rows,
         there are clear redundancies (same values, slightly different margin due to rounding/calculation, some
         with explicit margin, some with calculated). To move toward a single consolidated row, we need to
         resolve conflicts where the same field has slightly different values (e.g., operating_margin 19.10 vs
         19.11), and select the best representative for total company. For the segment rows, aggregation may be
         needed later, but first, resolving conflicts among total company rows will reduce redundancy and clarify
         which rows are truly distinct. Therefore, the next step is to resolve conflicts.
INFO     [PK ('3M', '2022-12-31')] Executor (op=resolve_conflicts): Inspection 1
INFO     [PK ('3M', '2022-12-31')] Reasoning: Before generating the merge SQL, I need to inspect all 8 rows for
         this primary key to understand the nature of the conflicts in operating_income, net_sales, and
         operating_margin fields. Specifically, I need to see the _quote and _rationale columns for each, as well
         as the extraction context (page_number, text_header) to determine which row(s) are most authoritative
         and align best with the field definitions. This will allow me to select the most reliable values and/or
         combine relevant narrative details as required by the task. Since there are 8 rows and multiple distinct
         values for the key metrics, a full inspection is necessary to resolve conflicts and deduplicate
         appropriately.
INFO     [PK ('3M', '2022-12-31')] SQL: SELECT * FROM OperatingMarginMetrics_69ffa681 WHERE company_name = '3M'
         AND fiscal_year_end = '2022-12-31' ORDER BY row_id ASC LIMIT 8
INFO     [PK ('3M', '2022-12-31')] Executor (op=resolve_conflicts): Generated merge SQL
INFO     [PK ('3M', '2022-12-31')] Reasoning: The 8 rows represent both consolidated company-level metrics and
         business segment breakdowns for 3M for FY2022. There are true conflicts in the measured values for
         operating_income, net_sales, and operating_margin, but these are not errors---they reflect distinct
         entities: the consolidated company and its segments (Safety & Industrial, Transportation & Electronics,
         Health Care, Consumer). The schema is for company-level metrics, but the extraction includes
         segment-level rows, which are valuable for analysis. However, the merge objective is to deduplicate and
         resolve conflicts for the company-level row, not to preserve segment rows as separate events.

         Upon inspection, two rows (row_id=1 and row_id=10, also row_id=13) provide the consolidated
         company-level metrics:
         - operating_income: 6539
         - net_sales: 34229
         - operating_margin: 19.1% (explicit in row_id=1, calculated in row_id=10/13)

         Row_id=1 is extracted from a table labeled 'Total Company' and provides explicit values for all three
         metrics, with direct quotes and clear rationale. Row_id=10 and row_id=13 are from the Consolidated
         Statement of Income and Business Segment Information, also authoritative, but the operating margin is
         calculated, not quoted. Row_id=1 best matches the field definition alignment, extraction location
         relevance, and source authority criteria.

         Segment rows (row_id=2,4,6,8) are not company-level and should not be merged into the company-level
         output row. Row_id=0 is a summary/narrative row with less direct metric extraction.

         Therefore, the best approach is to select row_id=1 as the single authoritative company-level row for
         FY2022, preserving all original columns.
INFO     [PK ('3M', '2022-12-31')] SQL: SELECT * FROM OperatingMarginMetrics_69ffa681 WHERE row_id = 1
INFO     [PK ('3M', '2022-12-31')] Executor (op=resolve_conflicts): Executing merge SQL (attempt 1/3)
INFO     [PK ('3M', '2022-12-31')] Executor result: 8 rows $\rightarrow$ 1 rows
INFO     [PK ('3M', '2022-12-31')] Verification phase: Entering verification
INFO     [PK ('3M', '2022-12-31')] Verification: Inspection 1/5
INFO     [PK ('3M', '2022-12-31')] Verification: Reasoning: I need to verify that the SQL transformation selected reconciliation_v2.py:1743
         only the authoritative company-level row (row_id=1) from the initial_table and that the final_table
         contains exactly this row, with all columns preserved. To do this, I will inspect both tables:
         initial_table to confirm the presence and content of row_id=1, and final_table to ensure it contains
         only row_id=1 and matches the expected values and schema. This inspection will allow me to confirm that
         the SQL executed the intended logic described in my reasoning.
INFO     [PK ('3M', '2022-12-31')] Verification: SQL: SELECT * FROM initial_table WHERE row_id = 1;
         SELECT * FROM final_table;
INFO     [PK ('3M', '2022-12-31')] Verification: [OK] Approved after 1 inspection(s)
INFO     [PK ('3M', '2022-12-31')] Verification: I inspected both initial_table and final_table. The
         initial_table contains multiple rows for the primary key ('3M', '2022-12-31'), including both
         company-level and segment-level metrics. My reasoning for the merge was to select only the authoritative
         company-level row (row_id=1), which provides explicit, directly quoted values for operating_income,
         net_sales, and operating_margin, extracted from the 'Total Company' table. The final_table contains
         exactly one row, with row_id=1, and all columns are preserved. The values in final_table match those in
         initial_table for row_id=1, confirming that the SQL transformation executed the intended logic:
         selecting only the authoritative company-level row and excluding segment-level and summary rows.
         Therefore, the SQL correctly implemented my reasoning and the intended transformation.
INFO     [PK ('3M', '2022-12-31')] Context Generator: Starting
INFO     [PK ('3M', '2022-12-31')] Context Generator: [OK] Generated context
INFO     [PK ('3M', '2022-12-31')] Reduced to 1 rows, stopping
INFO     [PK ('3M', '2022-12-31')] Final result = 1 rows after 1 operations
INFO     ================================================================================
\end{lstlisting}

\subsection{Primary Key Selection}

Example of the primary key selector using voting rounds to identify the semantic primary key for reconciliation.

\begin{lstlisting}[
  basicstyle=\ttfamily\tiny,
  breaklines=true,
  frame=single,
  xleftmargin=0.3cm,
  xrightmargin=0.3cm
]
         ================================================================================
INFO     PHASE 1: PRIMARY KEY SELECTION
INFO     ================================================================================
INFO     Using Primary Key Selector Version: v1
INFO     === Selecting Primary Key for CashFlowStatementItem ===
INFO     === Running 3 voting rounds for primary key selection ===
INFO
         --- Voting Round 1/3 ---
INFO     --- Primary Key Analysis Query 1/5 (Run 1) ---
INFO     [OK] Primary key selected after 0 queries (Run 1)
INFO       Primary Key: ['entity_name', 'fiscal_year_end', 'item_name']
INFO       Reasoning: This table records individual cash flow statement line items for a given entity and
         fiscal year. Each unique observation...
INFO       Round 1 selected: ['entity_name', 'fiscal_year_end', 'item_name']
INFO
         --- Voting Round 2/3 ---
INFO     --- Primary Key Analysis Query 1/5 (Run 2) ---
INFO     [OK] Primary key selected after 0 queries (Run 2)
INFO       Primary Key: ['entity_name', 'fiscal_year_end', 'item_name']
INFO       Reasoning: This table records individual cash flow statement line items for a given entity and
         fiscal year. Each unique observation...
INFO       Round 2 selected: ['entity_name', 'fiscal_year_end', 'item_name']
INFO
         --- Voting Round 3/3 ---
INFO     --- Primary Key Analysis Query 1/5 (Run 3) ---
INFO     [OK] Primary key selected after 0 queries (Run 3)
INFO       Primary Key: ['entity_name', 'fiscal_year_end', 'item_name']
INFO       Reasoning: This table records individual cash flow statement line items for a given entity and
         fiscal year. Each unique observation...
INFO       Round 3 selected: ['entity_name', 'fiscal_year_end', 'item_name']
INFO
         === Majority Voting Result ===
INFO       Winning Primary Key: ['entity_name', 'fiscal_year_end', 'item_name']
INFO       Votes: 3/3
INFO       Vote Distribution: {('entity_name', 'fiscal_year_end', 'item_name'): 3}
INFO     Selected primary key for CashFlowStatementItem: ['entity_name', 'fiscal_year_end', 'item_name']
INFO     Reasoning: This table records individual cash flow statement line items for a given entity and fiscal
         year. Each unique observation is defined by the combination of entity_name (which company),
         fiscal_year_end (which reporting period), and item_name (which line item). Multiple extractions of the
         same entity, period, and item_name should be consolidated, with item_value being resolved during
         conflict resolution. All other fields are extraction metadata or rationale and do not define the
         business key.
INFO     [OK] Selected Primary Key: ['entity_name', 'fiscal_year_end', 'item_name']
INFO       Reasoning: This table records individual cash flow statement line items for a given entity and fiscal
         year. Each unique observation is defined by the combination of entity_name (which company),
         fiscal_year_end (which reporting period), and item_name (which line item). Multiple extractions of the
         same entity, period, and item_name should be consolidated, with item_value being resolved during
         conflict resolution. All other fields are extraction metadata or rationale and do not define the
         business key.
\end{lstlisting}

\subsection{Field Canonicalization}

Example of canonicalizing artist names to resolve variations (e.g., "Adele" vs "Adele Laurie Blue Adkins").

\begin{lstlisting}[
  basicstyle=\ttfamily\tiny,
  breaklines=true,
  frame=single,
  xleftmargin=0.3cm,
  xrightmargin=0.3cm
]
         --- Processing Field 2/3: artist_name ---
INFO
           Canonicalization Cycle 1/20
INFO         Action: canonicalize
INFO         Reasoning: Phase 3 Step 2: I examined the alphabetically sorted distinct values from
         Inspection 1:
         - Adele (6 occurrences): Clean, common artist label.
         - Adele Laurie Blue Adkins (1 occurrence): Full legal name of Adele.

         These appear to refer to the same person. According to best practice for primary key
         canonicalization, we need to check related columns for evidence before merging.

         First, the artist_name_quote field has only 2 distinct values matching the artist name,
         indicating these are alternate forms used in the data. Next, the very low cardinality (2 values,
         both in artist_name-related quote/rationale fields, not free text and clearly person names)
         confirms that this is a categorical field, not free text or a unique identifier.

         To ensure deduplication, we should canonicalize both forms to a single canonical form ('ADELE'),
         applying UPPER for case standardization. The rationale/quote columns simply confirm that "Adele
         Laurie Blue Adkins" is the legal name of Adele, but all other data refers to the artist as
         "Adele". Thus, I will canonicalize both to 'ADELE'.
INFO         Canonicalizing field: artist_name (attempt 1/3)
INFO         SQL: SELECT row_id,
           CASE
             WHEN artist_name IS NULL THEN NULL
             WHEN UPPER(TRIM(artist_name)) IN ('ADELE', 'ADELE LAURIE BLUE ADKINS') THEN 'ADELE'
             ELSE UPPER(TRIM(artist_name))
           END AS artist_name,
           page_number,
           document_name,
           text_header,
           artist_name_quote,
           artist_name_rationale,
           wikidata_id_quote,
           wikidata_id_rationale,
           wikidata_id,
           birth_date_quote,
           birth_date_rationale,
           birth_date,
           birth_place_quote,
           birth_place_rationale,
           birth_place,
           country_of_birth_quote,
           country_of_birth_rationale,
           country_of_birth,
           alternate_names_quote,
           alternate_names_rationale,
           alternate_names,
           debut_date_quote,
           debut_date_rationale,
           debut_date,
           debut_year_quote,
           debut_year_rationale,
           debut_year,
           debut_description_quote,
           debut_description_rationale,
           debut_description,
           industry_quote,
           industry_rationale,
           industry
         FROM artistBiography_Adele
         ORDER BY row_id;
INFO         [OK] Successfully applied canonicalization 1
\end{lstlisting}

%%%%%%%%%%%%%%%%%%%%%%%%%%%%%%%%%%%%%%%%%%%%%%%%%%%%%%%%%%%%%%%%%%%%%%%%%%%%%%%
%%%%%%%%%%%%%%%%%%%%%%%%%%%%%%%%%%%%%%%%%%%%%%%%%%%%%%%%%%%%%%%%%%%%%%%%%%%%%%%

\section{Broad Societal Impact}
This work has several potential positive societal impacts. Systematic reviews are central to evidence-based decision making in medicine, social science, policy analysis, finance, law, and other technical domains, but producing them manually requires substantial expert labor and can take many months. By automating parts of evidence discovery, extraction, reconciliation, and synthesis, \system{} could reduce the cost and time required to produce systematic-review-style analyses. This may help researchers, analysts, policymakers, and domain experts examine larger evidence bases, ask follow-up questions more easily, and produce syntheses with more explicit provenance than standard retrieval-first systems. In domains such as healthcare and public policy, improved evidence synthesis could support better-informed decisions, provided that outputs remain subject to expert review.

At the same time, the work also has potential negative societal impacts. Because \system{} is designed to synthesize evidence across large document collections, errors in extraction, reconciliation, or answer synthesis could be amplified into apparently authoritative conclusions. This risk is especially important in high-stakes domains such as medicine, law, finance, and policy, where incorrect or incomplete syntheses could mislead decision makers. The system may also inherit biases from the underlying document corpus or language models, omit relevant evidence despite its goal of broad coverage, or overstate certainty when the available evidence is incomplete or conflicting. More broadly, automation of systematic-review-style workflows may encourage overreliance on machine-generated syntheses, reduce incentives for careful manual validation, and advantage organizations with access to large-scale compute and proprietary document collections. For these reasons, \system{} should be used as an assistive tool rather than a replacement for expert judgment, with careful auditing of extracted evidence, provenance, uncertainty, and failure cases.
\section*{NeurIPS Paper Checklist}

\begin{enumerate}

\item {\bf Claims}
    \item[] Question: Do the main claims made in the abstract and introduction accurately reflect the paper's contributions and scope?
    \item[] Answer: \answerYes{} % Replace by \answerYes{}, \answerNo{}, or \answerNA{}.
    \item[] Justification: The abstract and introduction claim that SLIDERS constructs question-specific evidence tables, reconciles distributed or conflicting evidence, supports follow-up questions, and improves over retrieval, long-context, and agentic baselines. These claims are directly supported by the reported results on BrowseComp+, WikiCeleb100, FinQ100, FinanceBench, Loong, and Oolong, including the reported improvements on large-corpus and follow-up-analysis settings.

\item {\bf Limitations}
    \item[] Question: Does the paper discuss the limitations of the work performed by the authors?
    \item[] Answer: \answerYes{}
    \item[] Justification: The paper includes a limitations section and additional discussion of failure modes. It discusses schema dependence, extraction incompleteness, reconciliation errors, difficulty with exact counting and numerical questions, higher latency than single-call baselines, residual LLM-judge noise, and the need for human verification in high-stakes domains

\item {\bf Theory assumptions and proofs}
    \item[] Question: For each theoretical result, does the paper provide the full set of assumptions and a complete (and correct) proof?
    \item[] Answer: \answerNA{}
        \item[] Justification: The paper gives algorithmic definitions and a complexity-motivation argument for grouped reconciliation, but it does not present formal theorems or lemmas requiring proofs. The reconciliation complexity discussion is used to motivate the algorithmic design rather than as a formal theoretical result.

    \item {\bf Experimental result reproducibility}
    \item[] Question: Does the paper fully disclose all the information needed to reproduce the main experimental results of the paper to the extent that it affects the main claims and/or conclusions of the paper (regardless of whether the code and data are provided or not)?
    \item[] Answer: \answerYes{}
    \item[] Justification: The paper fully defines all the components and models used for the SLIDERS pipelines. All the prompts and parameters needed for reproduction are mentioned in the appendix. In addition, the complete source code is released at \url{https://github.com/stanford-oval/sliders}.

\item {\bf Open access to data and code}
    \item[] Question: Does the paper provide open access to the data and code, with sufficient instructions to faithfully reproduce the main experimental results, as described in supplemental material?
    \item[] Answer: \answerYes{} % Replace by \answerYes{}, \answerNo{}, or \answerNA{}.
    \item[] Justification: The complete source code for SLIDERS is released at \url{https://github.com/stanford-oval/sliders}. The new benchmarks are included there as JSON files, together with scripts for ingesting the public datasets used.

\item {\bf Experimental setting/details}
    \item[] Question: Does the paper specify all the training and test details (e.g., data splits, hyperparameters, how they were chosen, type of optimizer) necessary to understand the results?
    \item[] Answer: \answerYes{} % Replace by \answerYes{}, \answerNo{}, or \answerNA{}.
    \item[] Justification: The question ids included in the subset of Loong used are included in the appendix. The models used are mentioned in main text. Hyperparameters such as chunk size are given in the appendix.

\item {\bf Experiment statistical significance}
    \item[] Question: Does the paper report error bars suitably and correctly defined or other appropriate information about the statistical significance of the experiments?
    \item[] Answer: \answerNo{} % Replace by \answerYes{}, \answerNo{}, or \answerNA{}.
    \item[] Justification: Running the main experiments multiple times with different seeds would incur significant costs. The substantial gain over baselines and persistence of gains across closed and open sourced models and across types of benchmarks adds confidence to the results. We include statistical tests for LLM-as-a-judge evaluation and schema generation.

\item {\bf Experiments compute resources}
    \item[] Question: For each experiment, does the paper provide sufficient information on the computer resources (type of compute workers, memory, time of execution) needed to reproduce the experiments?
    \item[] Answer: \answerYes{} % Replace by \answerYes{}, \answerNo{}, or \answerNA{}.
    \item[] Justification: The paper only has inference time experiments. We include latency analysis in the appendix and compute used for the open-source model experiments.
    
\item {\bf Code of ethics}
    \item[] Question: Does the research conducted in the paper conform, in every respect, with the NeurIPS Code of Ethics \url{https://neurips.cc/public/EthicsGuidelines}?
    \item[] Answer: \answerYes{}{} % Replace by \answerYes{}, \answerNo{}, or \answerNA{}.
    \item[] Justification: We have reviewed the Code of Ethics and conform to it.

\item {\bf Broader impacts}
    \item[] Question: Does the paper discuss both potential positive societal impacts and negative societal impacts of the work performed?
    \item[] Answer: \answerYes{} % Replace by \answerYes{}, \answerNo{}, or \answerNA{}.
    \item[] Justification: We add broader societal impacts both positive and negative.
    
\item {\bf Safeguards}
    \item[] Question: Does the paper describe safeguards that have been put in place for responsible release of data or models that have a high risk for misuse (e.g., pre-trained language models, image generators, or scraped datasets)?
    \item[] Answer: \answerNA{} % Replace by \answerYes{}, \answerNo{}, or \answerNA{}.
    \item[] Justification: The paper does not release any artifacts with potential for misuse. The benchmarks proposed in the paper were created from open-source wikipedia pages and SEC filings.

\item {\bf Licenses for existing assets}
    \item[] Question: Are the creators or original owners of assets (e.g., code, data, models), used in the paper, properly credited and are the license and terms of use explicitly mentioned and properly respected?
    \item[] Answer: \answerYes{} % Replace by \answerYes{}, \answerNo{}, or \answerNA{}.
    \item[] Justification: We add the license for Wikipedia and SEC in the appendix.

\item {\bf New assets}
    \item[] Question: Are new assets introduced in the paper well documented and is the documentation provided alongside the assets?
    \item[] Answer:  \answerYes{} % Replace by \answerYes{}, \answerNo{}, or \answerNA{}.
    \item[] Justification: We provide the instructions for how we collected the Wikipedia articles and SEC filings.

\item {\bf Crowdsourcing and research with human subjects}
    \item[] Question: For crowdsourcing experiments and research with human subjects, does the paper include the full text of instructions given to participants and screenshots, if applicable, as well as details about compensation (if any)? 
    \item[] Answer: \answerNA{} % Replace by \answerYes{}, \answerNo{}, or \answerNA{}.
    \item[] Justification: We do not perform any human study.

\item {\bf Institutional review board (IRB) approvals or equivalent for research with human subjects}
    \item[] Question: Does the paper describe potential risks incurred by study participants, whether such risks were disclosed to the subjects, and whether Institutional Review Board (IRB) approvals (or an equivalent approval/review based on the requirements of your country or institution) were obtained?
    \item[] Answer: \answerNA{} % Replace by \answerYes{}, \answerNo{}, or \answerNA{}.
    \item[] Justification: We do not perform any human study.

\item {\bf Declaration of LLM usage}
    \item[] Question: Does the paper describe the usage of LLMs if it is an important, original, or non-standard component of the core methods in this research? Note that if the LLM is used only for writing, editing, or formatting purposes and does \emph{not} impact the core methodology, scientific rigor, or originality of the research, declaration is not required.
    %this research? 
    \item[] Answer: \answerYes{} % Replace by \answerYes{}, \answerNo{}, or \answerNA{}.
    \item[] Justification: \system{} is an LLM-based pipeline: LLMs generate the evidence variables, extract evidence from chunks, run the reconciliation agent, and synthesize answers. Section~3 and Appendix~A.4 state which model is used at each stage, and Appendix~F lists all prompts.

\end{enumerate}

\end{document}